%% file: main.tex
\documentclass[10pt, twocolumn, letterpaper]{article}

\usepackage[pagenumbers]{cvpr} 

\input{preamble}
\definecolor{cvprblue}{rgb}{0.21,0.49,0.74}
\usepackage[pagebackref, breaklinks, colorlinks, allcolors=cvprblue]{hyperref}
\usepackage{microtype} 
\usepackage[utf8]{inputenc} 
\usepackage[T1]{fontenc} 
\usepackage{url} 
\usepackage{booktabs} 
\usepackage{amsfonts} 
\usepackage{nicefrac} 
\usepackage{xcolor} 
\usepackage{colortbl}
\usepackage{balance}
\usepackage{multirow}
\usepackage{graphicx}
\usepackage{subcaption}
\usepackage{float}
\usepackage{amsmath}
\usepackage{mathtools}
\usepackage{diagbox}
\usepackage{array}
\usepackage{ragged2e}
\usepackage{tabularx}
\usepackage{tabularray}
\usepackage[export]{adjustbox}
\usepackage{makecell}
\usepackage{nicematrix}
\usepackage{silence}
\colorlet{lightred}{red!15}
\newcommand{\blackhref}[2]{\href{#1}{\textcolor{black}{#2}}}
\def\paperID{43348} 
\def\confName{CVPR}
\def\confYear{2026}

\title{Unified Spherical Frontend: Learning Rotation-Equivariant Representations
of Spherical Images from Any Camera}

\makeatletter
\renewcommand{\@fnsymbol}[1]{\ifcase#1\or 1\or 2\or 3\or 4\or 5\else\arabic{#1}\fi}
\makeatother

\author{
{\large Mukai Yu \quad Mosam Dabhi \quad Liuyue Xie \quad Sebastian Scherer \quad László A. Jeni}\\
{\normalsize Carnegie Mellon University, Robotics Institute}\\
{\tt\footnotesize \{\blackhref{mailto:mukaiy@andrew.cmu.edu}{mukaiy},\blackhref{mailto:mdabhi@andrew.cmu.edu}{mdabhi},\blackhref{mailto:liuyuex@andrew.cmu.edu}{liuyuex},\blackhref{mailto:basti@andrew.cmu.edu}{basti},\blackhref{mailto:laszloaj@andrew.cmu.edu}{laszloaj}\}@andrew.cmu.edu}
}

\begin{document}
\maketitle

\input{sec/0_abstract}
\input{sec/1_introduction}
\input{sec/2_related_work}
\input{sec/3_method}
\input{sec/4_experiment}
\input{sec/5_conclusion}

\clearpage
\input{sec/6_acknowledgement}
{ \small \bibliographystyle{ieeenat_fullname} \bibliography{main} }

\input{sec/X_supplement}

\end{document}

%% file: sec/0_abstract.tex
\begin{abstract}
	Modern perception increasingly relies on fisheye, panoramic, and other wide field-of-view (FoV) cameras, yet most pipelines still apply planar CNNs designed for pinhole imagery on 2D grids, where pixel-space neighborhoods misrepresent physical adjacency and models are sensitive to global rotations.
	Traditional spherical CNNs partially address this mismatch but require costly spherical harmonic transform that constrains resolution and efficiency.
	We present Unified Spherical Frontend (USF), a distortion-free lens-agnostic framework that transforms images from any calibrated camera onto the unit sphere via ray-direction correspondences, and performs spherical resampling, convolution, and pooling canonically in the spatial domain.
	USF is modular: projection, location sampling, value interpolation, and resolution control are fully decoupled.
	Its configurable distance-only convolution kernels offer rotation-equivariance, mirroring translation-equivariance in planar CNNs while avoiding harmonic transforms entirely.
	We compare multiple standard planar backbones with their spherical counterparts across classification, detection, and segmentation tasks on synthetic (Spherical MNIST) and real-world (PANDORA, Stanford 2D-3D-S) datasets, and stress-test robustness to extreme lens distortions, varying FoV, and arbitrary rotations.
	USF scales efficiently to high-resolution spherical imagery and maintains \emph{less than 1\%} performance drop under random test-time rotations without training-time rotational augmentation, and enables \emph{zero-shot} generalization to any unseen (wide-FoV) lenses with minimal performance degradation.
	\footnote{Accepted to CVPR 2026. Website: \href{https://tomnotch.com/USF}{tomnotch.com/USF}}
\end{abstract}

%% file: sec/1_introduction.tex
\section{Introduction}
\label{sec:introduction}
Spherical signals arise naturally in many domains - from astrophysics and global climate modeling to omnidirectional perception in robotics and virtual reality.
However, modern computer vision pipelines overwhelmingly rely on planar convolutional neural networks (CNNs), which assume a standard pinhole camera model and operate strictly in the 2D image domain.
This design becomes problematic as real-world perception systems increasingly adopt fisheye, panoramic, and other wide field-of-view (FoV) lenses.
Directly applying planar CNNs to such distorted imagery often leads to suboptimal performance, as image-space neighborhoods do not reflect true physical adjacency.

\input{fig/Unified_Spherical_Representation}

The core issue is a mismatch between the domain of \emph{processing} and the domain of \emph{geometry}.
In a planar CNN, convolution aggregates information from adjacent pixels on a 2D plane, whose proximity is defined by the image grid instead of the physical configuration of light rays.
This discrepancy becomes especially pronounced in wide-FoV images, where physically adjacent pixels (e.g., near the poles of an equirectangular panorama) may appear far apart after projection.
This distortion is not incidental, but a fundamental limitation.
According to Gauss's Theorema Egregium~\cite{gauss1828,carmoDifferentialGeometryCurves1976}, no 2D projection can preserve the intrinsic curvature of the sphere.
As a result, any attempt to represent spherical signals on a flat image inevitably introduces distortion, undermining the spatial assumptions that planar CNNs rely on.
Moreover, since the convolution kernel is fixed in image coordinates, it inherently encodes the coordinate frame of the image, leading to models that are dependent on global rotations.

To address these limitations, we propose \emph{Unified Spherical Frontend (USF)}, a generic, modular, and lens-agnostic framework that lifts vision pipelines from the image plane to the sphere.
As illustrated in \cref{fig:framework}, USF transforms input planar images from cameras with known intrinsics into spherical signal. It then applies resampling, convolution, and pooling operations entirely on the sphere, abstracting away lens distortions and preserving physical geometry.
This representation enables a new class of spherical CNNs that exhibit built-in rotation-equivariance and cross-lens adaptability without requiring task-specific designs or heavy augmentation.

USF is built to be flexible and composable.
Every stage, including projection, resampling, convolution, and pooling, is decoupled, allowing users to swap in different location samplers, value interpolators, or output resolutions.
Moreover, we show that distance-only weighting functions guarantee rotation-equivariance by construction, enabling models to be robust to arbitrary SO(3) transformations via architectural bias rather than data augmentation.

We evaluate USF on a wide range of vision tasks: MNIST digit classification, object detection on panoramic images, and semantic segmentation across lenses.
Our results show that USF maintains competitive performance while demonstrating superior robustness to rotation and zero-shot generalization across unseen wide-FoV lenses.
In particular, our spherical CNN layers support plug-and-play replacement of planar layers, enabling general-purpose spherical vision.

Our contributions are summarized as follows:
\vspace{0.5em}
\begin{itemize}
	\item We propose a unified and lens-agnostic spherical vision pipeline that processes arbitrary camera inputs in a geometry-aware and rotation-equivariant manner.

	\item We design a modular spherical resampling module composed of decoupled and configurable location sampling and value interpolation stages.

	\item We introduce an expressive and efficient spherical convolution kernel with decomposable distance and direction weighting, and demonstrate how architectural design ensures equivariance.

	\item We validate our approach on three representative tasks, demonstrating zero-shot lens generalization, robustness against random rotation, and competitive performance compared to standard planar models.
\end{itemize}

%% file: fig/Unified_Spherical_Representation.tex
\begin{figure}[!t]
	\centering
	\includegraphics[width=0.8\linewidth]{
		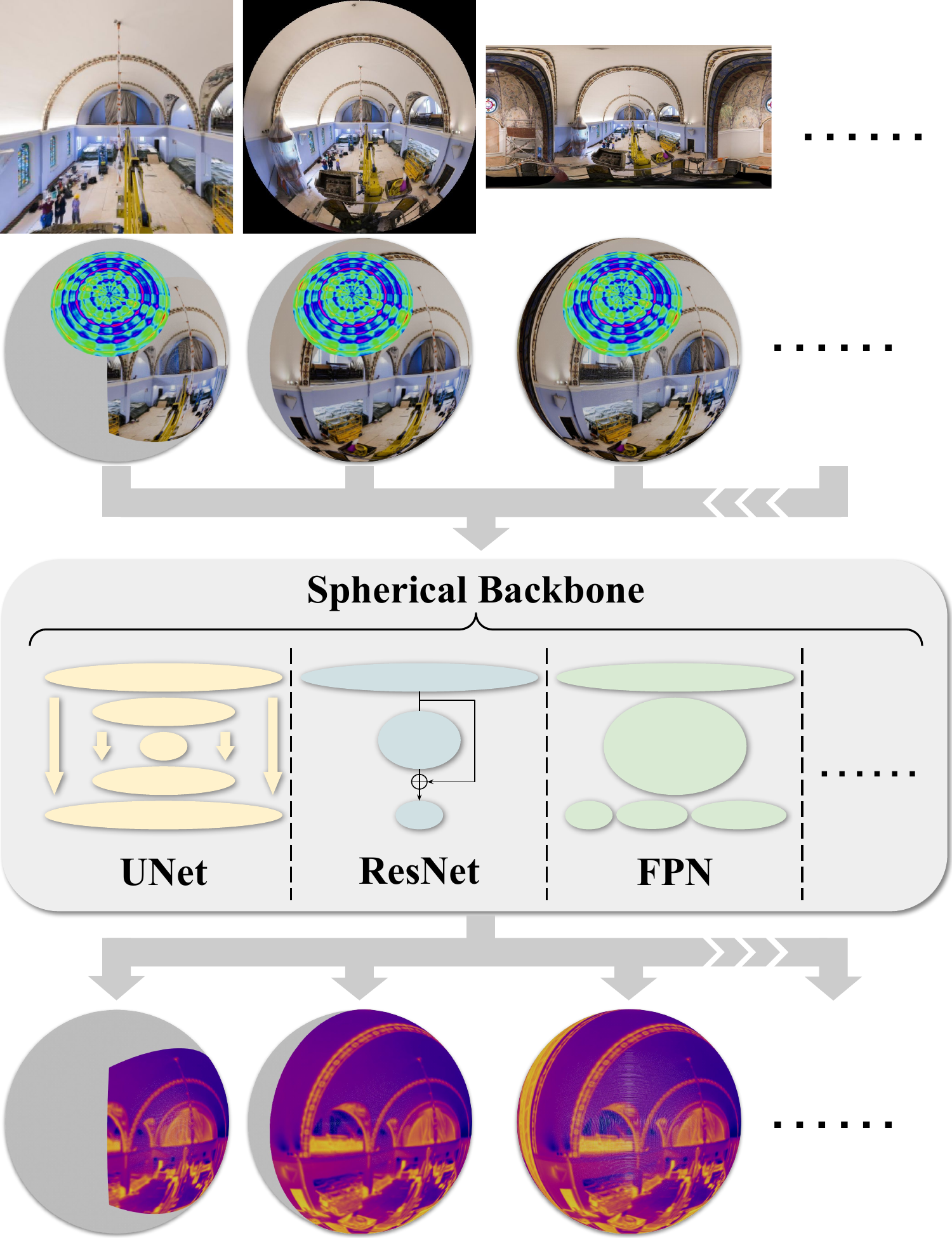
	}
	\caption{\textbf{Unified Spherical Representation.}
		From any camera to any architecture: a unified spherical pipeline for modern vision.}
	\label{fig:unified spherical representation}
\end{figure}

%% file: sec/2_related_work.tex
\section{Related Work}
\label{sec:related-work}
\input{fig/Equivariance_And_Invariance}

\paragraph{Lens Distortion and Panoramic Perception.}
Handling wide-FoV and heavily distorted imagery from fisheye and panoramic lenses challenges planar CNNs, which assume a pinhole model and fixed grid sampling.
Directly applying planar convolutions introduces spatial bias, as image-domain adjacency does not necessarily reflect physical proximity.
Several methods address this by adapting to spherical geometry: SphereNet~\cite{coorsSphereNetLearningSpherical2018} samples features on tangent planes, while ~\cite{leeSpherePHDApplyingCNNs2019a,zhangHexNetOrientationAwareDeep2023,yangUVSCNNsConstructingGeneral2024,defferrardDeepSphereGraphbasedSpherical2020,masciGeodesicConvolutionalNeural2015} define specialized kernels on polyhedral, mesh, or graph.
These techniques, while geometrically informed, depend on handcrafted grids, predefined connectivity, or structured sampling schemes tailored to full-sphere panoramic cameras.
In contrast, our approach treats spherical data as an \emph{unstructured}, \emph{unordered} set of points, enabling flexible processing across arbitrary lens types without assuming any mesh, grid, or sequential order.
\input{fig/Pipeline}

\paragraph{Spherical CNNs and Rotation Equivariance.}
Spectral methods~\cite{cohenSphericalCNNs2018,estevesLearningSO3Equivariant2020} achieve exact $\mathrm{SO}(3)$-equivariance (bottom-half of \cref{fig:equivariance & invariance diagram}) by performing convolution in the spherical harmonics domain.
Spatial-domain methods~\cite{jiangSphericalCNNsUnstructured2019,leeSpherePHDApplyingCNNs2019a} define convolution using local graph neighborhoods or predefined meshes, but often relax equivariance or assume specific structures.
DISCO~\cite{ocampoScalableEquivariantSpherical2023} advances spatial filtering via learned radial-directional kernels but focuses on dense full-sphere signals and fixed discretizations, thereby lacking flexibility for partial views.
Our method remains entirely in the spatial domain, supports partial-sphere coverage, scales efficiently to dense signals, and retains built-in rotation-equivariance, making it well-suited for high-resolution perception tasks with any lens.

\paragraph{Detection and Segmentation on the Sphere.}
Modern detection and segmentation architectures, such as YOLOv11~\cite{khanamYOLOv11OverviewKey2024}, DeepLab v3~\cite{chenRethinkingAtrousConvolution2017}, and UNet~\cite{ronnebergerUNetConvolutionalNetworks2015}, are designed primarily for distortion-free pinhole images and lack built-in rotation-equivariance.
Extensions like R-CenterNet~\cite{xuPANDORAPanoramicDetection2022} adapt detection heads to panoramic data but still depend on planar backbones, which degrade under arbitrary rotations.
While effective within their target domains, these models are limited in generalizability across lens types and robustness against test-time rotations.
In our experiments, we empower these architectures with spherical vision by replacing their planar layers with rotation-equivariant spherical counterparts, enhancing their robustness to unseen camera lenses and random rotations without altering their macro design.

%% file: fig/Equivariance_And_Invariance.tex
\begin{figure}[!t]
	\centering
	\includegraphics[width=\linewidth]{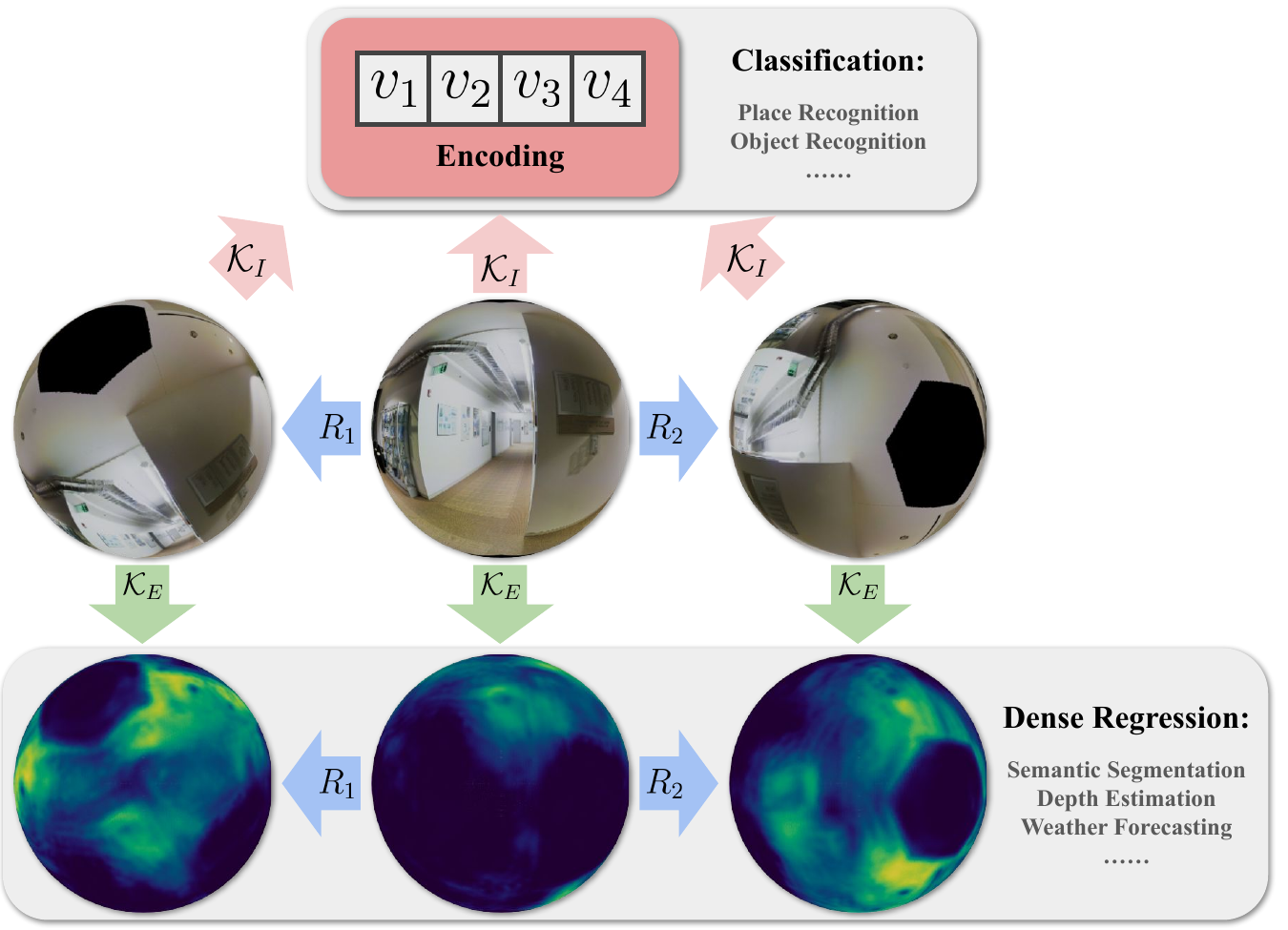}

	\caption{\textbf{Rotation Equivariance and Invariance.}
		A function $\mathcal{K}$ is \textcolor{OliveGreen}{rotation-equivariant} if $\textcolor{OliveGreen}{\mathcal{K}_E}(R \cdot \mathbf{x}) = R \cdot \textcolor{OliveGreen}{\mathcal{K}_E}(\mathbf{x})$, and \textcolor{BrickRed}{rotation-invariant} if $\textcolor{BrickRed}{\mathcal{K}_I}(R \cdot \mathbf{x}) = \textcolor{BrickRed}{\mathcal{K}_I}(\mathbf{x})$, for all $R \in \mathrm{SO}(3)$.}
	\label{fig:equivariance & invariance diagram}
\end{figure}

%% file: fig/Pipeline.tex
\begin{figure*}[!ht]
	\centering
	\includegraphics[width=\linewidth]{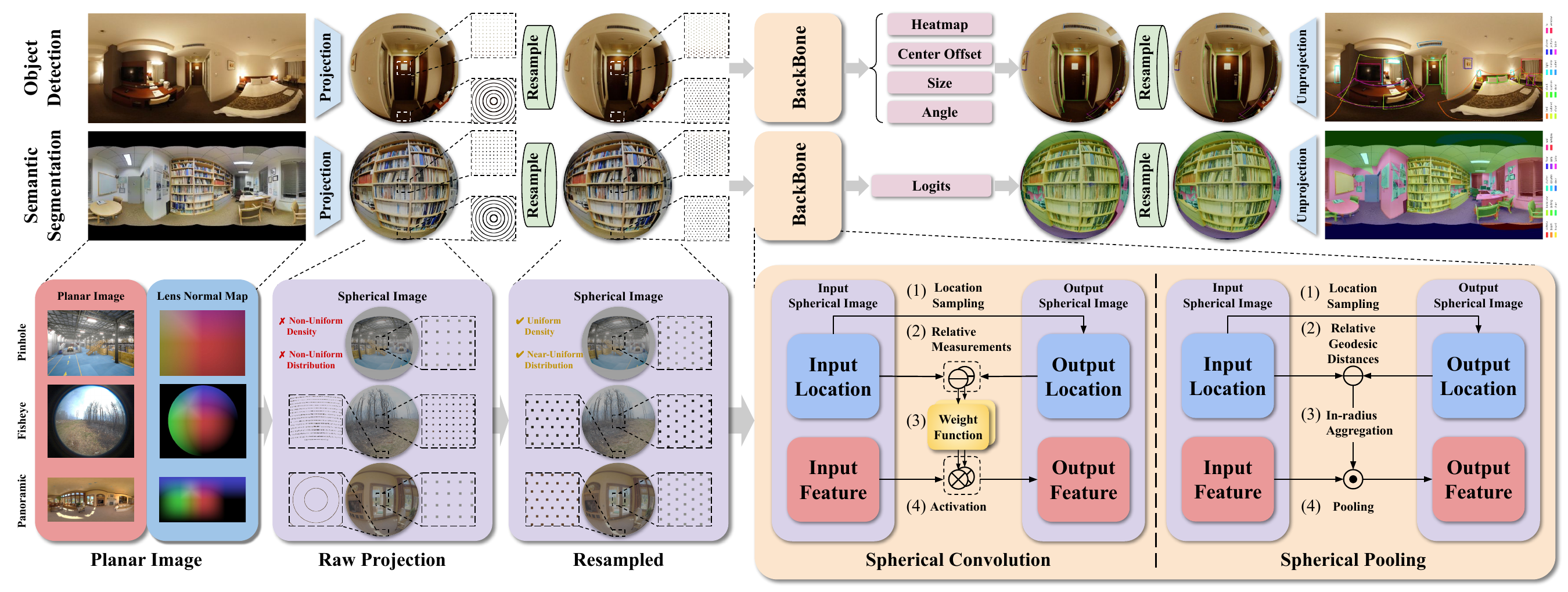}
	\caption{\textbf{Unified Spherical Frontend.}
		(i) A \textcolor{BrickRed}{planar image} and its \textcolor{blue}{lens normal map} can be combined to form a (ii) \textcolor{violet}{spherical image}.
		Cameras with different lenses produce spatially varying densities and distributions of pixels when projected onto the sphere.
		Thus, it is crucial to perform (iii) resampling before (iv) feeding into the backbone composed of spherical convolution and pooling layer.
		Optionally, the results can be (v) resampled back into the raw projected \textcolor{violet}{spherical image} pixel locations, and (vi) unproject back to the \textcolor{BrickRed}{planar image} for downstream integration.}
	\label{fig:framework}
\end{figure*}

%% file: sec/3_method.tex
\section{Methodology}
\label{sec:methodology}

We propose \emph{Unified Spherical Frontend}, a lens-agnostic and task-agnostic strategy to approach generic computer vision applications.
We begin by introducing its core components: spherical projection and resampling, spherical convolution, and pooling.
These components are then integrated into a complete, modular pipeline, which we instantiate on representative applications in \cref{sec:experiments}.


\subsection{Spherical Projection and Resampling}
\label{ssec:spherical-projection-and-resampling}

Given an input image with calibrated intrinsics under any camera model (e.g., pinhole, fisheye, or panoramic), we derive the per-pixel geometry as unit-norm $\mathbb{R} ^{3}$ vectors.
Collectively, they form a \emph{lens normal map}\footnote{Sometimes called \emph{ray map} in relevant literature~\cite{zhangCamerasRaysPose2024}}, which defines a \emph{bijective} mapping between image coordinates and spherical coordinates.
Formally, each image coordinate $\mathbf{u}\in \mathbb{R}^{2}$ maps to a ray direction $\mathbf{p}_{\mathbf{u}}\in \mathbb{S}^{2}$, with subpixel rays obtained via planar interpolation.
Traditional camera models approximate the mapping $\mathbf{p}_{\mathbf{u}}$ using a low-dimensional parametric form (e.g., 4-9 camera intrinsic parameters~\cite{hartleyMultipleViewGeometry2018,kannalaGenericCameraModel2006,wangNewCalibrationModel2006}).
In contrast, dense ray maps represent a full-rank mapping $\mathbb{R}^{2}\rightarrow \mathbb{S}^{2}$ over all pixel coordinates.

Projecting all pixels onto the unit sphere yields a spherical image, in which geometry (unit norm $\mathbb{R}^{3}$ vectors and polar coordinates) and scalar values (e.g., RGB colors, features) are explicitly separated and associated.
Unlike prior works~\cite{yangUVSCNNsConstructingGeneral2024,jiangSphericalCNNsUnstructured2019,coorsSphereNetLearningSpherical2018}, we do not assume any structured grid, mesh connectivity, or predefined sequential ordering over spherical points.
Instead, our formulation treats the input as an unordered set of sample values located on the sphere.
However, since the spherical points originate from the direct projection of planar images, they collectively exhibit non-uniform density and distribution, which hinders learning efficiency and creates undesired bias towards densely sampled regions, such as poles in panoramic images.
To mitigate this, it is crucial to resample the spherical image into a near-uniform distribution on the sphere with matching density, and without information loss.
This involves two decoupled steps: (1) location sampling to select new points with improved spatial distribution, and (2) value interpolation to assign feature values to those points based on inputs.

\subsubsection{Location Sampling}
\label{sssec:location-sampling}
This step selects new pixel locations with near-uniform distribution, matching the density and FoV coverage of the input spherical image from raw projection.

\input{fig/Sampling}

It is well known~\cite{alishahiSphericalEnsembleUniform2015,bonoMostUniformDistribution2024,cohenSphericalCNNs2018} that no perfectly uniform discretization of the sphere exists under all uniformity metrics.
Therefore, as illustrated in \cref{fig:sampling}, we provide several sampling schemes, including the Goldberg polyhedron~\cite{goldbergClassMultiSymmetricPolyhedra1937}, HEALPix~\cite{gorskiHEALPixFrameworkHighResolution2005,gorskiHEALPixPrimer1999a}, Fibonacci lattice~\cite{gonzalezMeasurementAreasSphere2010,swinbankFibonacciGridsNovel2006}, and quasi-random sampling~\cite{sobolDistributionPointsCube1967,haltonEfficiencyCertainQuasirandom1960} and systematically evaluate their effects on rotation-equivariance.\footnote{Sampler implementation details in the supplementary material}

To match the pixel density of an input spherical image, we estimate the average area per input pixel and adjust the number of output points accordingly.
The average pixel area is computed as the mean of the lower $75\%$ quantile of spherical-Voronoi cell areas, which is robust to outliers near the image boundary.

The raw output points of a location sampler distribute across the entire sphere.
Thus, to determine whether a newly sampled point falls within the input spherical image's FoV, we compare its geodesic distance to nearby input points.
Specifically, let the minimum radius be twice the average nearest-1-neighbor distance among all input spherical points; an output point is considered outside the FoV if its distance to the closest input point exceeds this threshold.

\subsubsection{Value Interpolation}
\label{sssec:value-interpolation}
This step assigns scalar values (pixel intensities) to the newly sampled spherical locations.
Since the input pixels do not have regular grid or mesh connectivity, standard bilinear interpolation is not applicable.
We therefore adopt a two-stage procedure consisting of neighborhood aggregation and local weighting.

\paragraph{Neighborhood aggregation.}
For each output location $\mathbf{p}_{o}$, its neighborhood $\mathcal{N}(\mathbf{p}_{o})$ is defined either as the $N$ nearest input points or as all input points within a circular cap of radius $r$ centered at the output.
Nearest-neighbor aggregation is useful for discrete label interpolation, similar to \textit{nearest\_exact} interpolation in planar images.
However, it can introduce discontinuities and weaken rotation equivariance, especially when points are sparse, because it selects the closest point based solely on distance, ignoring rotational symmetry.

\paragraph{Local weighting.}
Given an output point and its local neighborhood, interpolation weights can be computed using either radial or spectral methods.
A common approach is to use a \textit{radial basis function (RBF)}\footnote{RBF weighting scheme details in the supplementary material} with respect to the geodesic distance, which ensures rotational symmetry.
This yields interpolation of the form $x_{o}= \sum\omega_{k}x_{k}$, where each weight $\omega_{k}$ depends only on the distance between input and output locations.
Alternatively, one may fit a bandlimited spherical harmonic model to the neighborhood, analogous to \textit{moving-least-squares (MLS)} regression.

\vspace{0.5em}
Both stages of the resampling pipeline are deterministic with respect to geometry.
For location sampling, the set of output locations is fully determined by the input pixel locations and the chosen sampling method.
For value interpolation, the neighborhood structure and interpolation weights are fixed given two sets of input and output locations.
Although the pixel intensities or feature values may vary arbitrarily across frames, the geometric relations remain constant for a given camera.
Hence, the entire resampling pipeline is geometry-cacheable: once the geometric mappings are computed, subsequent inference reuses them with negligible overhead.


\subsection{Spherical Convolution and Pooling}
\label{ssec:spherical-convolution-and-pooling}
We begin by revisiting planar CNNs from the perspective of location-based feature aggregation, then generalize to the spherical domain.

\subsubsection{Planar CNN}
\label{sssec:planar-cnn}
In a standard planar CNN, each output feature is computed by aggregating values from a regular neighborhood around an output location using a square kernel:
\begin{align}
	x_{o}                       & = \mathcal{K}_{\text{conv}}(\mathcal{X}_{i}, \mathbf{p}_{o}) = \sum\limits_{k\in\mathcal{N}(\mathbf{p}_o)}x_{k}\; \omega_{k},             \\
	\mathcal{N}(\mathbf{p}_{o}) & = \bigg\{ k: \mathbf{p}_{k}\in \mathcal{P}_{i}, \; \mathbf{p}_{k}= \mathbf{p}_{o}\pm \lfloor \frac{\text{kernel size}}{2}\rfloor \bigg\}.
\end{align}

Planar CNN achieves the same effect as frequency-domain convolution by applying linear aggregation directly in the spatial domain, due to the convolution theorem.
Translation-equivariance arises from applying the same kernel at all positions, which also reduces the number of parameters compared to dense MLPs.
However, the relative distance and direction of neighbor pixels are inherently coupled in the kernel, and rotating the input image changes how the kernel aligns with features, since the kernel's orientation is fixed with the coordinate frame of the image plane, i.e., tied to a \textit{global gauge}.
As a result, planar convolutions are not rotation-equivariant and typically rely on random rotation data augmentation to stay robust against arbitrary unknown rotations at test time.

\input{tab/Generic_Spherical_CNN}

\subsubsection{Generic Spherical CNN}
\label{sssec:generic-spherical-cnn}


Similar to planar CNNs, spherical CNNs can achieve the same filtering function as convolution in the spherical harmonics domain like~\cite{estevesLearningSO3Equivariant2020,cohenSphericalCNNs2018,estevesScalingSphericalCNNs2023a}, by simply performing linear activation or correlation in the spatial domain.\footnote{Relevant proofs in the supplementary material} This avoids the high computational cost of spherical harmonic transforms, which require a large bandlimit $\ell$ for lossless reconstruction on dense signals, as dictated by the Nyquist–Shannon sampling theorem~\cite{driscollComputingFourierTransforms1994}.

However, unlike the planar counterpart, there is no structured grid or predefined mesh on the sphere.
An image projected onto the unit sphere becomes an unordered set of samples on $\mathbb{S}^{2}$, often non-uniform depending on the lens or projection, so each input pixel can lie at an arbitrary location relative to a given output point, which happens even after resampling.
Thus, we define spherical convolution as a local aggregation over input points within a circular cap centered at each output location.
In other words, the neighborhood $N(\mathbf{p}_{o})$ of an output point $\mathbf{p}_{o}$ is the set of all input points $\mathbf{p}_{k}$ whose geodesic distance $d(\mathbf{p}_{k}, \mathbf{p}_{o}) \le r$.
This respects the sphere's geometry and does not impose a rigid grid on the data.
On the contrary, irregular-shaped neighborhood aggregation, such as square~\cite{coorsSphereNetLearningSpherical2018}, hexagonal, or pentagonal kernels, would compromise rotation-equivariance because they are not radial.

This motivates our design of a generic spherical convolution kernel that computes weights from relative geometric measurements between each input neighbor $\mathbf{p}_{k}$ and the output $\mathbf{p}_{o}$.
Each output feature is computed as the average of input neighbor values, weighted by a learned function of their relative geometry measurement $\mathcal{M}_{m}$.
We adopt average reduction instead of summation because different output points may aggregate varying numbers of inputs due to non-uniform sampling.
Averaging ensures consistent scaling and better reflects the contribution of each neighbor, regardless of local density.

As illustrated in the bottom-right of \cref{fig:framework}, formally:
\begin{align}
	x_{o}                       & = \mathcal{K}_{\text{conv}}(\mathcal{X}_{i}, \mathbf{p}_{o})                                                                                                                                    \\
	                            & = \frac{1}{|\mathcal{N}(\mathbf{p}_{o})|}\sum\limits_{k\in\mathcal{N}(\mathbf{p}_o)}x_{k}\prod\limits_{m}f^{(m)}_{\text{weight}}\left( \mathcal{M}_{m}(\mathbf{p}_{k}, \mathbf{p}_{o})\right ), \\
	\mathcal{N}(\mathbf{p}_{o}) & = \{ k: \mathbf{p}_{k}\in \mathcal{P}_{i}, \; d \; (\mathbf{p}_{k}, \mathbf{p}_{o}) \leq r \}. \label{eq: spherical neighbor}
\end{align}

Where $x_{k}$ is the input feature at neighbor $\mathbf{p}_{k}$ and $\mathcal{M}_{m}(\mathbf{p}_{k}, \mathbf{p}_{o})$ denotes a relative measurement between $\mathbf{p}_{k}$ and $\mathbf{p}_{o}$.
In this work, we use two such measurements: (i) the geodesic distance $d(\mathbf{p}_{k}, \mathbf{p}_{o})$ and (ii) the local 1D direction of $\mathbf{p}_{k}$ on the tangent plane centered at $\mathbf{p} _{o}$.
We explicitly decouple these two measurements and assign each its own weighting function $f^{(m)}_{\text{weight}}$, whose outputs are multiplied to produce the final weight.
This factorization allows the kernel to modulate radial and angular sensitivity independently.

The kernel can take different forms depending on the weighting scheme (see \cref{tab: polymorphic kernel} for variations).
The choice of weighting function is up to design, which can be a discrete piecewise-constant (PWC) function, a continuous MLP, or a grid-sampled interpolant as in~\cite{ocampoScalableEquivariantSpherical2023}.

In the case of MLP, the direction is represented as a 1D bearing angle in $[-\pi, \pi]$, the relative azimuth from the tangent-projected north at $\mathbf{p}_o$.
The geodesic distance is embedded with cosines at integer multiples of a support-adapted base frequency $\lfloor \frac{\pi}{r} \rfloor$, which makes the MLP output both even and $2\pi$-periodic.
The bearing angle is embedded with sines and cosines at integer spaced frequencies excluding the raw angle to preserve strict $2\pi$-periodicity and ensure smooth weight transitions as the angle wraps across $-\pi$ and $\pi$.

By construction, if the direction branch is absent, the convolution reduces to a zonal/radial filter that depends only on relative distance, a measurement invariant to rotation, thus making the kernel trivially rotation-equivariant.\footnote{Proofs in~\cite{estevesLearningSO3Equivariant2020} and the supplementary material} However, once a directional component is introduced, the kernel becomes gauge-dependent, i.e., it relies on a locally defined \emph{up vector} in the tangent plane that rotates with the signal. As a result, the kernel's response varies under global rotation, breaking rotation-equivariance.

\subsubsection{Spherical Pooling}
\label{sssec:spherical-pooling}
Spherical pooling is defined analogously over the same geodesic neighborhood as \cref{eq: spherical neighbor}.

As illustrated in the bottom-right of \cref{fig:framework}, formally:
\begin{align}
	x_{o} & = \mathcal{K}_{\text{pool}}(\mathcal{X}_{i}, \mathbf{p}_{o}) = f_{\text{pool}}\left( x_{k}: k \in \mathcal{N}(\mathbf{p}_{o}) \right).
\end{align}
$f_{\text{pool}}$ can be a simple reducer like min, max, or average, or more complex local statistics such as the mean of the upper quartile.

The output sample locations for spherical convolution or pooling layers are chosen via a location sampler with a resolution factor that controls the density of output points for upsampling or downsampling.
Since these coordinates are fixed per layer, all geometric measurements between input and output can be cached and reused efficiently after the first forward pass, while preserving the dynamic configurability of the spherical layers.

\Cref{fig:spherical CNN & spherical Pool} visualizes spherical convolution and pooling on a sample input spherical image captured and projected from a fisheye camera.

\input{fig/Convolution_And_Pooling}

Putting all components together, the whole pipeline is shown in \cref{fig:framework}.

%% file: fig/Sampling.tex
\begin{figure}[!t]
	\centering

	\begin{subfigure}
		[t]{0.24\linewidth}
		\centering
		\includegraphics[width=\linewidth]{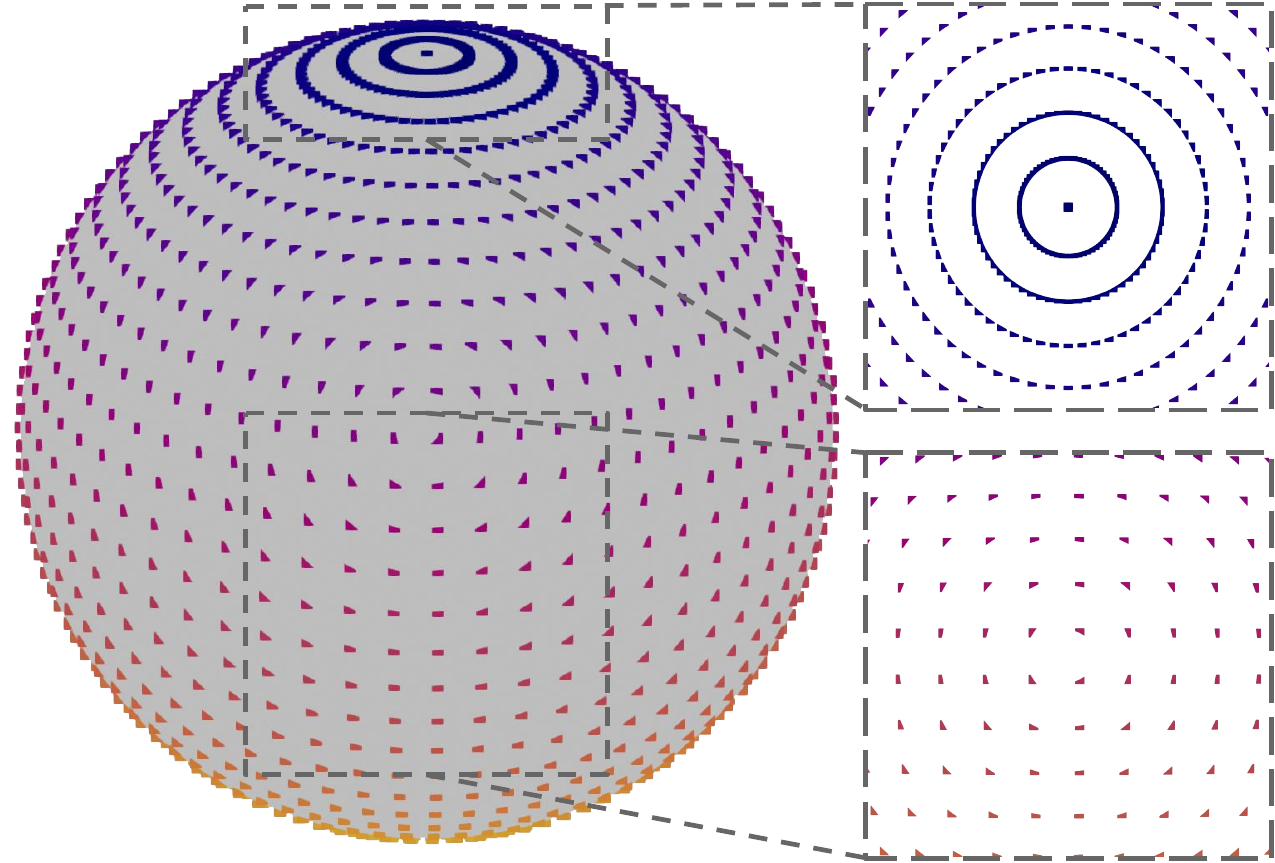}
		\caption{Equirect}
	\end{subfigure}
	\begin{subfigure}
		[t]{0.24\linewidth}
		\centering
		\includegraphics[width=\linewidth]{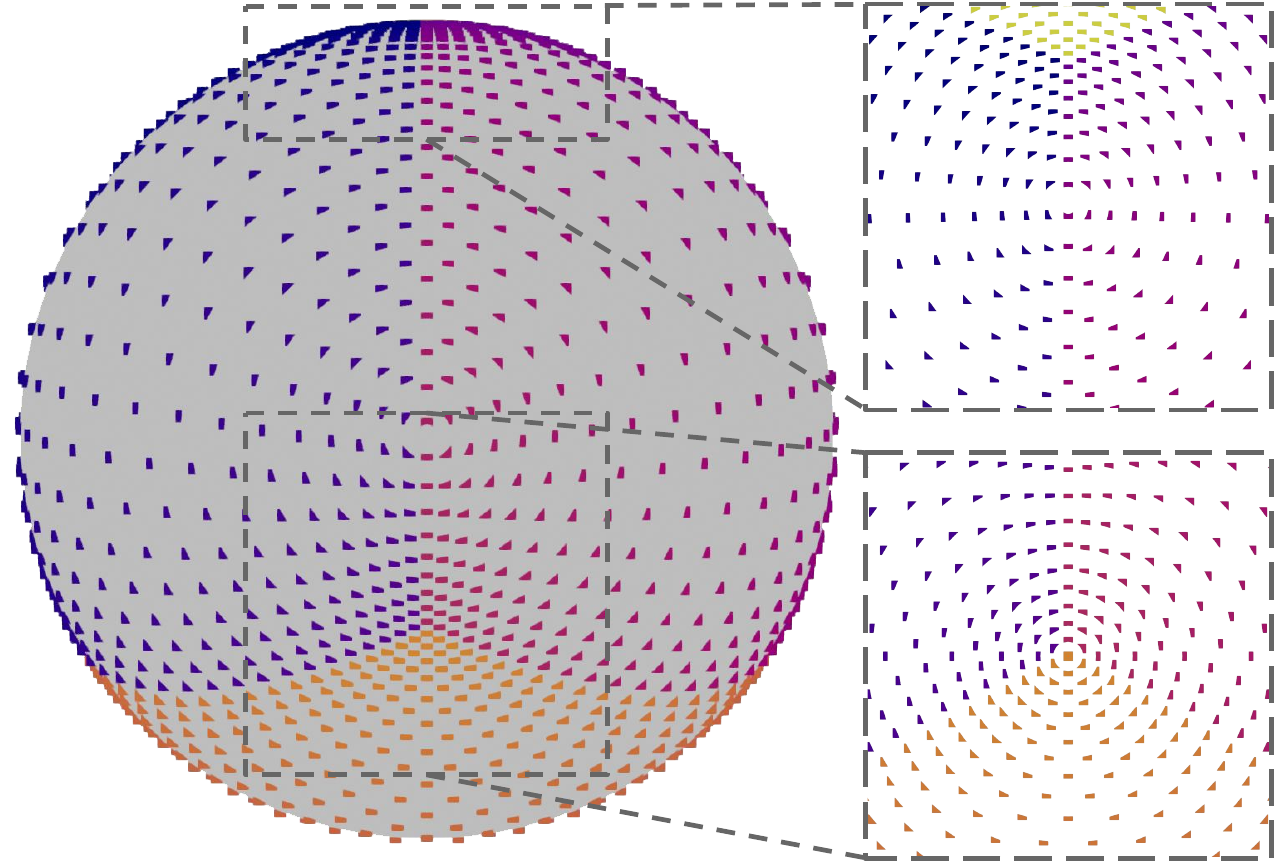}
		\caption{Tetrahedron}
	\end{subfigure}
	\begin{subfigure}
		[t]{0.24\linewidth}
		\centering
		\includegraphics[width=\linewidth]{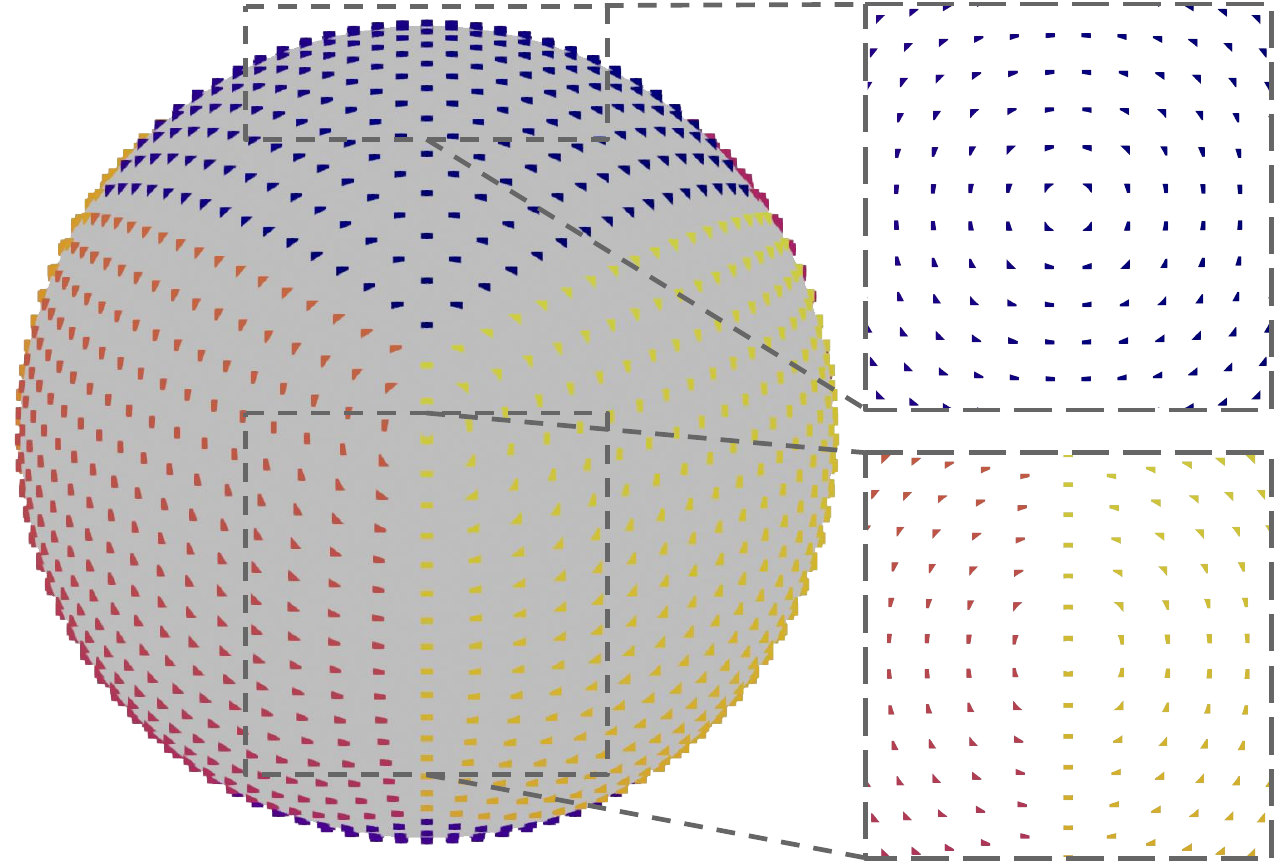}
		\caption{Hexahedron}
	\end{subfigure}
	\begin{subfigure}
		[t]{0.24\linewidth}
		\centering
		\includegraphics[width=\linewidth]{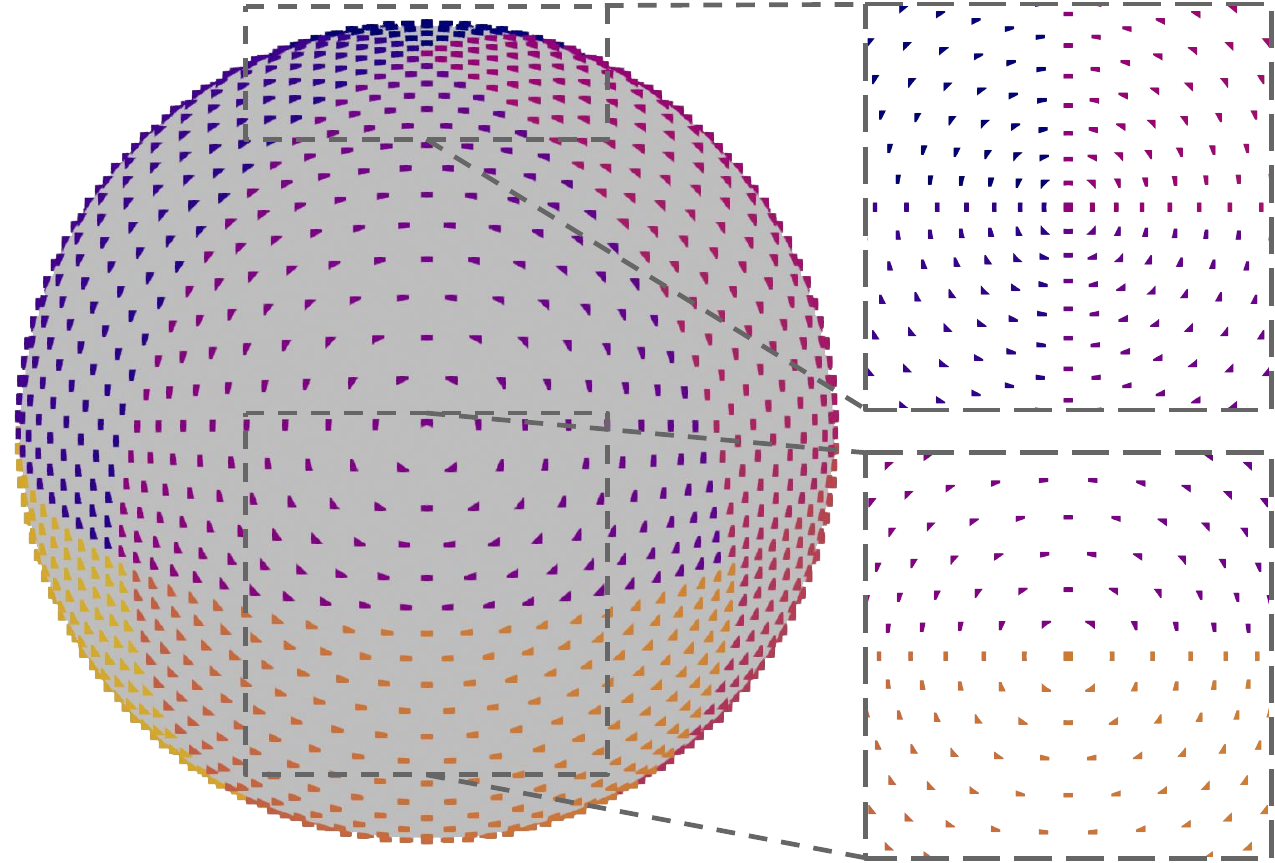}
		\caption{Octahedron}
	\end{subfigure}
	\\
	\begin{subfigure}
		[t]{0.24\linewidth}
		\centering
		\includegraphics[width=\linewidth]{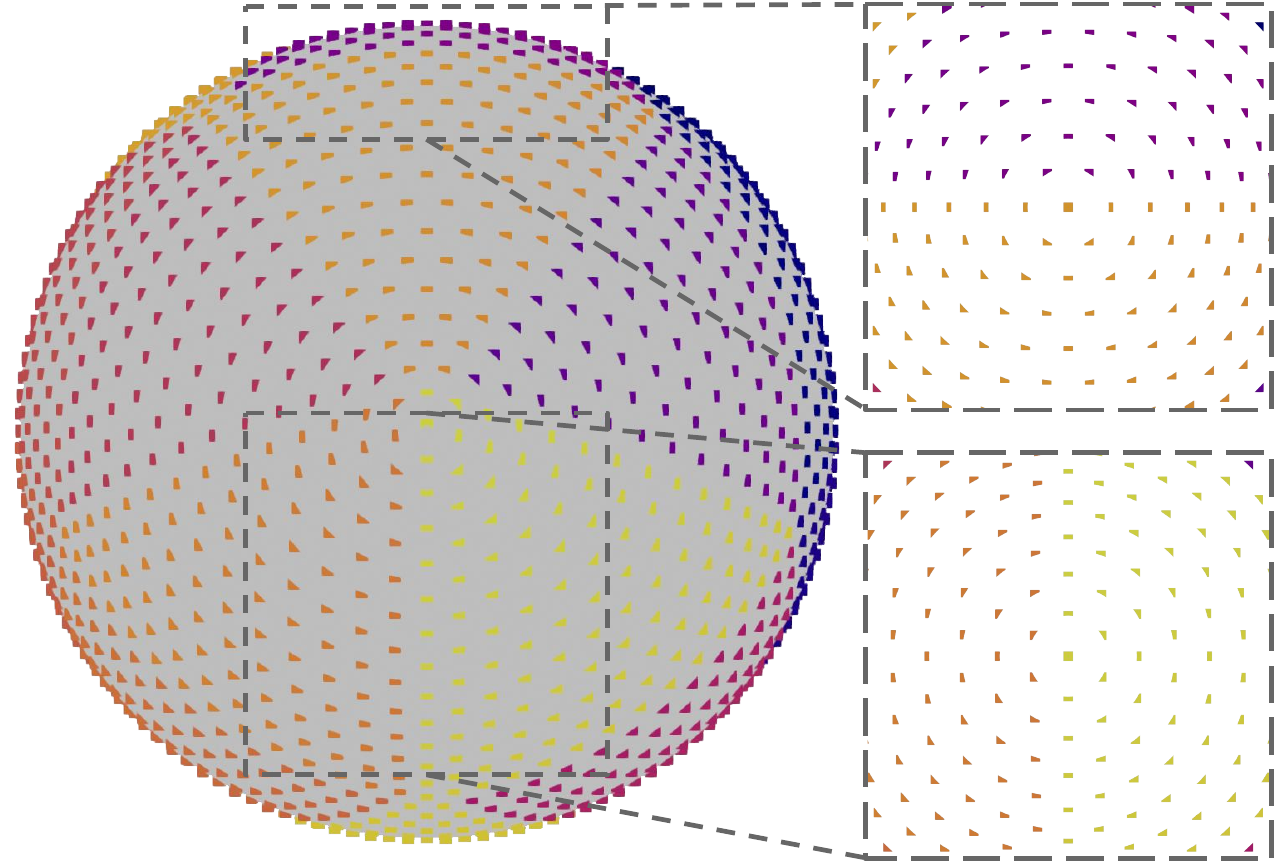}
		\caption{Icosahedron}
	\end{subfigure}
	\begin{subfigure}
		[t]{0.24\linewidth}
		\centering
		\includegraphics[width=\textwidth]{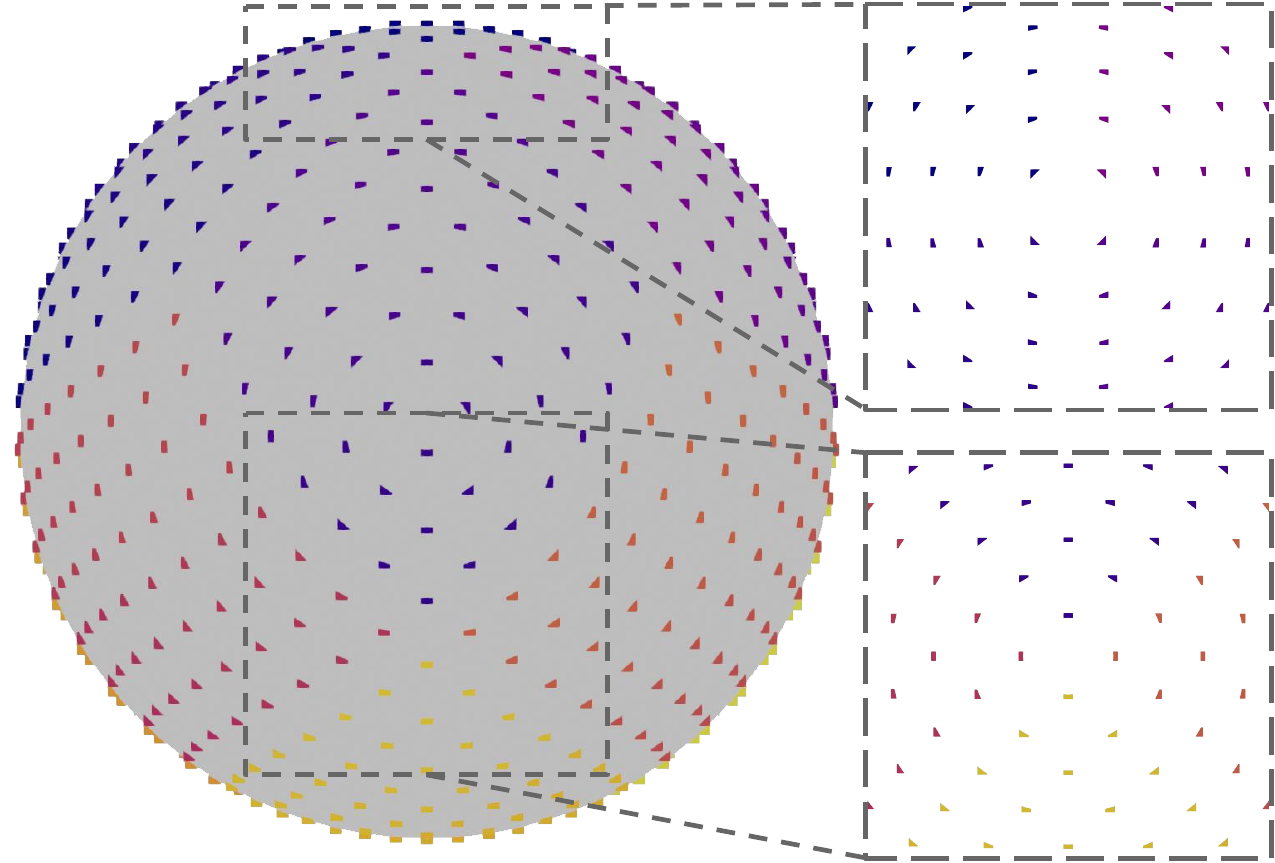}
		\caption{HEALPix}
	\end{subfigure}
	\begin{subfigure}
		[t]{0.24\linewidth}
		\centering
		\includegraphics[width=\textwidth]{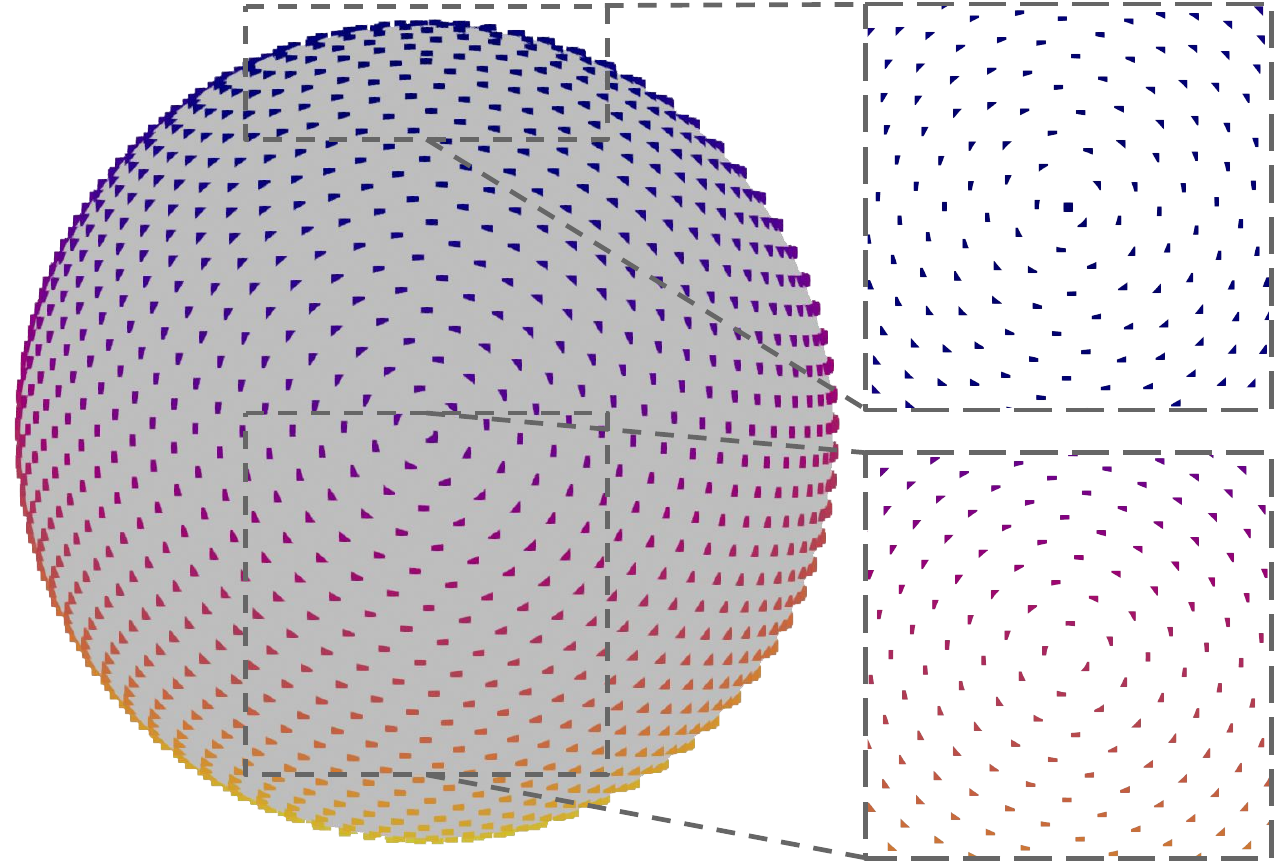}
		\caption{Fibonacci}
	\end{subfigure}
	\begin{subfigure}
		[t]{0.24\linewidth}
		\centering
		\includegraphics[width=\textwidth]{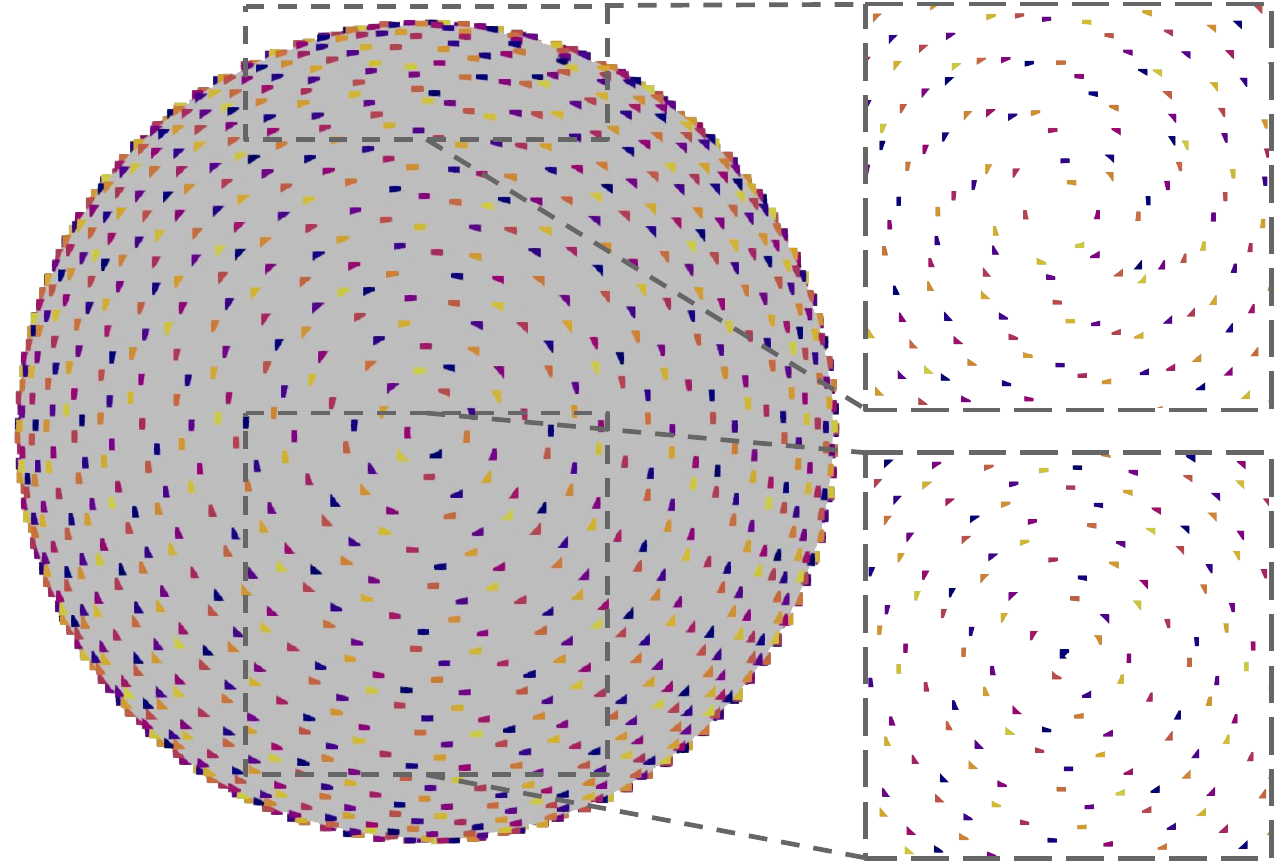}
		\caption{Quasi-random}
	\end{subfigure}

	\caption{\textbf{Spherical Sampling Methods.}
		Various location sampling strategies produce different levels of uniformity across the sphere.
		The bottom row displays point distributions with higher uniformity compared to coarser Goldberg polyhedron discretizations. }
	\label{fig:sampling}
\end{figure}

%% file: tab/Generic_Spherical_CNN.tex
\begin{table}[!t]
	\centering
	\setlength{\tabcolsep}{2pt}
	\renewcommand{\arraystretch}{0.95}
	\resizebox{0.99\linewidth}{!}{
		\begin{tabular}{@{} c | c c c @{}}
			\diagbox[width=6.5em,outerleftsep=1pt]{\textbf{Distance}} {\textbf{Direction}}                    & \textbf{Continuous}                                                              & \textbf{Discrete ($\times$6)}                                                    & \textbf{None}                                                                   \\
			\noalign{\hrule height 0.5pt} \makebox[5em][c]{\rotatebox{90}{\hspace{0.5em}\textbf{Continuous}}} & \rotatebox[origin=c]{90}{\includegraphics[width=2.5cm]{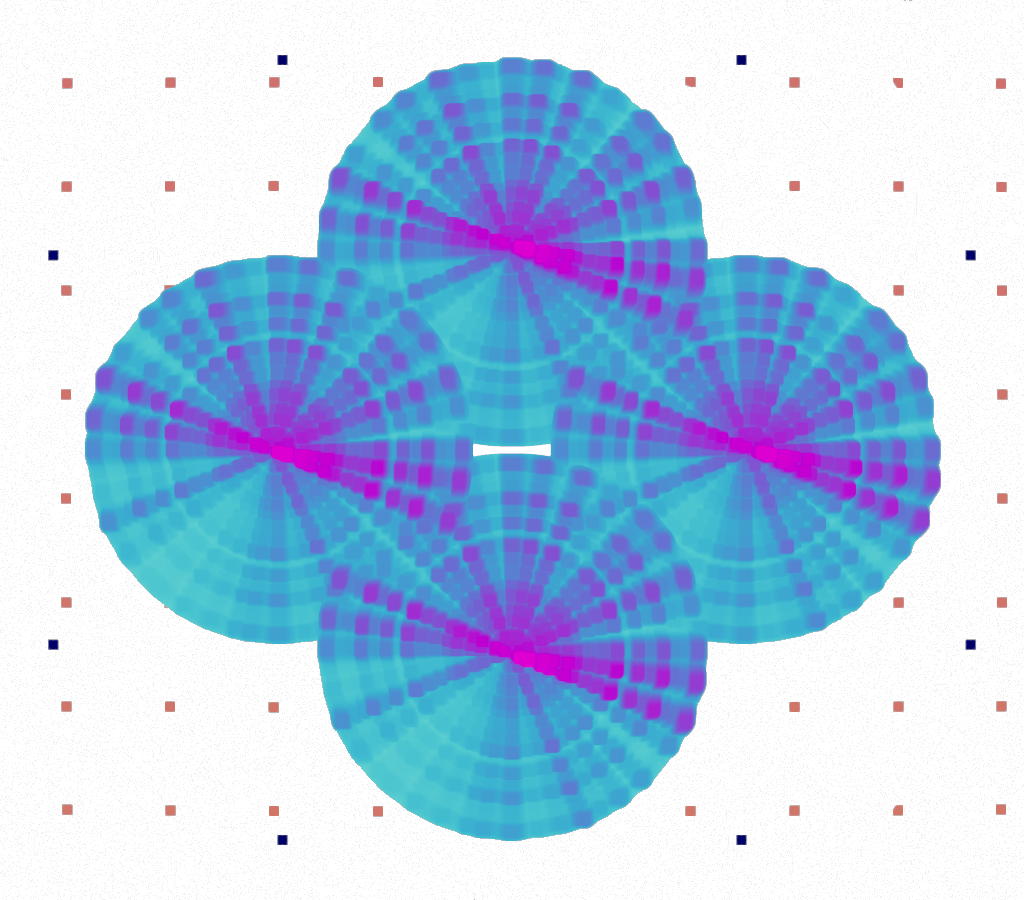}} & \rotatebox[origin=c]{90}{\includegraphics[width=2.5cm]{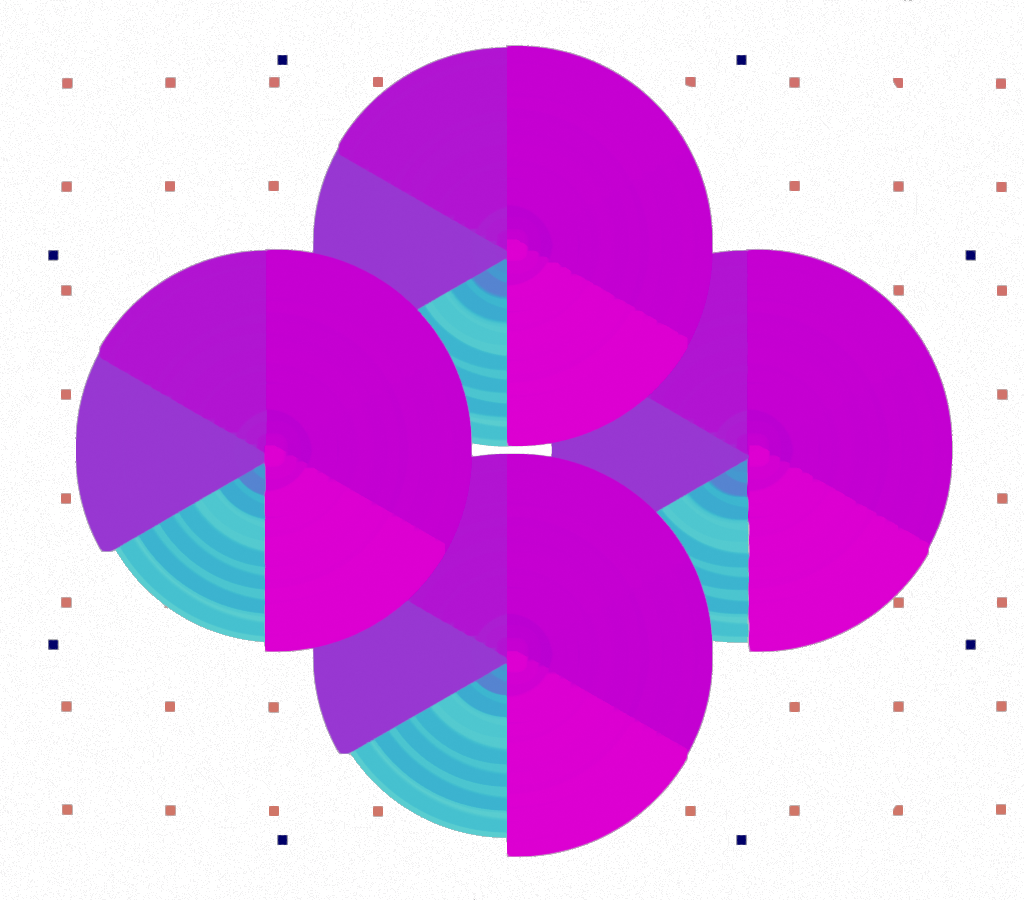}} & \rotatebox[origin=c]{90}{\includegraphics[width=2.5cm]{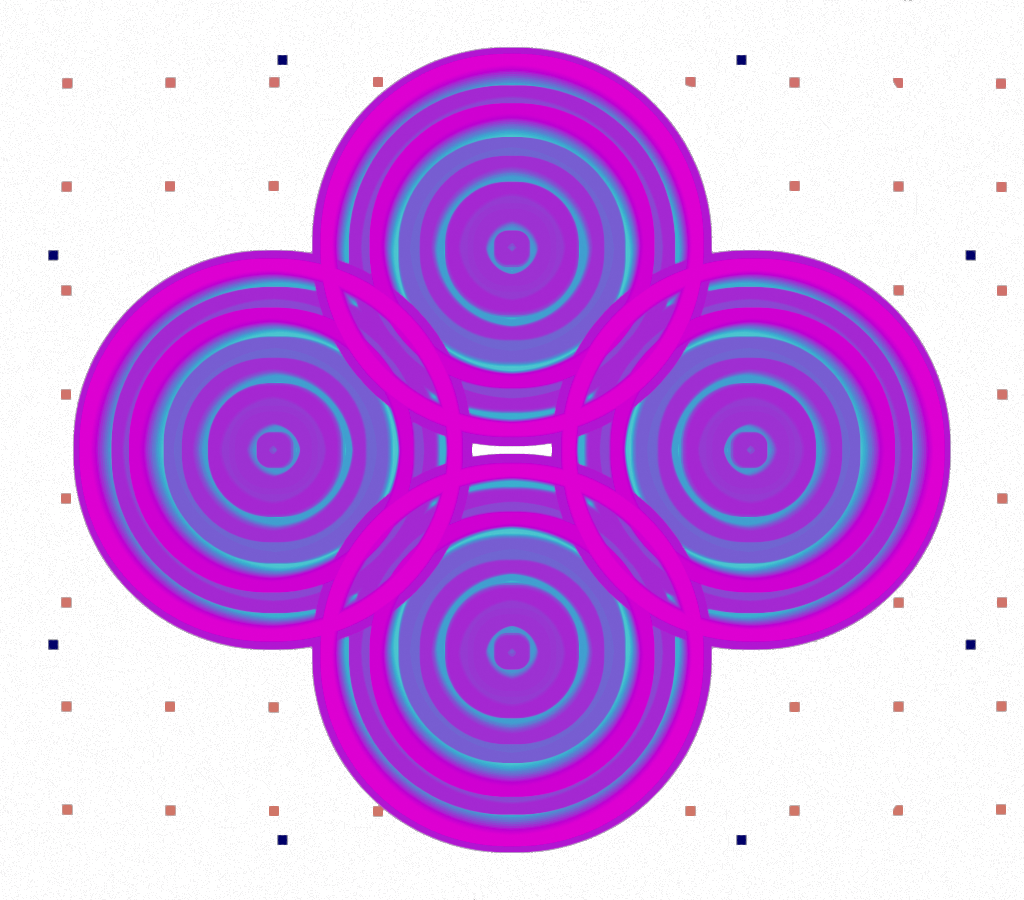}} \\
			\makebox[5em][c]{\rotatebox{90}{\hspace{0.25em}\textbf{Discrete ($\times$3)}}}                    & \rotatebox[origin=c]{90}{\includegraphics[width=2.5cm]{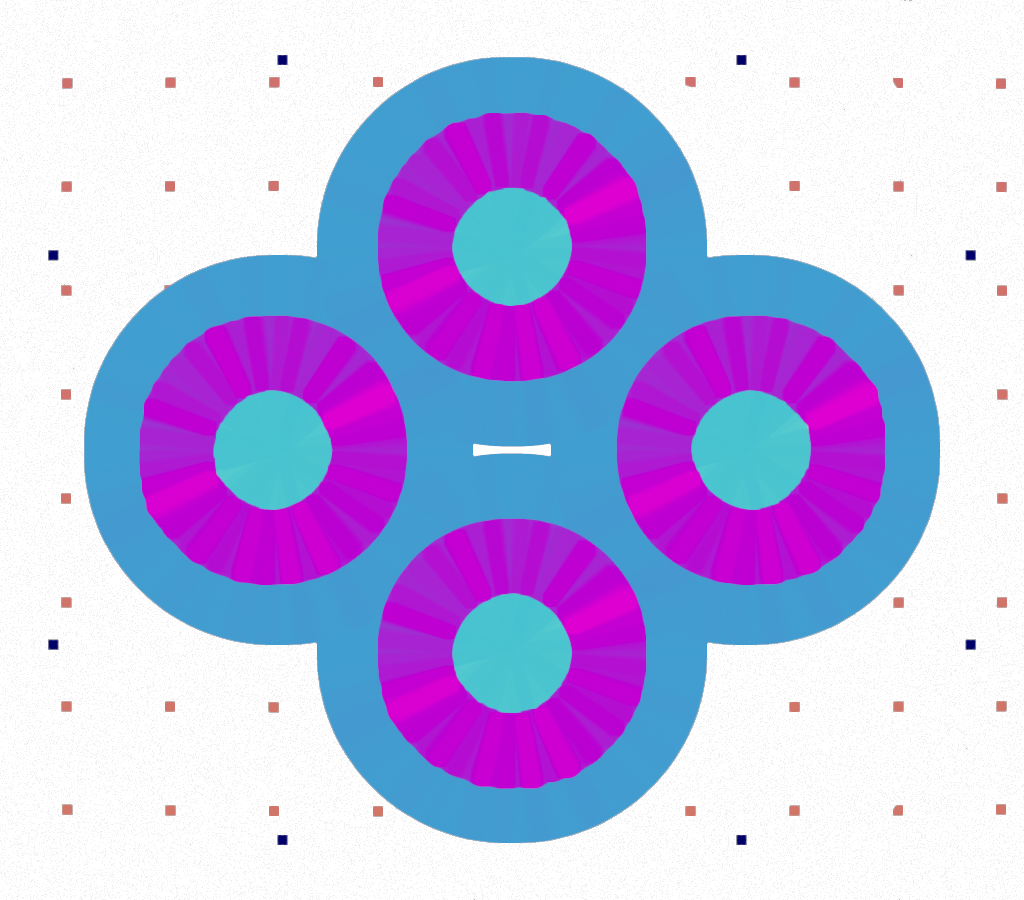}} & \rotatebox[origin=c]{90}{\includegraphics[width=2.5cm]{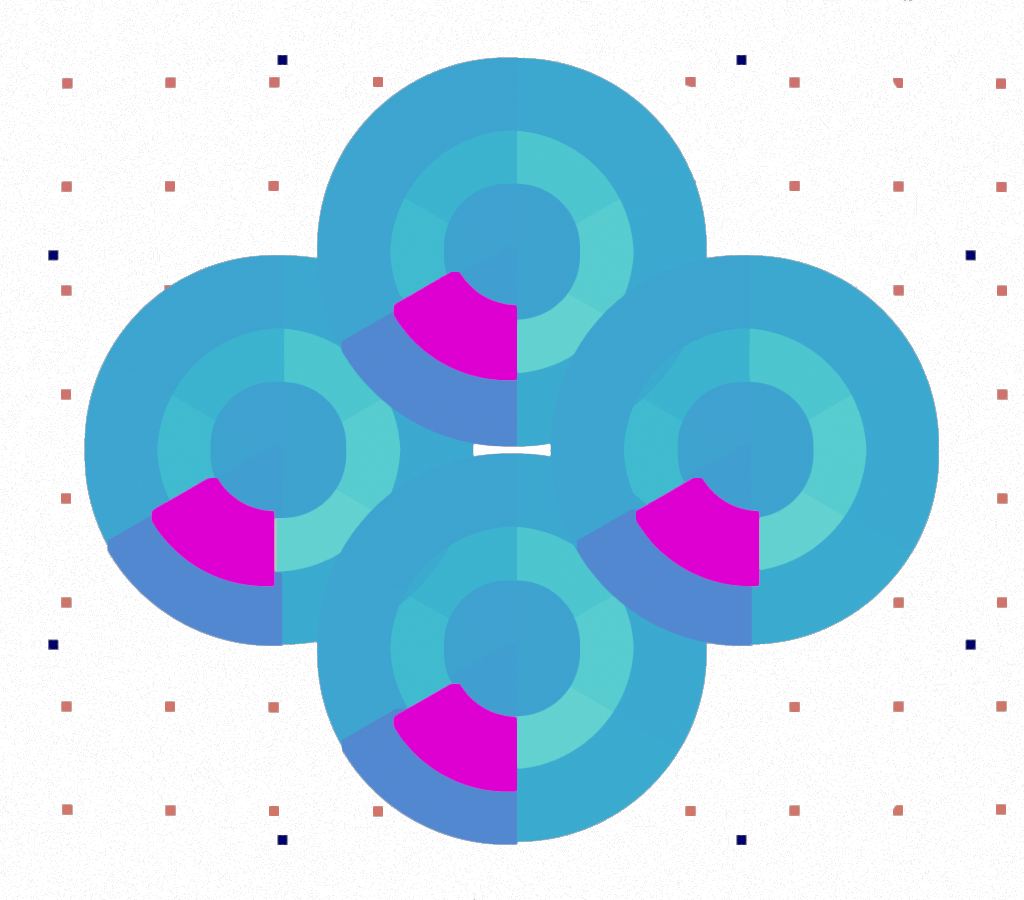}} & \rotatebox[origin=c]{90}{\includegraphics[width=2.5cm]{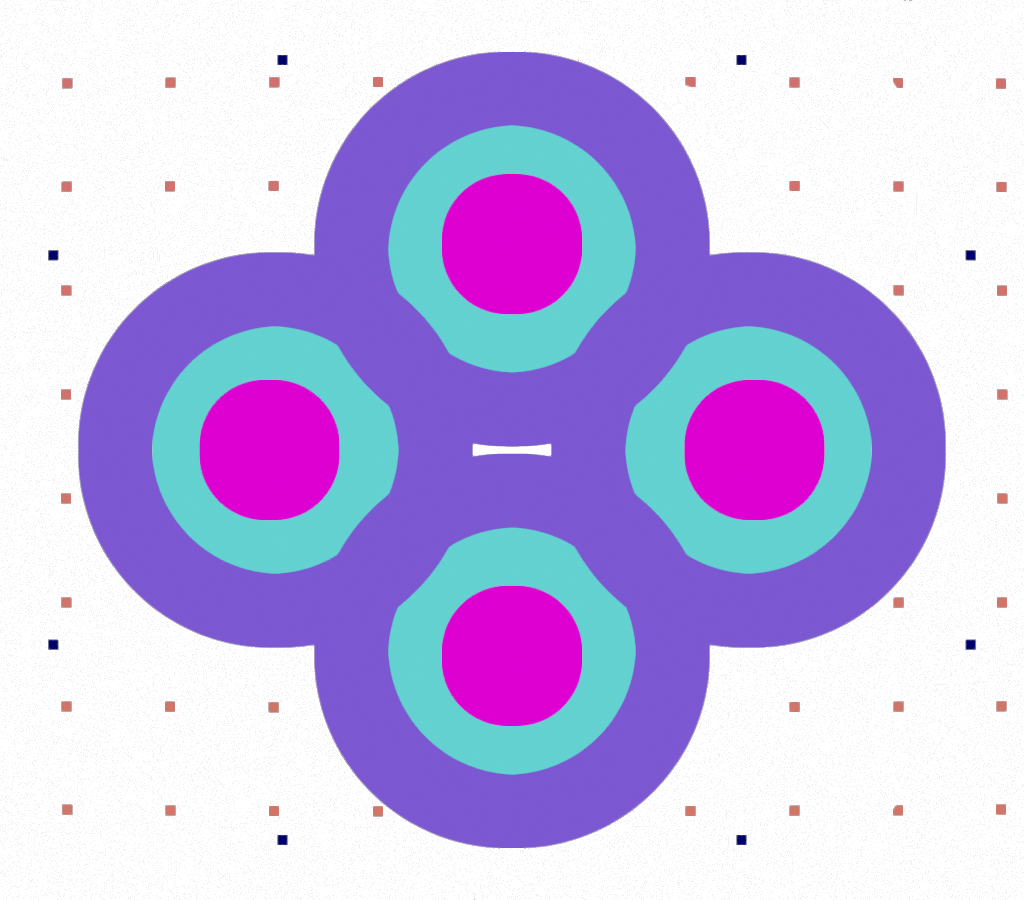}} \\
		\end{tabular}
	}
	\caption{\textbf{Generic Spherical CNN.}
		\textcolor{brown}{Brown} and \textcolor{blue}{blue} dots denote \textcolor{brown}{input} and \textcolor{blue}{output} locations.
		Colors visualize the activation weights between a given input-output channel pair.}
	\label{tab: polymorphic kernel}
\end{table}

%% file: fig/Convolution_And_Pooling.tex
\begin{figure}[!t]
	\centering
	\begin{subfigure}
		[t]{0.32\linewidth}
		\centering
		\includegraphics[width=\linewidth]{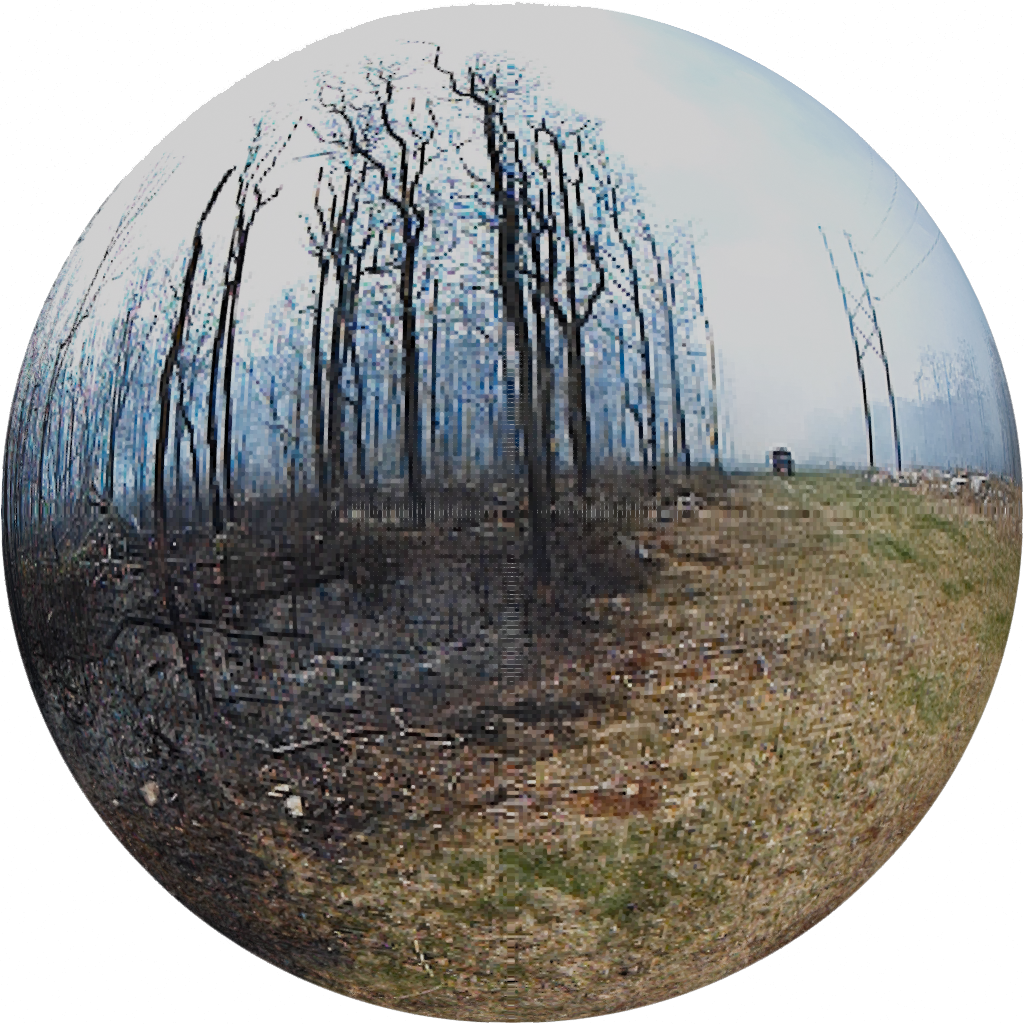}
		\caption{Spherical Image}
	\end{subfigure}
	\hfill
	\begin{subfigure}
		[t]{0.32\linewidth}
		\centering
		\includegraphics[width=\linewidth]{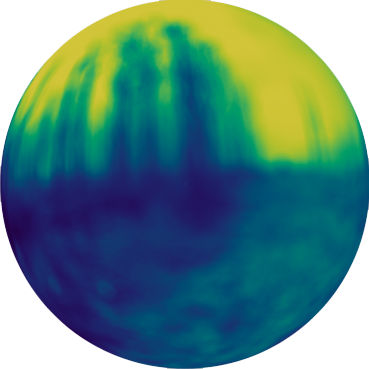}
		\caption{Convolution}
	\end{subfigure}
	\hfill
	\begin{subfigure}
		[t]{0.32\linewidth}
		\centering
		\includegraphics[width=\linewidth]{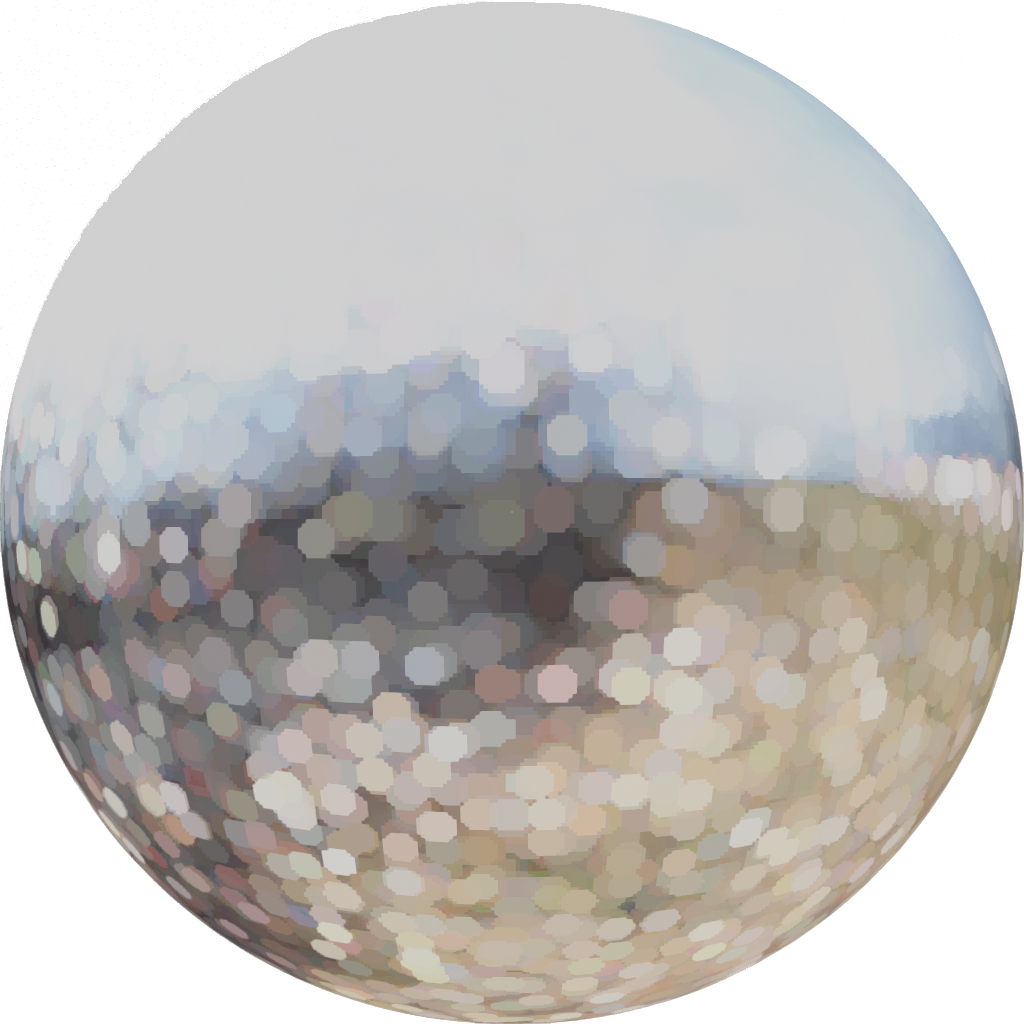}
		\caption{Max Pooling}
	\end{subfigure}
	\caption{\textbf{Spherical Convolution and Pooling.}
		The output locations are set to be identical to the input locations.
		(b) visualizes a channel of convolution output with weight = 1 and bias = 0, effectively a summation operator.}
	\label{fig:spherical CNN & spherical Pool}
\end{figure}

%% file: sec/4_experiment.tex
\section{Experiments}
\label{sec:experiments}
In this section, we demonstrate the effectiveness and versatility of the proposed \textit{Unified Spherical Frontend} on several representative vision tasks, instantiating our generic pipeline for each.
In each case, we replace standard planar layers with our spherical layers to construct the spherical variant, while keeping other aspects of the models and training protocol the same for a fair comparison.
All benchmarks are performed in the planar domain to ensure extra consistency.
These tasks demonstrate that our approach matches the accuracy of conventional models while preserving robustness under arbitrary image rotations and varying fields of view.

Prior spherical CNNs operate in the harmonics domain and incur prohibitively high computational and memory costs that scale with spatial resolution, we therefore compare with such methods only in the low-resolution MNIST experiment.

In the following tables, \emph{NR} and \emph{RR} denote training or testing under \emph{non-rotated} and \emph{randomly rotated} settings.\footnote{Training details and additional results in the supplementary material}


\subsection{MNIST Classification}
\label{ssec:mnist-classification}
\subsubsection{Experiment Setup}
\label{sssec:mnist-classification-experiment-setup}
We first evaluate our pipeline on the Spherical MNIST benchmark by stereographically projecting MNIST digits onto the sphere following~\cite{cohenSphericalCNNs2018}.
We apply global average pooling over feature values before fully-connected layers to ensure rotation-invariance.
Training is performed on 6,000 spherical digits for 100 epochs (batch size 1024) without augmentation.
To avoid projection distortion for the planar model, random rotations are applied only around the central axis.

In this experiment, we also ablate 4 weighting function parameterizations: (1) Radial Discrete: PWC radial function with 3 segments on distance; (2) Radial Continuous: MLP with [8, 8] hidden channels on unembedded distance; (3) High-Frequency Radial Continuous: similar to (2) but embed distance with 6 fourier levels; and (4) Distance $\times$ Direction: MLP with [16, 16] hidden channels on both distance and direction embedded with 8 fourier levels.
Variants (1) and (4) correspond to the bottom-right and top-left cells of \cref{tab: polymorphic kernel}.

\subsubsection{Results and Discussion}
\label{sssec:mnist-classification-results-and-discussion}
\input{tab/MNIST}
As shown in \cref{tab: mnist}, spherical CNNs with radial kernel maintain strong performance under rotation, while planar CNNs degrade sharply when test digits are randomly rotated.

Radial-only variants offer built-in rotation-equivariance and perform best under random rotations.
In contrast, the distance $\times$ direction kernel captures orientation-sensitive cues (e.g., distinguishing “6” vs. “9”) and matches the planar CNN on upright digits, but sacrifices equivariance.

Finally, expressivity depends heavily on kernel parameterization: a low-frequency MLP underperforms a simple PWC design, highlighting the importance of appropriate embeddings.


\subsection{Object Detection}
\label{ssec:object-detection}
\subsubsection{Experiment Setup}
\label{sssec:object-detection-experiment-setup}
Next, we evaluate our pipeline on object detection in 360$^{\circ}$ panoramic images on the PANDORA dataset~\cite{xuPANDORAPanoramicDetection2022}.
In this experiment, we focus on demonstrating rotation robustness purely from architectural inductive bias.
PANDORA contains 3,000 panorama images of resolution $1920 \times 960$, annotated with 94,353 oriented bounding boxes of 47 categories.
Each box is represented as Rotated Bounding Field-of-View (RBFoV) with 6 parameters: $(\theta, \phi, \alpha, \beta, \gamma, \text{category})$, as articulated in~\cite{xuPANDORAPanoramicDetection2022}.
Raw images are downsampled to $960 \times 480$ before training.
We adapt the recent YOLOv11~\cite{khanamYOLOv11OverviewKey2024} as the backbone and R-CenterNet~\cite{xuPANDORAPanoramicDetection2022}'s detection head for both planar and spherical models.
Detection performance is evaluated with mean Average Precision (mAP) at Intersection over Union (IoU) thresholds of 10\% and 50\%.\footnote{Additional implementation details in the supplementary material}

All spherical convolutions adopt the 3-segment discrete PWC weighting function on geodesic distance, identical to scheme (1) from the MNIST experiments.

To apply random rotation on a spherical image, we rotate the spherical image's vectors and then resample back to the canonical unrotated vectors.
This ensures that the pixel values reflect a globally rotated view, while maintaining consistent geometry across batches and samples.

\subsubsection{Results and Discussion}
\label{sssec:object-detection-results-and-discussion}
\input{tab/Object_Detection_Quantitative}
\Cref{tab: object detection quantitative} shows that our spherical pipeline offers improved robustness to random rotations compared to the planar baseline.
Notably, the planar YOLOv11 model achieves strong performance when trained and tested without rotation, even surpassing the original R-CenterNet~\cite{xuPANDORAPanoramicDetection2022} due to a more powerful backbone.
However, its performance collapses when evaluated under rotations unless trained with explicit rotation augmentation.

In contrast, our spherical model demonstrates rotation-equivariance without augmentation, maintaining stable performance across rotation conditions.
This robustness aligns with the MNIST results in \cref{tab: mnist}, where distance-only spherical kernels preserved rotation consistency, though with some reduction in raw accuracy due to limited expressiveness.
The same trade-off recurs here: using discrete radial weights ensures equivariance but restricts directional sensitivity, which is often important for capturing orientation-specific patterns.

It is also worth noting that certain prediction targets, such as angular offsets or bounding box orientation, are inherently gauge-dependent and cannot be preserved under global rotation simply by a rotation-equivariant model.
Capturing such directional cues may require either directional kernels with explicit data-driven learning through augmentation, or gauge-equivariant architectures that estimate local frame direction.


\input{tab/Semantic_Segmentation_Random_Rotation_Quantitative}

\subsection{Semantic Segmentation}
\label{ssec:semantic-segmentation}
\subsubsection{Experiment Setup}
\label{sssec:semantic-segmentation-experiment-setup}
Finally, we evaluate our framework on semantic segmentation and assess its ability to generalize across different camera lenses on Stanford 2D-3D-S~\cite{armeniJoint2D3DSemanticData2017}, which contains 1,413 equirectangular RGB-D panoramas of resolution $4096 \times 2048$ with per-pixel semantic labels for 13 classes.
To isolate the effect of geometry-aware processing, we use only RGB inputs, omitting available depth data.
Input image downsampling and random rotation are performed similarly to object detection.

We experiment with three different backbones to show the plug-and-play nature of our spherical pipeline: DeepLab v3~\cite{chenRethinkingAtrousConvolution2017}, UNet~\cite{ronnebergerUNetConvolutionalNetworks2015}, and YOLOv11~\cite{khanamYOLOv11OverviewKey2024}.
All spherical models adopt the same distance-only 3-segment PWC kernel.

To simulate images captured by other lens types, we generate lens normal maps for two other camera models: (1) a 90$^{\circ}$ horizontal and vertical FoV pinhole camera with $280 \times 280$ resolution, and (2) a 180$^{\circ}$ FoV fisheye camera with $560 \times 560$ resolution (yielding a valid pixel ratio of $\frac{\pi}{4}$) The resolutions are chosen to ensure that the number of pixels is proportional to the FoV coverage area on the sphere.
To ensure broad spatial coverage from the original equirectangular image, each lens normal map is randomly oriented toward one of the six cube face directions prior to value interpolation.

To evaluate cross-lens generalization, we perform a single-batch overfitting experiment without applying random rotation.
Specifically, we overfit each model on a batch of samples from one lens type and evaluate its zero-shot performance on the same batch resampled into other lens types.
This setting explicitly isolates cross-lens adaptability from conventional train-to-test generalization.\footnote{Full-scale dataset experiment in the supplementary material}

\subsubsection{Results and Discussion}
\label{sssec:semantic-segmentation-results-and-discussion}
\input{tab/Single_Batch_Overfit_Test}
The results in \cref{tab: semantic segmentation random rotation quantitative} confirm that our spherical models exhibit significantly greater robustness to random rotations compared to their planar counterparts.
This pattern mirrors the results in MNIST and object detection, where radial-only spherical kernels preserved rotation performance but incurred a small drop in peak accuracy.

The equivariance diagram in \cref{fig:equivariance & invariance diagram} provides more than just intuition.
It is a visualization of a logit channel from a spherical segmentation model \emph{trained without random rotation}.
This demonstrates the model's true equivariant behavior under global $\mathrm{SO}(3)$ transformations, providing compelling visual evidence of the theory in practice.

As shown in \cref{tab: single batch overfit test}, planar models show clear performance drops when evaluated on different lenses from training, especially in off-diagonal entries.
Spherical models perform more consistently across lenses, especially when the source and target lens share similar FoV coverage.
Degradation is more noticeable when moving between views with drastically different FoV, such as from pinhole to panoramic, due to mismatched pixel counts.
While neither model achieves perfect performance across all lens combinations, the spherical model limits the drop to fewer and less severe cases.
Finally, since the 2D-3D-S dataset excludes polar regions from evaluation and no rotation is applied here, a setting that favors planar models.
Despite this, the spherical model still shows better generalization across lenses.

\subsection{Ablation Study}
\label{ssec:ablation-study}
We ablate two critical design factors: location samplers, which affect rotation-equivariance through spatial uniformity, and the number of distance segments in PWC function, which controls kernel expressivity.

\input{tab/Ablation}

\Cref{tab: ablation} shows that icosahedron with 3-segment discretization consistently delivers the best trade-off between accuracy and rotational stability.
Using more bins may lead to overfitting, as each segment contains fewer samples.
Non-uniform sampling schemes, on the other hand, introduce spatial bias that degrades rotation-equivariance, reinforcing the importance of uniformity in spherical sampling.

%% file: tab/MNIST.tex
\begin{table}[!t]
	\centering
	\renewcommand{\arraystretch}{1.05}
	\setlength{\tabcolsep}{2pt}
	\resizebox{0.99\linewidth}{!}{
		\begin{tabular}{@{} >{\raggedright\arraybackslash}p{6.5cm} | c c@{}}
			\toprule \multicolumn{1}{c|}{\textbf{Model}}                      & \textbf{NR $\uparrow$}       & \textbf{RR $\uparrow$}       \\
			\midrule
			\hspace{1.2em} Planar                                             & 98.45\%                      & \cellcolor{lightred} 41.08\% \\
			\hspace{1.2em} S$^2$CNN~\cite{cohenSphericalCNNs2018}             & 96\%                         & 94\%                         \\
			\hspace{1.2em} SO(3) CNN~\cite{estevesLearningSO3Equivariant2020} & 98.7\%                       & 98.1\%                       \\
			\hline
			(1) Spherical Dis PWC $\times$3                                   & 87.18\%                      & 85.43\%                      \\
			(2) Spherical Dis MLP [8, 8], $L=0$                               & \cellcolor{lightred} 67.01\% & \cellcolor{lightred} 65.74\% \\
			(3) Spherical Dis MLP [8, 8], $L=6$                               & 92.13\%                      & 91.50\%                      \\
			(4) Spherical Dis$\times$Dir MLP [16, 16], $L=8$                  & 98.28\%                      & \cellcolor{lightred} 43.54\% \\
			\bottomrule
		\end{tabular}
	}
	\caption{\textbf{MNIST Classification Results.}
		All models are trained without random rotation.
		$L$ denotes embedding levels.}
	\label{tab: mnist}
\end{table}

%% file: tab/Object_Detection_Quantitative.tex
\begin{table}[!t]
	\centering
	\scriptsize
	\setlength{\tabcolsep}{2pt}
	\renewcommand{\arraystretch}{1.05}
	\resizebox{0.99\linewidth}{!}{
		\begin{tabular}{@{}c c c c c c@{}}
			\toprule \multirowcell{2}{\diagbox[outerleftsep=-18pt,outerrightsep=-50pt,width=7.5em]{\textbf{Train}}{\textbf{Test}}}                                                                                                                                                            & \multirow{2}{*}{}        & \multicolumn{2}{c}{NR} & \multicolumn{2}{c}{RR}                                                            \\
			\cmidrule(lr){3-4} \cmidrule(lr){5-6}                                                                                                                                                                                                                                             &                                                                                                                                       
			                                                                                                                                                                                                                                                                                  & mAP$_{10}\uparrow$       & mAP$_{50}\uparrow$     & mAP$_{10}\uparrow$     & mAP$_{50}\uparrow$                                       \\
			\midrule R-CenterNet\cite{xuPANDORAPanoramicDetection2022}                                                                                                                                                                                                                        & \multicolumn{1}{|c|}{NR} & 35.73\%                & 22.7\%                 & N/A                         & N/A                        \\
			\midrule \multirow{2}{*}{\makecell[c]{Planar                                                                                                                                                                                       \\YOLOv11\cite{khanamYOLOv11OverviewKey2024}}} & \multicolumn{1}{|c|}{NR} & 39.65\%                & 24.41\%                & \cellcolor{lightred}12.71\% & \cellcolor{lightred}4.66\% \\
			                                                                                                                                                                                                                                                                                  & \multicolumn{1}{|c|}{RR} & 27.76\%                & 9.99\%                 & 28.01\%                     & 10.24\%                    \\
			\midrule                                                                                                                
			\makecell[c]{Spherical                                                                                                                                                                                                             \\YOLOv11}                                     & \multicolumn{1}{|c|}{NR} & 29.54\%                & 11.41\%                & 29.59\%                     & 7.90\%                     \\
			\bottomrule
		\end{tabular}
	}
	\caption{\textbf{Object Detection Results on PANDORA Dataset.}}
	\label{tab: object detection quantitative}
\end{table}

%% file: tab/Semantic_Segmentation_Random_Rotation_Quantitative.tex
\begin{table}[!t]
	\centering
	\scriptsize
	\setlength{\tabcolsep}{2pt}
	\renewcommand{\arraystretch}{1.05}
	\resizebox{0.99\linewidth}{!}{
		\begin{tabular}{cc cc cc}
			\toprule \toprule                                                                                                      
			\multicolumn{2}{c}{\multirowcell{2}{\diagbox[outerleftsep=-20pt,outerrightsep=-40pt]{\textbf{Train}}{\textbf{Test}}}}                                                                                                                                                            & \multicolumn{2}{c}{NR}   & \multicolumn{2}{c}{RR}                                                                                 \\
			\cmidrule(lr){3-4} \cmidrule(lr){5-6}                                                                                                                                                                                                                                            &                                                                                                                                   
			                                                                                                                                                                                                                                                                                 & mIoU $\uparrow$          & mAcc $\uparrow$        & mIoU $\uparrow$ & mAcc $\uparrow$                                             \\
			\midrule \midrule \multirow{2}{*}{\makecell{Planar                                                                                                                                                                      \\DeepLab v3\cite{chenRethinkingAtrousConvolution2017}}} & \multicolumn{1}{|c|}{NR} & 35.01\%                & 58.30\%         & \cellcolor{lightred} 12.11\% & \cellcolor{lightred} 22.50\% \\
			                                                                                                                                                                                                                                                                                 & \multicolumn{1}{|c|}{RR} & 32.29\%                & 52.89\%         & 38.30\%                      & 53.99\%                      \\
			\midrule \multirow{2}{*}{\makecell{Planar                                                                                                                                                                               \\UNet\cite{ronnebergerUNetConvolutionalNetworks2015}}}  & \multicolumn{1}{|c|}{NR} & 33.33\%                & 55.48\%         & \cellcolor{lightred} 12.91\% & \cellcolor{lightred} 23.40\% \\
			                                                                                                                                                                                                                                                                                 & \multicolumn{1}{|c|}{RR} & 33.75\%                & 51.13\%         & 35.91\%                      & 51.52\%                      \\
			\midrule \multirow{2}{*}{\makecell{Planar                                                                                                                                                                               \\YOLOv11\cite{khanamYOLOv11OverviewKey2024}}}           & \multicolumn{1}{|c|}{NR} & 28.32\%                & 48.09\%         & \cellcolor{lightred} 8.17\%  & \cellcolor{lightred} 16.43\% \\
			                                                                                                                                                                                                                                                                                 & \multicolumn{1}{|c|}{RR} & 28.53\%                & 44.39\%         & 30.62\%                      & 45.13\%                      \\
			\midrule \midrule \multirow{2}{*}{\makecell{Spherical                                                                                                                                                                   \\DeepLab v3}}                                           & \multicolumn{1}{|c|}{NR} & 28.78\%                & 45.27\%         & 28.09\%                      & 41.18\%                      \\
			                                                                                                                                                                                                                                                                                 & \multicolumn{1}{|c|}{RR} & 30.55\%                & 44.58\%         & 32.59\%                      & 45.38\%                      \\
			\midrule \multirow{2}{*}{\makecell{Spherical                                                                                                                                                                            \\UNet}}                                                 & \multicolumn{1}{|c|}{NR} & 25.72\%                & 42.20\%         & 22.99\%                      & 35.29\%                      \\
			                                                                                                                                                                                                                                                                                 & \multicolumn{1}{|c|}{RR} & 25.07\%                & 40.85\%         & 27.83\%                      & 41.81\%                      \\
			\midrule \multirow{2}{*}{\makecell{Spherical                                                                                                                                                                            \\YOLOv11}}                                              & \multicolumn{1}{|c|}{NR} & 24.29\%                & 40.59\%         & 15.61\%                      & 28.08\%                      \\
			                                                                                                                                                                                                                                                                                 & \multicolumn{1}{|c|}{RR} & 21.52\%                & 38.88\%         & 24.05\%                      & 38.98\%                      \\
			\bottomrule \bottomrule
		\end{tabular}
	}
	\caption{\textbf{Semantic Segmentation Results on Stanford 2D-3D-S.}}
	\label{tab: semantic segmentation random rotation quantitative}
\end{table}

%% file: tab/Single_Batch_Overfit_Test.tex
\begin{table}[!t]
	\centering
	\setlength{\tabcolsep}{2pt}
	\renewcommand{\arraystretch}{1.05}
	\resizebox{0.99\linewidth}{!}{
		\begin{tabular}{cc cc cc cc}
			\toprule                                                                                                               
			\multicolumn{2}{c}{\multirowcell{2}{\diagbox[outerleftsep=-10pt,outerrightsep=-10pt]{\textbf{Train}}{\textbf{Test}}}}                                                                                                                                                                                                                                                                                       & \multicolumn{2}{c}{Pinhole}     & \multicolumn{2}{c}{Fisheye}  & \multicolumn{2}{c}{Panoramic}                                                                                                                             \\
			\cmidrule(lr){3-4} \cmidrule(lr){5-6} \cmidrule{7-8}                                                                                                                                                                                                                                                                                                                                                        &                                                                                                                                                                                                                            
			                                                                                                                                                                                                                                                                                                                                                                                                            & mIoU $\uparrow$                 & mAcc $\uparrow$              & mIoU $\uparrow$               & mAcc $\uparrow$              & mIoU $\uparrow$              & mAcc $\uparrow$                                             \\
			\midrule \multirow{3}{*}[-0.0em]{\makecell{Planar                                                                                                                                                                                                                                                                                                  \\DeepLab v3\cite{chenRethinkingAtrousConvolution2017}}} & \multicolumn{1}{|c|}{Pinhole}   & 53.75\%                      & 59.70\%                       & \cellcolor{lightred} 33.47\% & \cellcolor{lightred} 45.73\% & \cellcolor{lightred} 19.57\% & \cellcolor{lightred} 36.40\% \\
			                                                                                                                                                                                                                                                                                                                                                                                                            & \multicolumn{1}{|c|}{Fisheye}   & 67.95\%                      & 81.58\%                       & 68.54\%                      & 82.70\%                      & \cellcolor{lightred} 57.46\% & \cellcolor{lightred} 77.46\% \\
			                                                                                                                                                                                                                                                                                                                                                                                                            & \multicolumn{1}{|c|}{Panoramic} & \cellcolor{lightred} 51.56\% & \cellcolor{lightred} 62.24\%  & \cellcolor{lightred} 55.57\% & \cellcolor{lightred} 67.91\% & 71.20\%                      & 92.12\%                      \\
			\midrule \multirow{3}{*}[-0.0em]{\makecell{Spherical                                                                                                                                                                                                                                                                                               \\DeepLab v3}}                                           & \multicolumn{1}{|c|}{Pinhole}   & 48.71\%                      & 62.21\%                       & \cellcolor{lightred} 36.51\% & \cellcolor{lightred} 62.07\% & \cellcolor{lightred} 35.62\% & \cellcolor{lightred} 61.05\% \\
			                                                                                                                                                                                                                                                                                                                                                                                                            & \multicolumn{1}{|c|}{Fisheye}   & \cellcolor{lightred} 40.27\% & \cellcolor{lightred} 45.45\%  & 54.65\%                      & 66.21\%                      & 48.04\%                      & 63.85\%                      \\
			                                                                                                                                                                                                                                                                                                                                                                                                            & \multicolumn{1}{|c|}{Panoramic} & \cellcolor{lightred} 36.54\% & \cellcolor{lightred} 42.38\%  & 58.52\%                      & 69.75\%                      & 65.71\%                      & 90.44\%                      \\
			\bottomrule
		\end{tabular}
	}
	\caption{ \textbf{Zero-shot Lens Generalizability Test.}
		Overfitted and tested on the same batch. Random rotation is disabled.}
	\label{tab: single batch overfit test}
\end{table}

%% file: tab/Ablation.tex
\begin{table}[H]
	\centering
	\resizebox{0.99\linewidth}{!}{
		\begin{tabular}{c c cc cc}
			\toprule                                               
			\multirowcell{2}{\textbf{Location}                                                                                                                       \\\textbf{Sampler}} & \multirowcell{2}{\textbf{Distance}\\\textbf{Bins}} & \multicolumn{2}{c}{NR} & \multicolumn{2}{c}{RR}                                                              \\
			\cmidrule(lr){3-4} \cmidrule(lr){5-6}                                                                                                                                        &                                                                                                                                                                   
			                                                                                                                                                                             & mIoU $\uparrow$                                    & mAcc $\uparrow$        & mIoU $\uparrow$        & mAcc $\uparrow$                                            \\
			\midrule Icosahedron                                                                                                                                                         & 3                                                  & 28.78\%                & 45.27\%                & 28.09\%                     & 41.18\%                      \\
			\midrule Icosahedron                                                                                                                                                         & 4                                                  & 27.99\%                & 45.52\%                & 23.50\%                     & 35.57\%                      \\
			Icosahedron                                                                                                                                                                  & 5                                                  & 29.66\%                & 45.36\%                & 22.82\%                     & 34.69\%                      \\
			Icosahedron                                                                                                                                                                  & 6                                                  & 29.02\%                & 45.95\%                & 21.39\%                     & 33.40\%                      \\
			\midrule Fibonacci                                                                                                                                                           & 3                                                  & 31.69\%                & 47.63\%                & 12.60\%                     & 22.79\%                      \\
			HEALPix                                                                                                                                                                      & 3                                                  & 29.59\%                & 46.98\%                & 13.87\%                     & 25.20\%                      \\
			Quasi-random                                                                                                                                                                 & 3                                                  & 29.85\%                & 48.06\%                & \cellcolor{lightred} 8.70\% & \cellcolor{lightred} 17.73\% \\
			Octahedron                                                                                                                                                                   & 3                                                  & 28.96\%                & 44.12\%                & 14.05\%                     & 24.57\%                      \\
			Hexahedron                                                                                                                                                                   & 3                                                  & 29.25\%                & 45.41\%                & 18.06\%                     & 29.27\%                      \\
			Equirectangular                                                                                                                                                              & 3                                                  & 30.25\%                & 46.69\%                & 12.87\%                     & 23.48\%                      \\
			\bottomrule
		\end{tabular}
	}
	\caption{\textbf{Ablation Study on Hyperparameters.}
		Random rotation is disabled during training.}
	\label{tab: ablation}
\end{table}

%% file: sec/5_conclusion.tex
\section{Conclusion}
\label{sec:conclusion}

We presented the \emph{Unified Spherical Frontend (USF)}, a modular and lens-agnostic framework for generic vision tasks.
USF separates projection, resampling, convolution, and pooling into independent components, and supports configurable location sampling, value interpolation, and per-layer output resolutions.
This modularity allows task-specific architectures to be composed with minimal changes to existing pipelines while scaling efficiently to high-resolution inputs.

\vspace{0.5em}
Crucially, using distance-only weighting functions guarantees rotation-equivariance by construction.
Instead of relying on heavy augmentation to approximate symmetry, USF builds equivariance directly into the geometric formulation.

\vspace{0.5em}
By representing all camera types within a single spherical domain, USF enables a unified processing space where lens-specific distortions are eliminated.
This provides a practical foundation for spherical vision as a general-purpose framework across diverse perception systems in robotics, AR/VR/MR/XR, and beyond.

%% file: sec/6_acknowledgement.tex
\section{Acknowledgement}
\label{sec:acknowledgement}
We thank Yuheng Qiu and Yuchen Zhang for helpful discussions.
This work was supported by computational resources from AirLab Cloud, CUBE Lab clusters, and Pittsburgh Supercomputing Center (PSC) Bridges-2~\cite{brownBridges2PlatformRapidlyEvolving2021}.
We also gratefully acknowledge NVIDIA for providing GPUs through academic hardware grants.

%% file: sec/X_supplement.tex
\clearpage
\setcounter{section}{0}
\renewcommand{\thesection}{\Alph{section}}
\maketitlesupplementary

\section{Conventions}
\label{sec:conventions}

In this paper, \emph{location}, \emph{coordinate}, or \emph{geometry} refers to a point on the unit sphere, represented either by Cartesian coordinates in a right-handed system:
\begin{align}
	\mathbf{p}= (x, y, z) \in \mathbb{R}^{3}, \quad \| \mathbf{p}\| = 1
\end{align}
or by polar coordinates:
\begin{align}
	(\theta, \phi) \in \mathbb{S}^{2}, \quad \theta \in [-\frac{\pi}{2}, \frac{\pi}{2}], \quad \phi \in [-\pi, \pi]
\end{align}
where $\theta$ and $\phi$ are latitudes and longitudes.
$(x, y, z) = (1, 0, 0)$ corresponds to $(\theta, \phi) = (0, 0)$, and the bidirectional mapping between Cartesian and polar coordinates is given below:
\begin{align}
	\begin{cases}x = \cos{\theta} \cos{\phi} \\ y = \cos{\theta} \sin{\phi} \\ z = \sin{\theta}\end{cases}, \begin{cases}\theta = \arctan{\left(\frac{z}{\sqrt{x^{2}+ y^{2}}}\right)} \\ \phi = \arctan{\left(\frac{y}{x}\right)}\end{cases}
\end{align}
Both Cartesian and polar coordinates are \emph{2D} parameterizations of the same manifold.
However, the spherical parameterization exhibits singularities at the poles: all $(\frac{\pi}{2}, \phi)$ map to the north pole, and all $(-\frac{\pi}{2}, \phi )$ to the south pole, regardless of $\phi$.


\input{fig/YOLOv11}

\section{Location Sampling Method}
\label{sec:location-sampling-method}
This section details point generation methods.

\subsection{Goldberg Polyhedron}
\label{ssec:goldberg-polyhedron}
Polyhedral sampling generates spherical points by subdividing each polygon face of a base convex polyhedron (e.g., tetrahedron, hexahedron, dodecahedron) with generalized barycentric coordinates.
Let a polygon face have $m$ vertices $\mathbf{v} _{1}, \dots, \mathbf{v}_{m}\in \mathbb{S}^{2}$.
For a subdivision level $n_{\text{side}}$, we enumerate all integer tuples:
\begin{align}
	(i_{1}, \dots, i_{m}), \; i_{k}\geq 0, \; \sum\limits_{k = 1}^{m}i_{k}= n_{\text{side}},
\end{align}

Each tuple defines a generalized barycentric combination inside the face:
\begin{align}
	\mathbf{p}= \sum\limits_{k = 1}^{m}\frac{i_{k}}{n_{\text{side}}}\mathbf{v}_{k}.
\end{align}

After normalization to unit norm, this yields a point on the sphere.
Applying this process to all faces yields a near-uniform spherical point set.
Different polyhedra use the same subdivision rule but differ in how $N_{\text{side}}$ relates to the target average pixel area.

In this work, “Goldberg polyhedron” refers only to the \emph{spherical Voronoi} tessellation induced by the sampled points, which typically manifests as a mixture of pentagons and hexagons.
The sampling itself is purely barycentric.

\subsection{HEALPix}
\label{ssec:healpix}
Refer to the HEALPix Primer\cite{gorskiHEALPixPrimer1999a}.

\subsection{Fibonacci Lattice}
\label{ssec:fibonacci-lattice}
\begin{align}
	\mathbf{p}_{i} & = (\sin{\phi_i}\cos{\theta_i}, \; \sin{\phi_i}\sin{\theta_i}, \; \cos{\phi_i}), \\
	\phi_{i}       & = \arccos{\left(1 - \frac{2i}{N}\right)},                                       \\
	\theta_{i}     & = \frac{2 \pi \cdot i}{\varphi}, \; \varphi = \frac{1 + \sqrt{5}}{2}.
\end{align}

Where $\varphi$ is the golden ratio, $N$ is the total number of points.
Note that $\phi$ and $\theta$ here are different from the definition in \cref{sec:conventions}.

\subsection{Quasi-Random Sampling}
\label{ssec:quasi-random-sampling}
Quasi-random sampling generates low-discrepancy sequences similar to Sobol or Halton sequences.

First, generate evenly distributed points in $[0, 1]^{2}$ with irrational-ratio recurrence.
Specifically, the plastic constant $\psi \approx 1.32471795724474602596$, the unique real root of $\psi^{3}- \psi - 1 = 0$, is adopted as the irrational number.
Define two incommensurate step sizes $\begin{cases} \alpha_{u} & = \psi^{-1}\approx 0.7549, \\ \alpha_{v} & = \psi^{-2}\approx 0.5698 \end{cases}$ together with starting offsets $s_{0, u}= s_{0, v}= 0.5$, 2D quasi-random sequence can be generated by:
\begin{align}
	u_{i} & = (s_{0, u}+ i \cdot \alpha_{u}) \bmod 1, \\
	v_{i} & = (s_{0, v}+ i \cdot \alpha_{v}) \bmod 1,
\end{align}
for $i = 0, \dots, N - 1$.

Then, each point $(u_{i}, v_{i}) \in [0, 1]^{2}$ is mapped to the unit sphere with the Lambert equal-area projection:
\begin{align}
	\mathbf{p}_{i} & = \left(r_{i}\cos{\phi_i}, \; r_{i}\sin{\phi_i}, z_{i}\right), \\
	z_{i}          & = 1 - 2 u_{i},                                                 \\
	r_{i}          & = \sqrt{1 - z_{i}^{2}},                                        \\
	\phi_{i}       & = 2 \pi \cdot v_{i}.
\end{align}


\section{Value Interpolation RBF Kernel}
\label{sec:value-interpolation-rbf-kernel}
For an output location $\mathbf{p}_{o}$, let $\mathcal{N}(p_{o})$ be the collection of neighbor input points and $\omega_{k}$ be the weight applied to input value $x_{k}$ at input location $\mathbf{p}_{k}$.
$d(\mathbf{p}_{i}, \mathbf{p} _{j})$ is the geodesic distance between point $\mathbf{p}_{i}$ and $\mathbf{p}_{j}$.

The interpolated value $x_{o}$ at a new output location $\mathbf{p}_{o}$ is expressed as:
\begin{align}
	x_{o} & = \sum\limits_{k\in\mathcal{N}(\mathbf{p}_o)}\omega_{k}\; x_{k}
\end{align}

Where weights $\omega_{k}$ may come from different RBF kernels.
Weights are normalized so that their sum equals 1 for an output value.

\input{fig/UNet}

\subsection{Softmax}
\label{ssec:softmax}
\begin{align}
	\omega_{k} & = \frac{e^{-\frac{d(\mathbf{p}_k, \mathbf{p}_o)}{\mathcal{T}}}}{\sum\limits_{i\in\mathcal{N}(\mathbf{p}_o)}e^{-\frac{d(\mathbf{p}_i, \mathbf{p}_o)}{\mathcal{T}}}}
\end{align}

$\mathcal{T}$ is the temperature hyperparameter that controls sharpness.

\subsection{Gaussian}
\label{ssec:gaussian}
\begin{align}
	\omega_{k} & = e^{-\frac{d(\mathbf{p}_{k}, \mathbf{p}_{o})^{2}}{2\sigma^{2}}}
\end{align}


\subsection{Hann}
\label{ssec:hann}
\cite{kivinukkSamplingOperatorsDefined2003}:
\begin{align}
	\omega_{k} & = \frac{1}{2}\left(1 + \cos{(\pi \cdot d(\mathbf{p}_k, \mathbf{p}_o)}\right)
\end{align}

\subsection{Wendland-C2}
\label{ssec:wendland-c2}
\cite{biancoliniRadialBasisFunctions2020}:
\begin{align}
	\omega_{k} & = (1 - d(\mathbf{p}_{k}, \mathbf{p}_{o}))^{4}\cdot (1 + 4 \cdot d(\mathbf{p}_{k}, \mathbf{p}_{o}))
\end{align}


\section{Continuous Spherical Spatial Correlation}
\label{sec:continuous-spherical-spatial-correlation}
Analogous to a corollary in Fourier analysis for planar CNNs, we prove that spherical correlation in the spatial domain corresponds to multiplication in the spherical harmonic (frequency) domain.

Let $f, \mathcal{K}: \mathbb{S}^{2}\rightarrow \mathbb{C}$ denote signals on $\mathbb{S}^{2}$, we define their spherical correlation $f \star \mathcal{K}$ by:
\begin{align}
	(f \star \mathcal{K})(\omega) = \int\limits_{\mathbb{S}^2}f(\omega') \; \overline{\mathcal{K}(R_\omega^{-1} \omega')}\; d\Omega(\omega'),
\end{align}
Where $\omega, \omega' \in \mathbb{S}^{2}$, $R_{\omega}$ is any fixed rotation in $\mathrm{SO}(3)$ such that $R_{\omega}(\mathbf{n}) = \omega$ with $\mathbf{n}= (0 , 0, 1)$ denoting the north pole.
$d\Omega(\omega')$ is the surface measure on $\mathbb{S}^{2}$, and $\overline{\cdot}$ denotes complex conjugation.

Any $f \in L^{2}(\mathbb{S}^{2}): \mathbb{S}^{2}\rightarrow \mathbb{C}$ admits a spherical harmonic expansion:
\begin{align}
	f(\omega)         & = \sum\limits_{\ell = 0}^{\infty}\sum\limits_{m = -\ell}^{\ell}\hat{f}_{\ell, m}\; Y_{\ell}^{m}(\omega), \\
	\hat{f}_{\ell, m} & = \int\limits_{\mathbb{S}^2}f(\omega) \; \overline{Y_\ell^m(\omega)}\; d\Omega (\omega),
\end{align}

Where $\{Y_{\ell}^{m}\}$ are harmonic functions that form an orthonormal basis of $L^{2}(\mathbb{S}^{2})$, likewise for $\mathcal{K}(\omega)$.

We now derive the spherical harmonic coefficients of $f \star \mathcal{K}$ as:
\begin{align}
	  & \; \widehat{(f \star \mathcal{K})}_{\ell, m}\notag                                                                                                                                               \\
	= & \int\limits_{\mathbb{S}^2}(f \star \mathcal{K}) (\omega) \; \overline{Y_\ell^m(\omega)}\; d\Omega (\omega)                                                                                       \\
	= & \int\limits_{\mathbb{S}^2}\left[ \int\limits_{\mathbb{S}^2}f(\omega') \; \overline{\mathcal{K}(R_\omega^{-1} \omega')}\; d\Omega(\omega') \right] \overline{Y_\ell^m (\omega)}\; d\Omega(\omega) \\
	= & \int\limits_{\mathbb{S}^2}f(\omega') \left[ \int\limits_{\mathbb{S}^2}\overline{\mathcal{K}(R_\omega^{-1} \omega')}\; \overline{Y_\ell^m (\omega)}\; d\Omega(\omega)\right] d\Omega(\omega')
\end{align}

For general anisotropic kernel $\mathcal{K}$, $\widehat{(f \star \mathcal{K})}_{\ell, m}$ becomes a finite sum of products $\hat{f}_{\ell, m}\; \overline{\hat{\mathcal{K}}_{\ell, m}}$ multiplied by algebraic factors\cite{driscollComputingFourierTransforms1994,kostelecFFTsRotationGroup2008}:
\begin{align}
	\widehat{(f \star \mathcal{K})}_{\ell, m}= \sum\limits_{m' = -\ell}^{\ell}\hat{f}_{\ell, m'}\; \overline{\hat{\mathcal{K}}_{\ell, m'}}\; \Lambda_{\ell}(m, m').
\end{align}

Which corresponds to multiplication in the spherical harmonic domain.


\section{Rotation Equivariant Correlation}
\label{sec:rotation-equivariant-correlation}
As detailed in\cite{estevesLearningSO3Equivariant2020}, when the kernel $\mathcal{K}$ is isotropic, meaning that it is zonal, radial, isotropic, and its value only depends on the relative geodesic distance, $\widehat{(f \star \mathcal{K})} _{\ell, m}$ simplifies to:
\begin{align}
	\widehat{(f \star \mathcal{K})}_{\ell, m}= \alpha_{\ell}\; \hat{f}_{\ell, m}
\end{align}

Which is purely diagonal pointwise multiplication in $(\ell, m)$, where $\alpha_{\ell}$ can be regarded as the kernel's "\emph{frequency response}", instead of "\emph{phase response}" ($m$-mode coefficients) that changes with rotation, hence preserving rotation-equivariance.

Intuitively, if we rotate all the coordinates $\omega$ of a spherical image with any rotation $R \in \mathrm{SO}(3)$, $\mathcal{K}$ activates $R\omega$ with the same set of weights for local signals around $R\omega$ regardless of orientation because it's isotropic.


\input{fig/Heatmap}

\section{Computation Optimization}
\label{sec:computation-optimization}
A key design consideration of USF is that operations contraction schemes changes only with respect to geometry and not feature values.
This allows expensive geometric computations such as neighborhood construction, interpolation weights, and sparse aggregation structures to be precomputed once and reused across subsequent forward passes.
This geometry caching mechanism is critical for making high-resolution spherical processing computationally feasible in practice.

Our implementation utilizes several optimized libraries for efficient spherical processing.
We use \textit{FAISS}~\cite{johnsonBillionScaleSimilaritySearch2021} for fast nearest-neighbor search during neighborhood construction, \textit{torch\_scatter}~\cite{feyTorchScatter2023} for efficient neighborhood aggregation and interpolation, and \textit{opt\_einsum}~\cite{smithOpt_einsumPythonPackage2018} for optimized tensor contraction ordering in dense operations.
These components significantly reduce overhead compared to naive implementations and make large-scale spherical processing feasible within the PyTorch framework, even without custom CUDA kernels.

In all benchmarks, the input to both spherical and planar pipelines is a $960 \times 480$ panorama image with batch size 8 and RGB channels.
All experiment results are averaged over 10 runs on an NVIDIA H200 GPU using PyTorch 2.8.0, CUDA 12.8, torch\_scatter 2.1.2, and float32 precision.

\subsection{Location Sampling and Value Interpolation}
\label{ssec:location-sampling-and-value-interpolation}
The location sampling resolution factor is set to 1.0, meaning the number of output spherical samples is matched to the number of input pixels with less than 1\% difference.
The sampling benchmark measures the runtime of location sampling and value interpolation separately under both cold-start (first run) and sustained (cached) settings.
In the cold-start scenario, geometric structures such as spherical neighborhood structure and interpolation weights must be constructed.
In the sustained setting, these structures are reused and only value aggregation via matrix multiplication is performed.

\input{fig/Sampling_Computation_Benchmark}

As shown in \cref{fig:sampling_computation_benchmark}, the cold-start cost is dominated by geometric preprocessing and can take several seconds depending on the sampling method and interpolation kernel.
However, once geometry is cached, the runtime drops to the millisecond level across all sampling methods, representing orders-of-magnitude speedup.
This demonstrates that the computational bottleneck lies in geometry construction rather than interpolation itself, and confirms that geometry caching is essential for practical spherical processing.

This result highlights a fundamental property of spherical pipelines: while geometry-aware processing introduces an initial preprocessing cost, this cost is amortized over all subsequent forward passes, making sustained runtime comparable to standard image processing pipelines.

\subsection{Spherical Convolution and Pooling}
\label{ssec:spherical-layer-computation-benchmark}
We next benchmark individual spherical convolution and pooling operators and compare them with planar Conv2d and MaxPool2d layers.
For fairness, the planar convolution uses a $5 \times 5$ kernel, since each spherical output location aggregates approximately $25 - 30$ neighboring samples on average, making the receptive field sizes comparable.

Three spherical convolution implementations are evaluated:

\begin{enumerate}
	\item \textbf{Continuous Distance $\times$ Direction MLP:} Distance and direction weighting functions are parameterized by an MLP with hidden dims $[16, 16]$ and positional encoding $L = 8$.
	During training, the MLP must be evaluated for each neighbor pair, but during inference, weights can be fully cached once geometry is fixed.
	\item \textbf{Discrete Distance Piecewise-Constant (PWC) (Dense):} The MLP is replaced by discrete distance bins implemented using torch embedding.
	\item \textbf{Discrete Distance Piecewise-Constant (PWC) (Sparse):} Aggregation is performed by sparse matrix multiplication.
\end{enumerate}

\input{fig/Spherical_Layer_Computation_Benchmark}

\Cref{fig:spherical_layer_computation_benchmark} show that cold-start runtime is significantly higher due to kernel construction and sparse structure generation.
However, sustained runtime drops substantially once geometry and weights are cached.
Among spherical implementations, the discrete PWC (sparse) implementation achieves the fastest sustained runtime, while the continuous MLP variant benefits significantly from weight caching during inference.

Although spherical operators remain slower than highly optimized planar Conv2d kernels, the gap narrows significantly in sustained execution.
This difference is largely due to the fact that planar convolutions are implemented as heavily optimized CUDA kernels, whereas spherical operators are currently implemented using PyTorch scatter and sparse operations rather than custom fused CUDA kernels.

\subsection{Network-Level Comparison}
\label{ssec:network-level-comparison}
Finally, we compare full network runtime between planar and spherical versions of YOLOv11, DeepLab v3, and UNet using identical macro-architectures, input resolution, batch size, and channel configurations.
The only difference between models is the replacement of planar convolution and pooling layers with spherical counterparts, making the comparison strictly architecture-level and apples-to-apples.

\input{fig/Network_Computation_Benchmark}

\Cref{fig:network_computation_benchmark} shows that cold-start runtime for spherical networks is significantly higher due to geometry construction and kernel initialization.
However, in sustained execution, spherical networks become much closer to planar networks in runtime.
Across models, spherical networks take approximately $\mathbf{2 \times}$ \textbf{the overall time} of their planar counterparts in full-scale training, which we consider acceptable given the additional geometric processing and built-in rotation-equivariance.

Interestingly, the spherical networks often require \textbf{fewer FLOPs} than the planar counterparts while still taking longer in wall-clock time.
This phenomenon is common in custom operator research: theoretical arithmetic complexity (FLOPs) does not directly translate to runtime when standard operators benefit from highly optimized low-level CUDA kernels, while custom operators rely on higher-level sparse and scatter operations.
Further performance gains could be achieved by implementing dedicated CUDA kernels for spherical sampling and convolution.

\vspace{1.0em}
Overall, these benchmarks demonstrate that \textbf{geometry caching is the key enabler} that makes high-resolution spherical processing practical.
Without caching, the cost of geometric preprocessing would dominate runtime and make spherical pipelines infeasible for large-scale vision tasks.
With caching, however, sustained runtime becomes comparable to planar networks while providing additional geometric consistency, rotation-equivariance, and lens-agnostic processing capabilities.


\section{Common Training Setup}
\label{sec:common-training-setup}
If not otherwise mentioned in the main paper, all the full-scale experiments share the same configurations as follows:

\begin{itemize}
	\item Icosahedron location sampler for both the resampling stage and all the intermediate output locations of spherical convolution and pooling layers.

	\item \textit{AdamW} optimizer with learning rate $1 \times 10^{-3}$, weight decay $1 \times 10^{-2}$, learning rate scheduler warms up at the first $40\%$ steps and drops to $1 \times 10^{-2}$ of the learning rate at the end.

	\item Batch size of 8 with 2 distributed data parallel (DDP), trained for 200 epochs on NVIDIA A100 and H200 GPUs.

	\item Data augmentation consists of $50\%$ chance of random chroma, luma jitter, gaussian blur, $5\%$ chance of grayscaling, and $50\%$ chance of random horizontal reflection.
\end{itemize}


\input{tab/Full-dataset_Lens_Adaptation_Quantitative}

\section{Backbone Architecture}
\label{sec:backbone-architecture}
We applied our framework to three representative backbone architectures: YOLOv11 (\cref{fig:YOLOv11}), UNet (\cref{fig:UNet}), and DeepLab v3 (\cref{fig:DeepLab v3}).

\input{fig/DeepLab_v3}

We incorporate attention bias as positional encoding in the C2PSA self-attention layers with pairwise Euclidean distance in the planar domain and geodesic distance on the sphere, analogous to the mechanism in ALiBi\cite{pressTrainShortTest2022}.

Note that atrous (dilated) convolution is not defined for our generic spherical CNN, so we use larger kernels instead in ASPP block of spherical DeepLab v3.

Each downsampling layer in the spherical model reduces the number of output points to $\frac{1}{4}$ of the input, analogous to stride-2 downsampling in planar CNNs.
Because location sampling is not invertible with respect to resolution factors, we align the output locations of spherical CNN upsampling layers with corresponding downsampling layers to ensure valid channel concatenations.


\section{Object Detection}
\label{sec:object-detection}
The loss function includes focal loss for category classification and L1 loss for bounding box center $(\theta, \phi)$, size $(\alpha, \beta)$, and angle $\gamma$ regression.
Formally, focal loss:
\begin{align}
	L_{\text{focal}}    & = \frac{1}{N}\sum\limits_{i=1}^{N}[ L_{\text{positive}}+ L_{\text{negative}}],          \\
	L_{\text{positive}} & = -\ln(p_{i}) (1 - p_{i})^{\lambda_\text{positive}}y_{i},                               \\
	L_{\text{negative}} & = -\ln(1 - p_{i}) p_{i}^{\lambda_\text{positive}}(1 - y_{i})^{\lambda_\text{negative}}.
\end{align}

Where $p_{i}$ is the predicted probability, $y_{i}$ is the ground truth heatmap, and $\lambda_{\text{positive}}, \lambda_{\text{negative}}$ are hyperparameter exponents controlling the relative focus on positive versus negative detections.

\paragraph{Efficient Pairwise IoU.}
We propose a generic and vectorized method to compute pairwise IoU between arbitrary-shaped bounding boxes.
Each bounding box is converted into a binary mask over the same set of dense and uniformly sampled spherical points.
The intersection and union are then approximated by applying logical AND and OR operations on these masks, with IoU estimated as the ratio between the number of points inside the intersection and those in the union.

\vspace{0.5em}

Finally, we apply Matrix Non-Maximum Suppression (NMS)\cite{wangSOLOv2DynamicFast2020} with a Gaussian decay factor $\sigma = 5.0$ and score threshold 0.3 to suppress overlapping but inconfident proposals.

As illustrated in \cref{fig:heatmap and regression mask}, classification heatmap and bounding box regression are supervised with separate loss functions at different locations, and regression losses are only applied to points near the object's center.

Qualitative results are in \cref{tab: object detection qualitative} in addition to the quantitative results in the main paper.
Observe the \textcolor{cyan}{cyan boxes} carefully for light patches on the ground in the NR and RR cases of Spherical YOLOv11, which exhibits the same orientation with or without random rotation.
This behavior is expected because orientation is the raw output of one head, which should not change under rotation because of rotation-equivariance.


\section{Semantic Segmentation}
\label{sec:semantic-segmentation}
Training uses a composite loss of 70\% cross-entropy and 30\% Dice loss, with proper class weights and 0.05 label smoothing.
Formally, Dice loss:
\begin{align}
	L_{\text{Dice}}   & = 1 - \frac{1}{|S|}\sum\limits_{c \in S}\mathrm{Dice}_{c},                                                                          \\
	\mathrm{Dice}_{c} & = \frac{2 \cdot \sum\limits_{i \in \Omega}p_{i,c}\; y_{i,c}}{\sum\limits_{i \in \Omega}p_{i,c}+ \sum\limits_{i \in \Omega}y_{i,c}}.
\end{align}

Where $p_{i, c}$ and $y_{i, c}$ denote the predicted probability and ground-truth label for class $c$ at pixel $i$, $\Omega$ is the set of valid pixels, and $S$ is the set of non-ignored classes in $\Omega$.

We follow the official 3-fold cross-validation scheme as a benchmark on the Stanford-2D-3D-S dataset.

Random rotation qualitative results are visualized in \cref{tab: semantic segmentation random rotation qualitative} in addition to the quantitative results in the main paper.
Zero-shot lens adaptability results trained on the full-scale dataset are shown in \cref{tab: full-dataset lens adaptation quantitative} and \cref{tab: full-dataset lens adaptation qualitative}.
Ideally, a perfect model that adapts to all lenses would have the same performance in all the entries of the square performance matrix.
USF reduces off-diagonal performance degradations compared to planar models, but not yet completely eliminates them.

\input{tab/Object_Detection_Qualitative}

\input{tab/Semantic_Segmentation_Random_Rotation_Qualitative}

\input{tab/Full-dataset_Lens_Adaptation_Qualitative}

%% file: fig/YOLOv11.tex
\begin{figure*}[!ht]
	\centering
	\includegraphics[width=\linewidth]{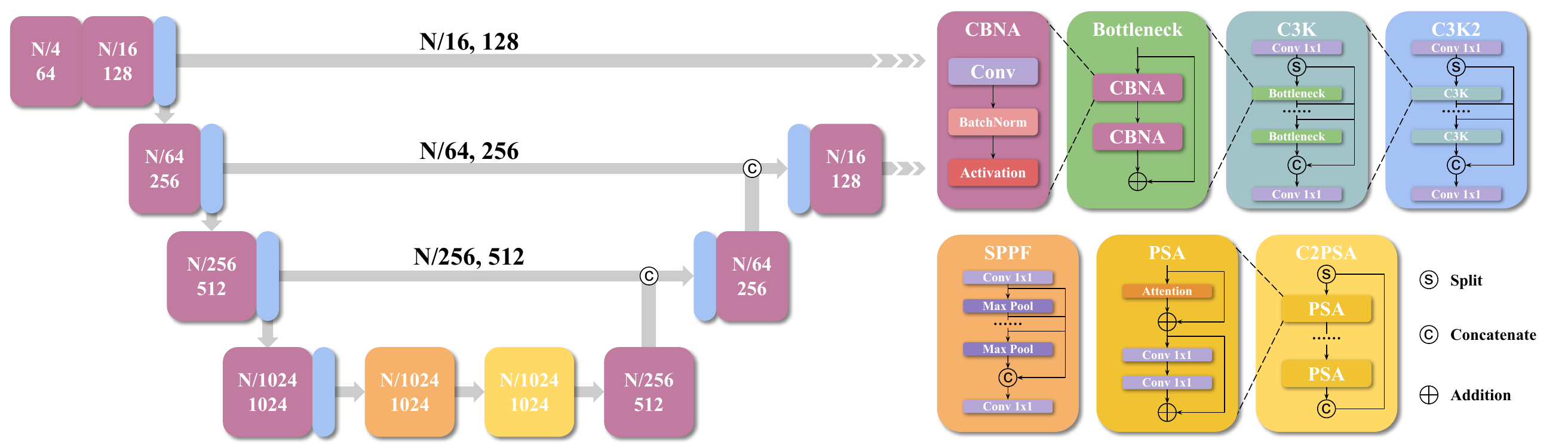}
	\caption{\textbf{YOLOv11.}
		Core components include: \textbf{CBNA}: convolution, batch normalization, activation; \textbf{SPPF}: spatial pyramid pooling-fast; \textbf{PSA}: Partial Spatial Attention.}
	\label{fig:YOLOv11}
\end{figure*}

%% file: fig/UNet.tex
\begin{figure*}[!ht]
	\centering
	\includegraphics[width=\linewidth]{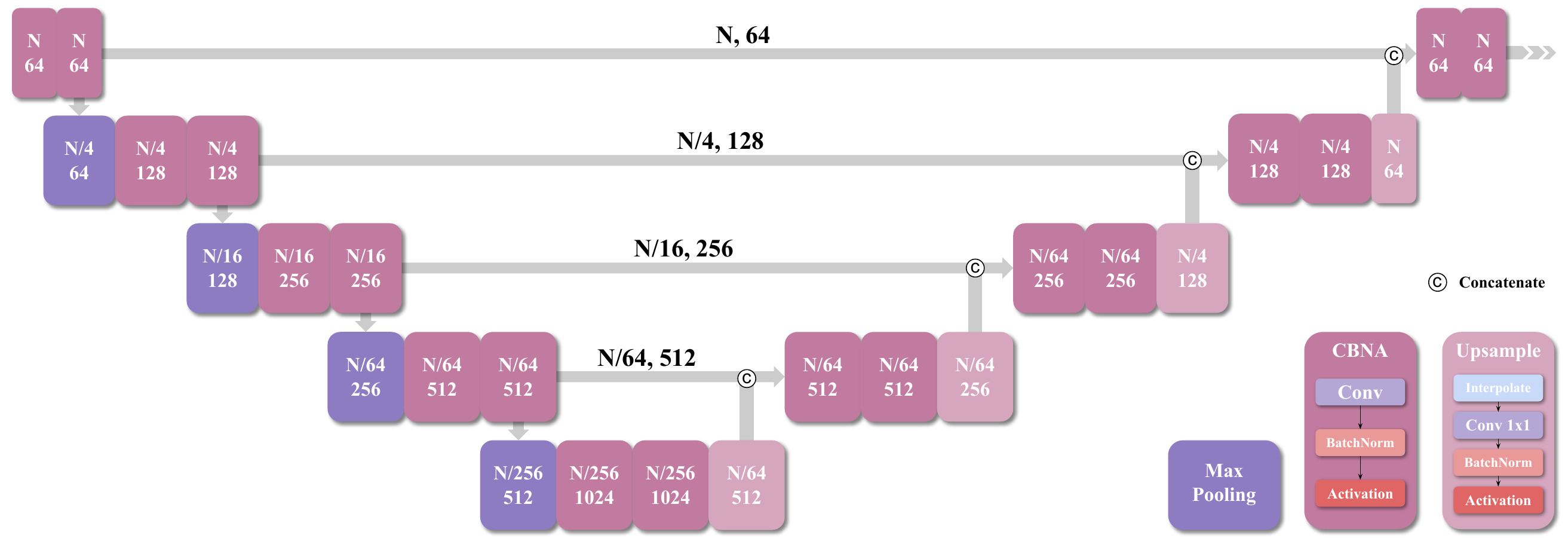}
	\caption{\textbf{UNet.}
		Core components include: \textbf{CBNA}: convolution, batch normalization, activation; \textbf{Upsample}: value interpolation, 1x1 channel-wise convolution, batch normalization, activation.}
	\label{fig:UNet}
\end{figure*}

%% file: fig/Heatmap.tex
\begin{figure*}[!ht]
	\centering

	\begin{subfigure}
		[t]{0.24\linewidth}
		\centering
		\includegraphics[width=.9\linewidth]{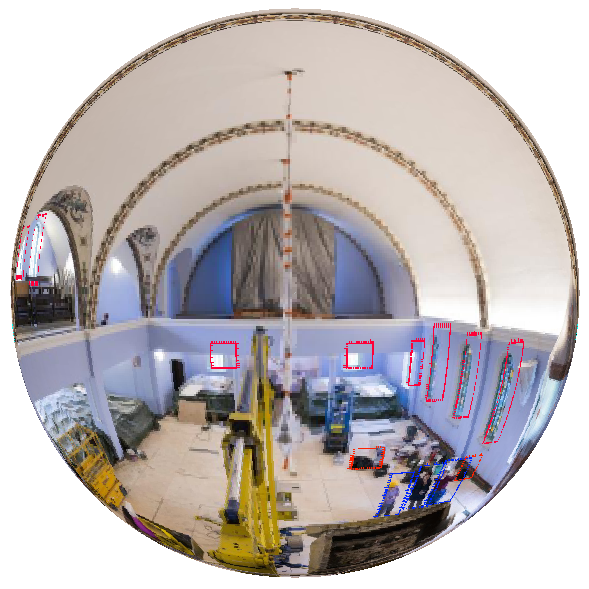}
		\caption{\textbf{Spherical Image with RBFoVs.}
			\textcolor{red}{Red} boxes denote windows.}
	\end{subfigure}
	\hfill
	\begin{subfigure}
		[t]{0.24\linewidth}
		\centering
		\includegraphics[width=.9\textwidth]{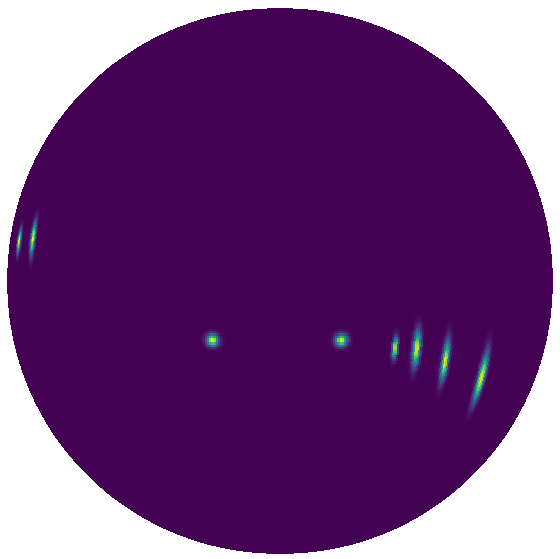}
		\caption{\textbf{Continuous Heatmap.}
			Per-category gaussian heatmaps are generated through gnomonic projection regardless of the point distribution on the sphere.}
	\end{subfigure}
	\hfill
	\begin{subfigure}
		[t]{0.24\linewidth}
		\centering
		\includegraphics[width=.9\linewidth]{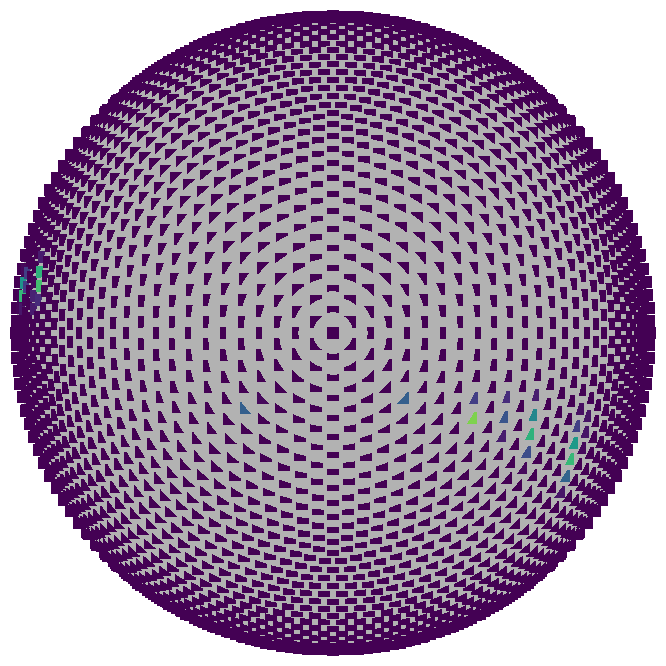}
		\caption{\textbf{Discrete Heatmap.}
			Used in actual supervision.}
	\end{subfigure}
	\hfill
	\begin{subfigure}
		[t]{0.24\linewidth}
		\centering
		\includegraphics[width=.9\linewidth]{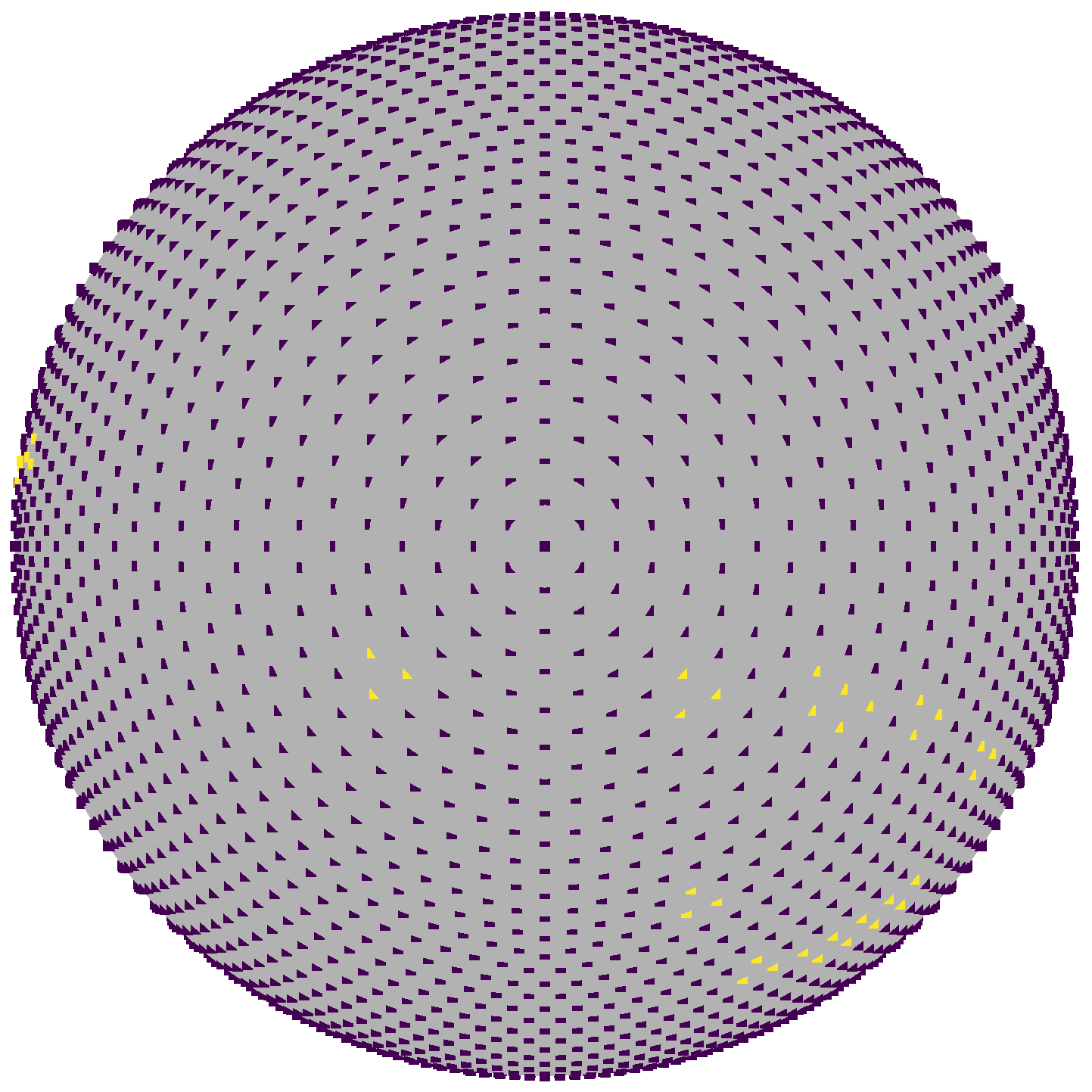}
		\caption{\textbf{Regression Mask.}
			L1 regression loss only supervises the \textcolor{Dandelion}{yellow} points of regression head outputs, where RBFoVs are centered.}
	\end{subfigure}

	\caption{\textbf{Object Detection Supervision.}}
	\label{fig:heatmap and regression mask}
\end{figure*}

%% file: fig/Sampling_Computation_Benchmark.tex
\begin{figure}[ht]
	\centering

	\includegraphics[width=\linewidth]{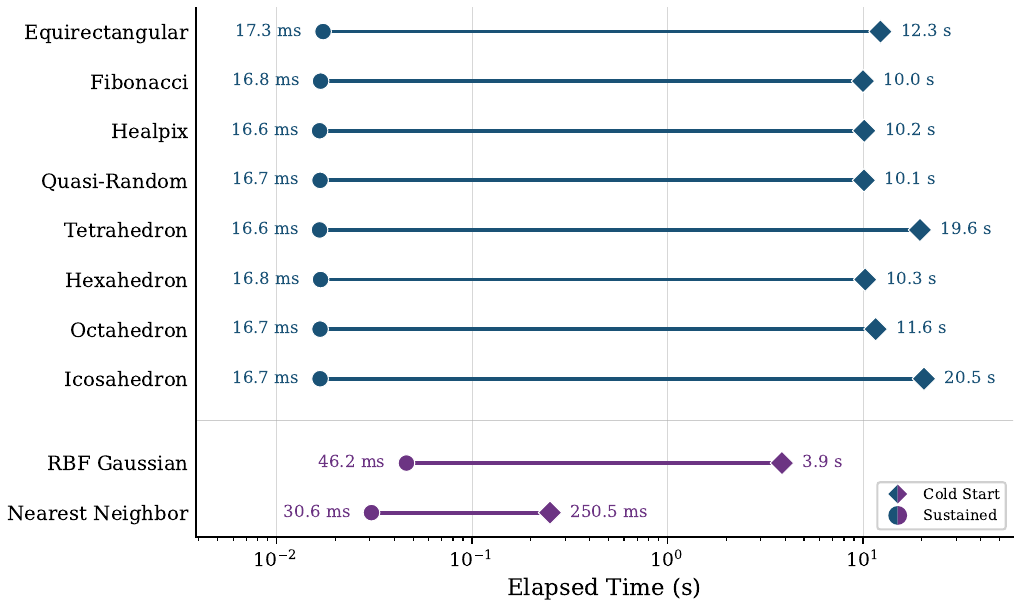}
	\caption{\textbf{Location Sampling and Value Interpolation Benchmarks.}
		Cold-start runtime includes geometric preprocessing such as neighborhood construction and interpolation weight computation.
		Sustained runtime reuses geometry and only performs value aggregation via matrix multiplication.}
	\label{fig:sampling_computation_benchmark}
\end{figure}

%% file: fig/Spherical_Layer_Computation_Benchmark.tex
\begin{figure}[ht]
	\centering

	\includegraphics[width=\linewidth]{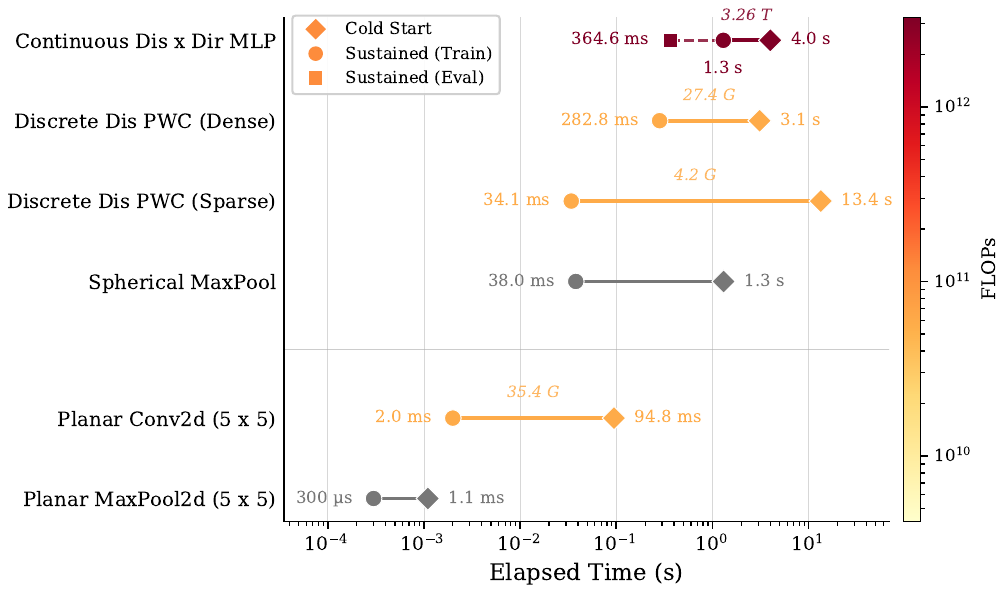}
	\caption{\textbf{Spherical Convolution and Pooling Benchmarks.}
		Sustained inference benefits from weight caching and reduces spherical convolution to matrix multiplication.}
	\label{fig:spherical_layer_computation_benchmark}
\end{figure}

%% file: fig/Network_Computation_Benchmark.tex
\begin{figure}[ht]
	\centering

	\includegraphics[width=\linewidth]{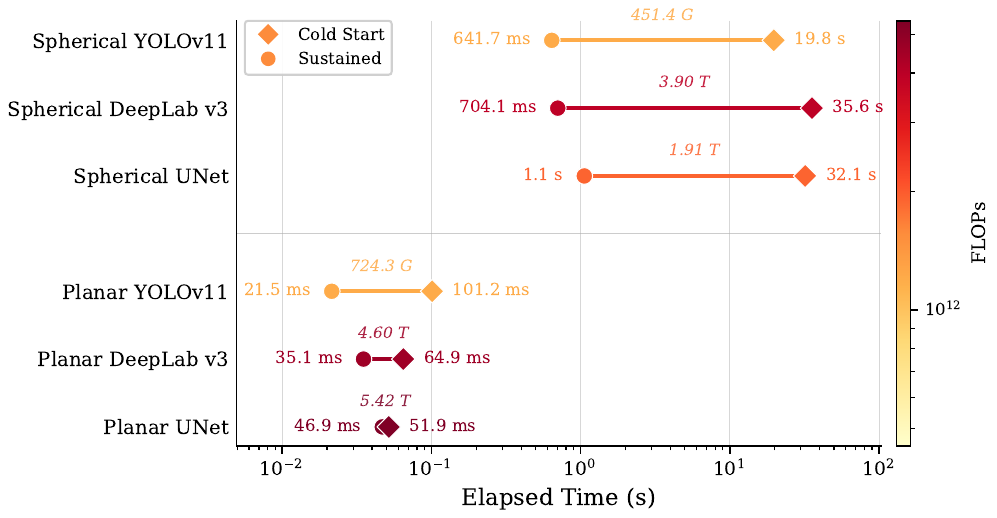}
	\caption{\textbf{End-to-End Network-Level Comparison Benchmarks.}
		Spherical networks achieve comparable runtime while requiring fewer FLOPs but currently lack fully optimized CUDA kernels.}
	\label{fig:network_computation_benchmark}
\end{figure}

%% file: tab/Full-dataset_Lens_Adaptation_Quantitative.tex
\begin{table*}
	[ht]
	\centering
	\begin{tabular}{cc cc cc cc}
		\toprule \multicolumn{2}{c}{\multirowcell{2}{\diagbox[outerleftsep=-10pt,outerrightsep=-10pt]{\textbf{Train}}{\textbf{Test}}}}                                                                                                                                                                                                                                                                                       & \multicolumn{2}{c}{Pinhole}     & \multicolumn{2}{c}{Fisheye}  & \multicolumn{2}{c}{Panoramic}                                                                                                                             \\
		\cmidrule(lr){3-4} \cmidrule(lr){5-6} \cmidrule{7-8}                                                                                                                                                                                                                                                                                                                                                                 &                                                                                                                                                                                                                            
		                                                                                                                                                                                                                                                                                                                                                                                                                     & mIoU $\uparrow$                 & mAcc $\uparrow$              & mIoU $\uparrow$               & mAcc $\uparrow$              & mIoU $\uparrow$              & mAcc $\uparrow$                                             \\
		\midrule \multirow{3}{*}{\makecell{Planar                                                                                                                                                                                                                                                                                                                   \\DeepLab v3\cite{chenRethinkingAtrousConvolution2017}}} & \multicolumn{1}{|c|}{Pinhole}   & 42.53\%                      & 55.75\%                       & \cellcolor{lightred} 33.53\% & \cellcolor{lightred} 47.36\% & \cellcolor{lightred} 36.07\% & \cellcolor{lightred} 53.61\% \\
		                                                                                                                                                                                                                                                                                                                                                                                                                     & \multicolumn{1}{|c|}{Fisheye}   & 39.88\%                      & 53.46\%                       & 40.05\%                      & 55.86\%                      & \cellcolor{lightred} 33.11\% & \cellcolor{lightred} 56.53\% \\
		                                                                                                                                                                                                                                                                                                                                                                                                                     & \multicolumn{1}{|c|}{Panoramic} & 29.66\%                      & 43.85\%                       & \cellcolor{lightred} 24.91\% & \cellcolor{lightred} 37.54\% & 35.01\%                      & 58.30\%                      \\
		\midrule \multirow{3}{*}{\makecell{Spherical                                                                                                                                                                                                                                                                                                                \\DeepLab v3}}                                           & \multicolumn{1}{|c|}{Pinhole}   & 34.76\%                      & 47.47\%                       & \cellcolor{lightred} 22.36\% & \cellcolor{lightred} 35.52\% & \cellcolor{lightred} 19.70\% & \cellcolor{lightred} 35.09\% \\
		                                                                                                                                                                                                                                                                                                                                                                                                                     & \multicolumn{1}{|c|}{Fisheye}   & \cellcolor{lightred} 19.44\% & \cellcolor{lightred} 31.52\%  & 30.44\%                      & 44.21\%                      & 28.16\%                      & 43.99\%                      \\
		                                                                                                                                                                                                                                                                                                                                                                                                                     & \multicolumn{1}{|c|}{Panoramic} & \cellcolor{lightred} 12.57\% & \cellcolor{lightred} 23.05\%  & 28.35\%                      & 41.58\%                      & 28.78\%                      & 45.27\%                      \\
		\bottomrule
	\end{tabular}
	\caption{ \textbf{Semantic Segmentation Full-dataset Lens Adaptability Test.}
		Random Rotation is disabled.}
	\label{tab: full-dataset lens adaptation quantitative}
\end{table*}

%% file: fig/DeepLab_v3.tex
\begin{figure}[hb]
	\centering
	\includegraphics[width=\linewidth]{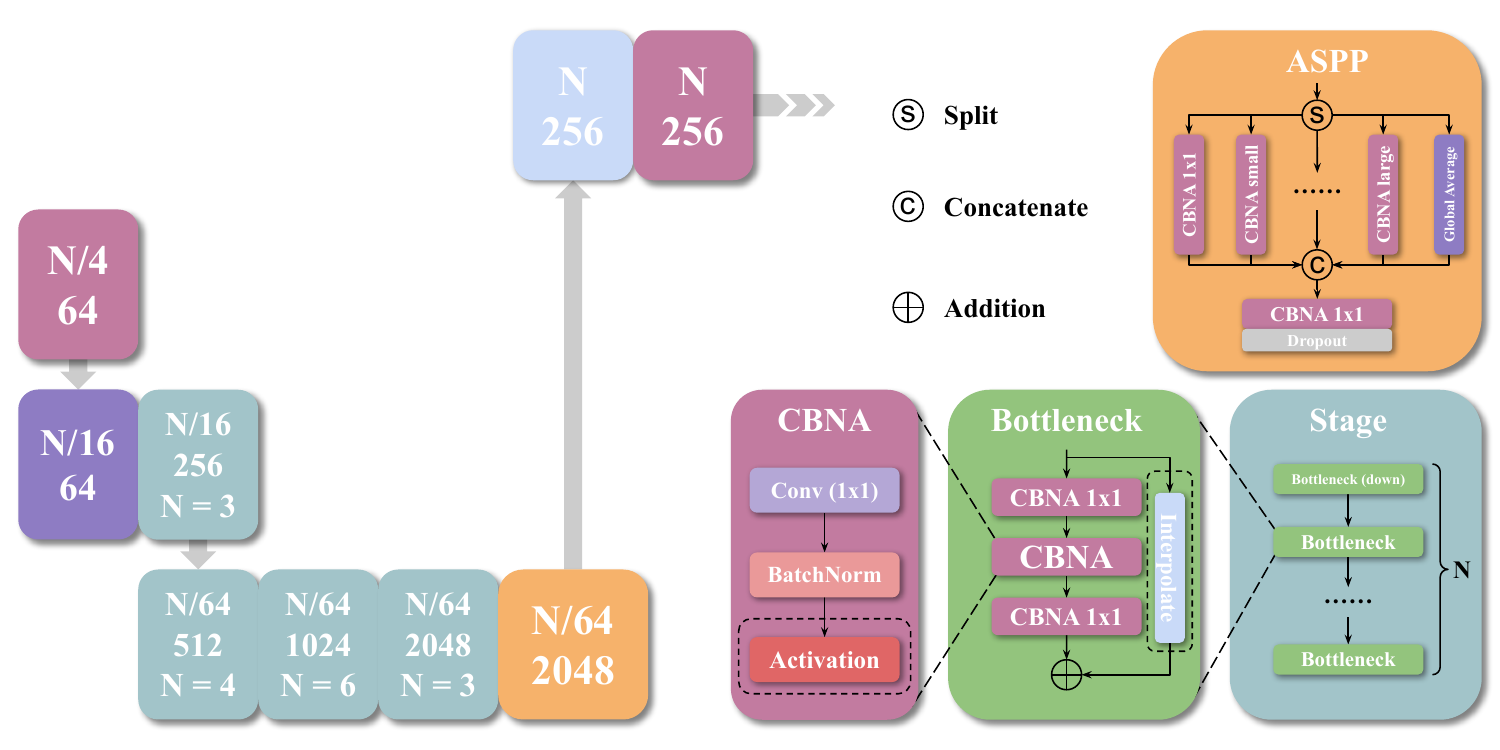}
	\caption{\textbf{DeepLab v3.}
		Core components include: \textbf{CBNA}: convolution, batch normalization, activation; \textbf{Stage}: N Bottleneck blocks; \textbf{ASPP}: atrous spatial pyramid pooling.}
	\label{fig:DeepLab v3}
\end{figure}

%% file: tab/Object_Detection_Qualitative.tex
\begin{table*}
	[!ht]
	\centering
	\begin{tabular}{cc cc}
		\toprule                                                                                              
		\multicolumn{2}{c|}{\diagbox[outerleftsep=-40pt,outerrightsep=-40pt]{\textbf{Train}}{\textbf{Test}}}                                                                                                                                                                                                                                                                                                                         & NR                                                                                & RR                                                                                                                                                                                                           \\
		\midrule \multirow{2}{*}[-4.5em]{\makecell{Planar                                                                                                                                                                                                                                                                                                                             \\YOLOv11\cite{khanamYOLOv11OverviewKey2024}}} & \multicolumn{1}{|c|}{NR}                                                          & \includegraphics[valign=c,width=.35\linewidth,keepaspectratio]{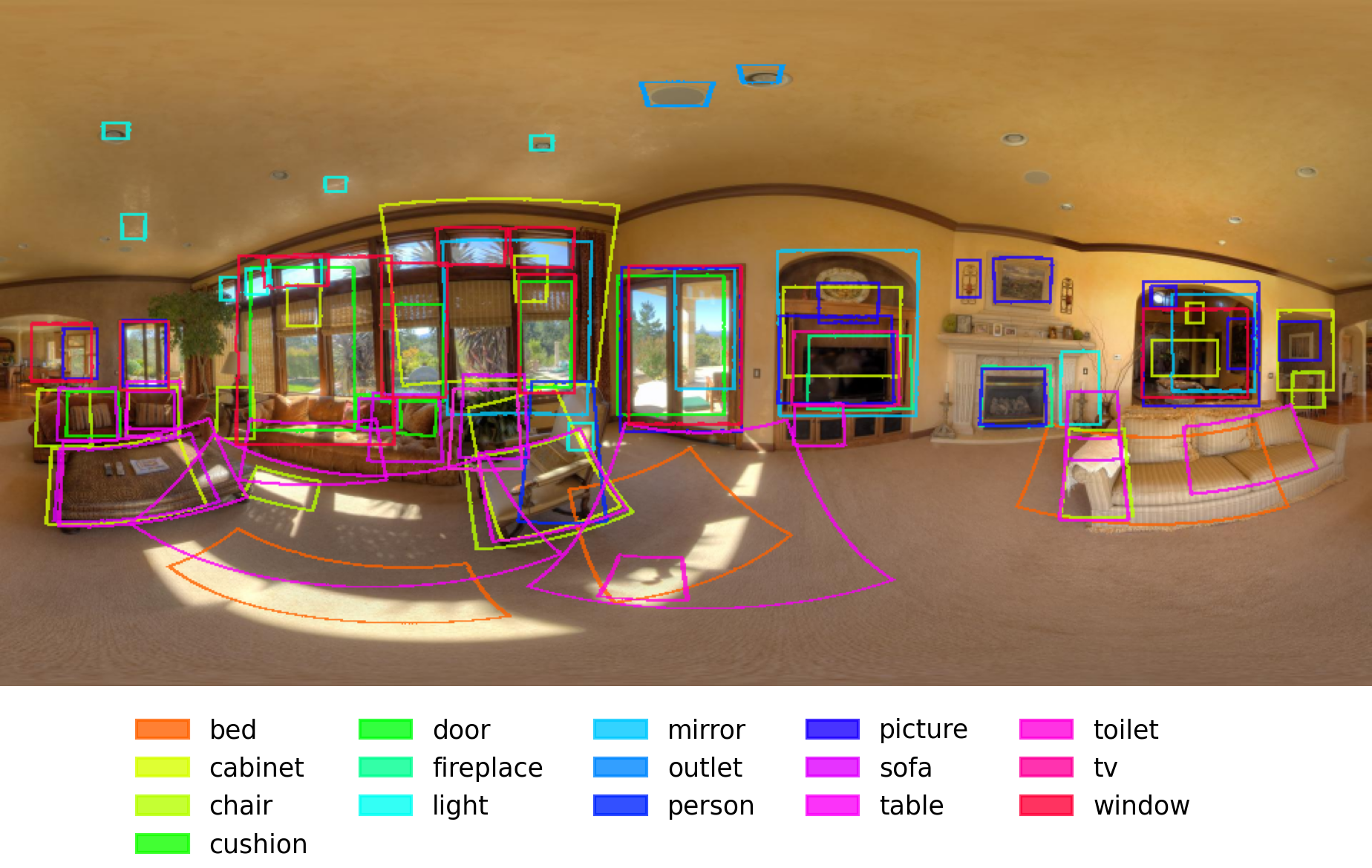}    & \cellcolor{lightred} \includegraphics[valign=c,width=.35\linewidth,keepaspectratio]{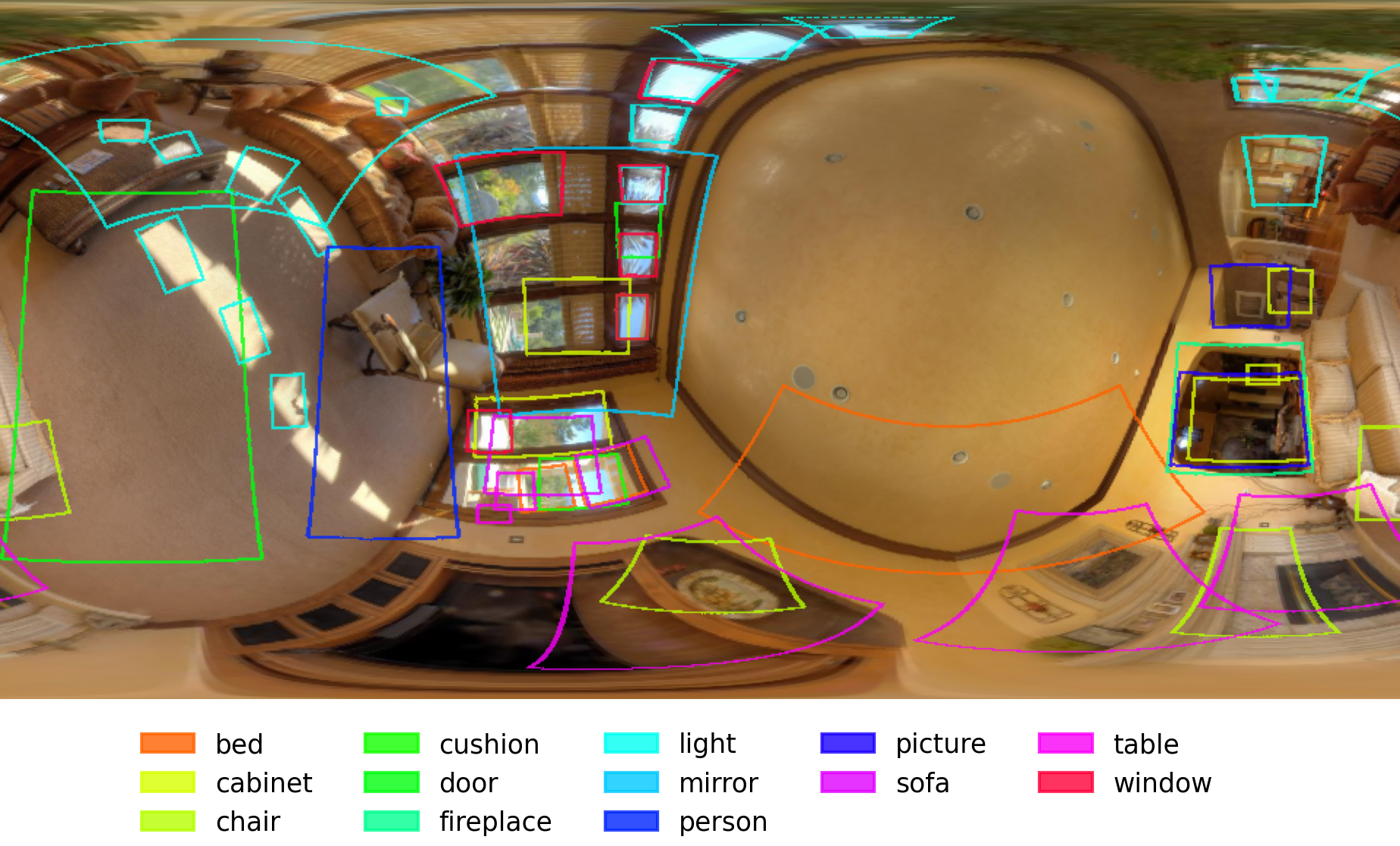} \\
		                                                                                                                                                                                                                                                                                                                                                                                                                             & \multicolumn{1}{|c|}{RR}                                                          & \includegraphics[valign=c,width=.35\linewidth,keepaspectratio]{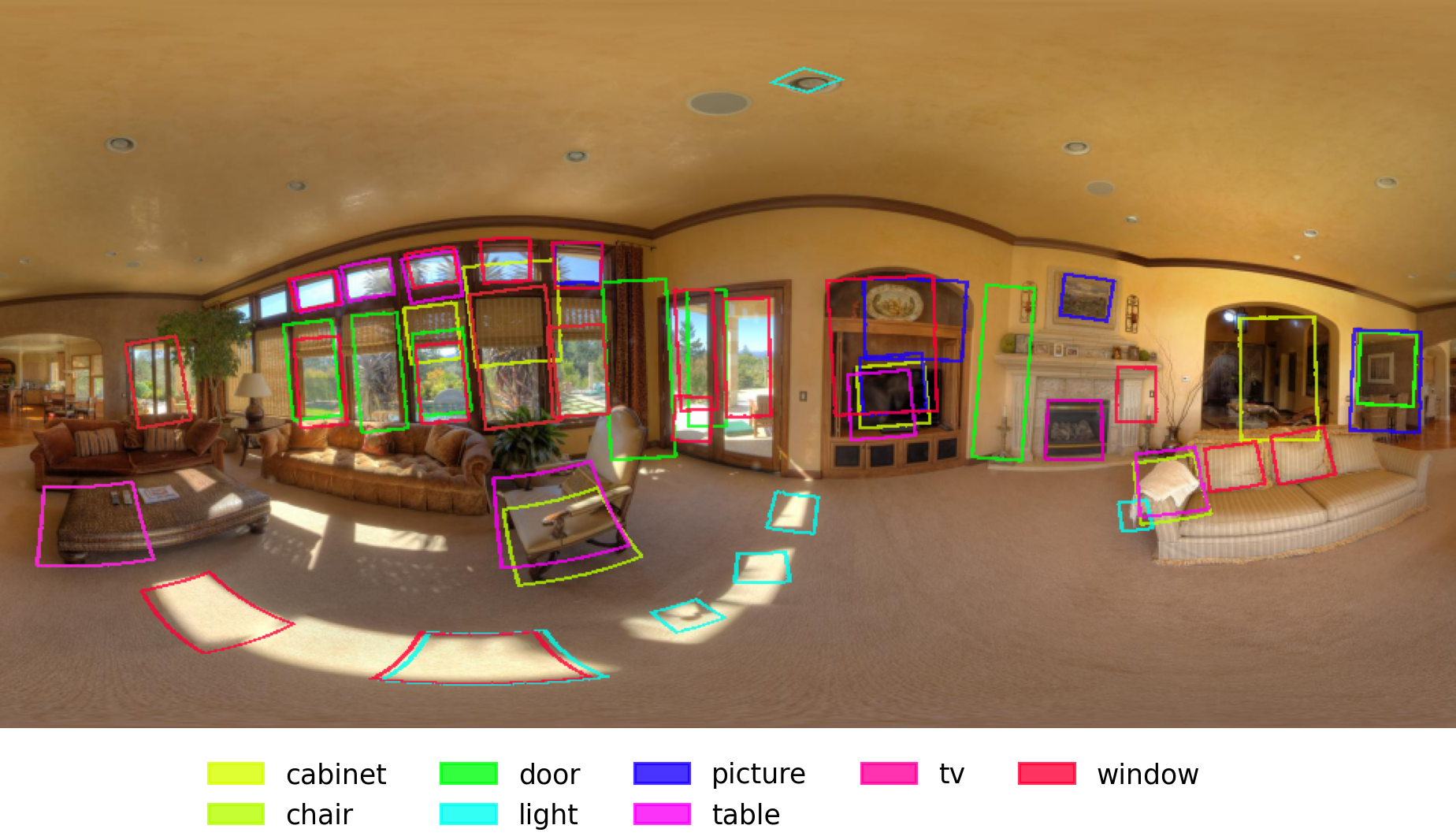}     & \includegraphics[valign=c,width=.35\linewidth,keepaspectratio]{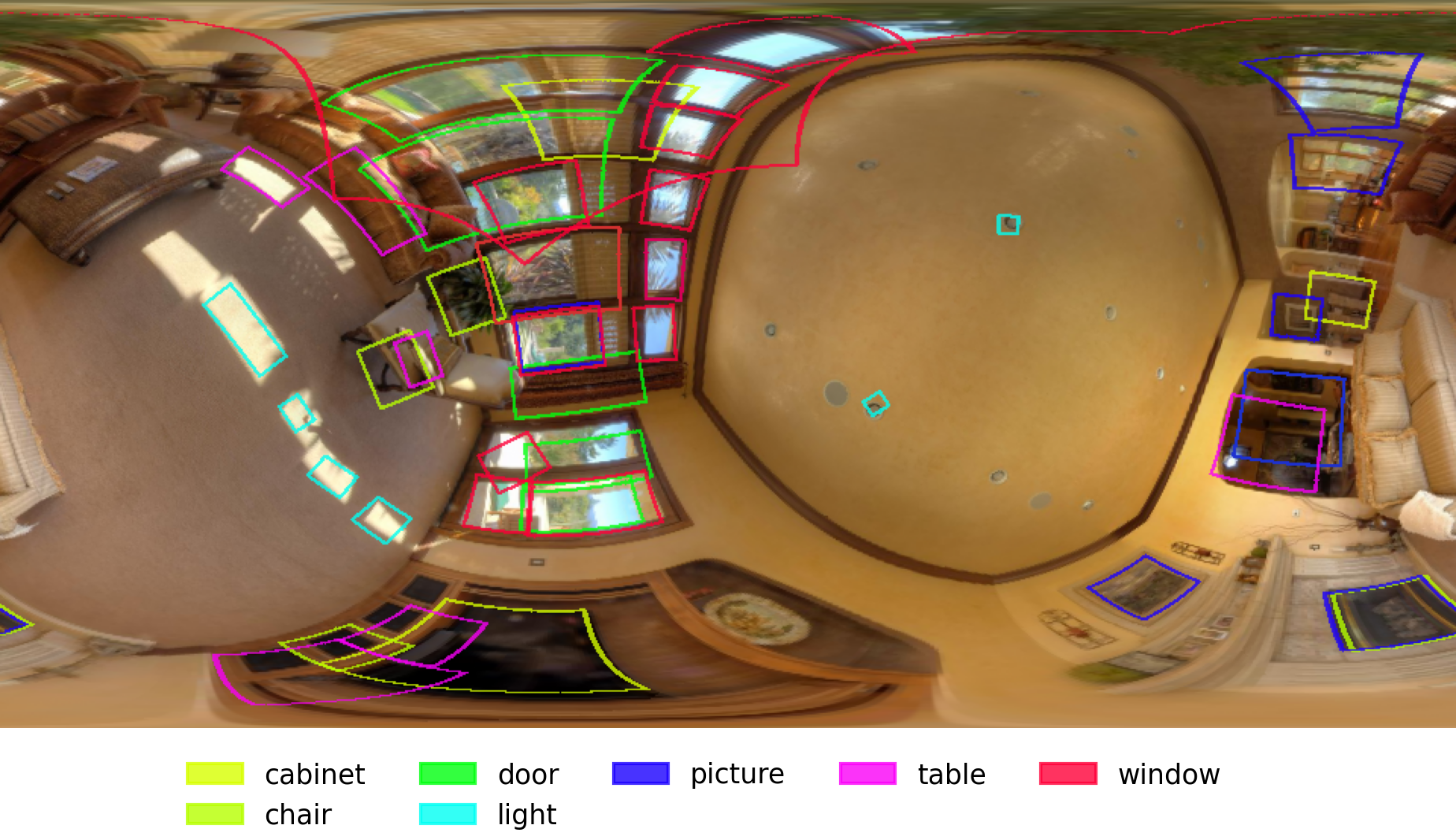}                       \\
		\midrule \makecell{Spherical                                                                                                                                                                                                                                                                                                                                                  \\YOLOv11}                                     & \multicolumn{1}{|c|}{NR}                                                          & \includegraphics[valign=c,width=.35\linewidth,keepaspectratio]{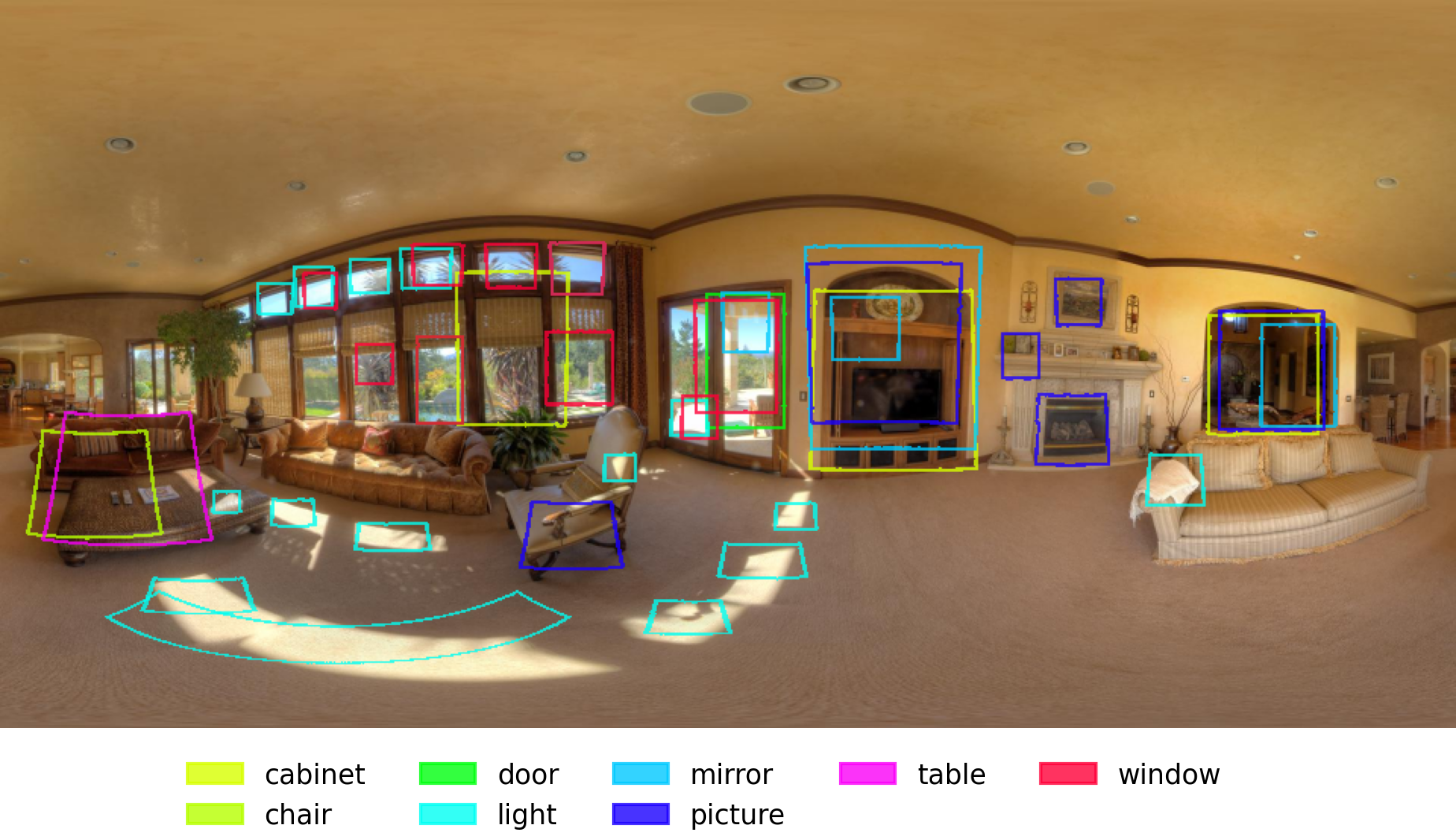} & \includegraphics[valign=c,width=.35\linewidth,keepaspectratio]{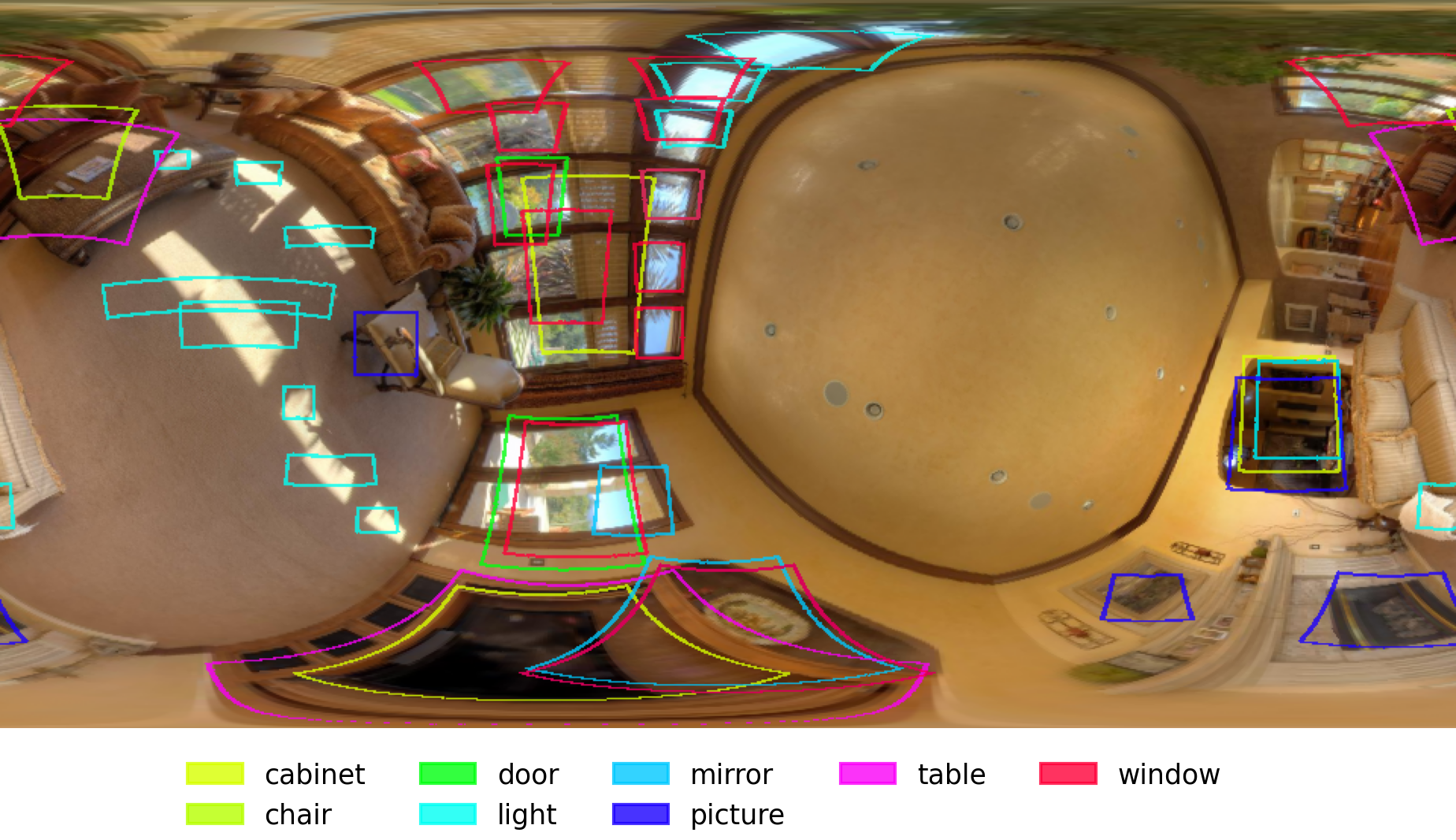}                   \\
		\midrule \multicolumn{2}{c|}{Ground Truth}                                                                                                                                                                                                                                                                                                                                                                                   & \includegraphics[valign=c,width=.35\linewidth,keepaspectratio]{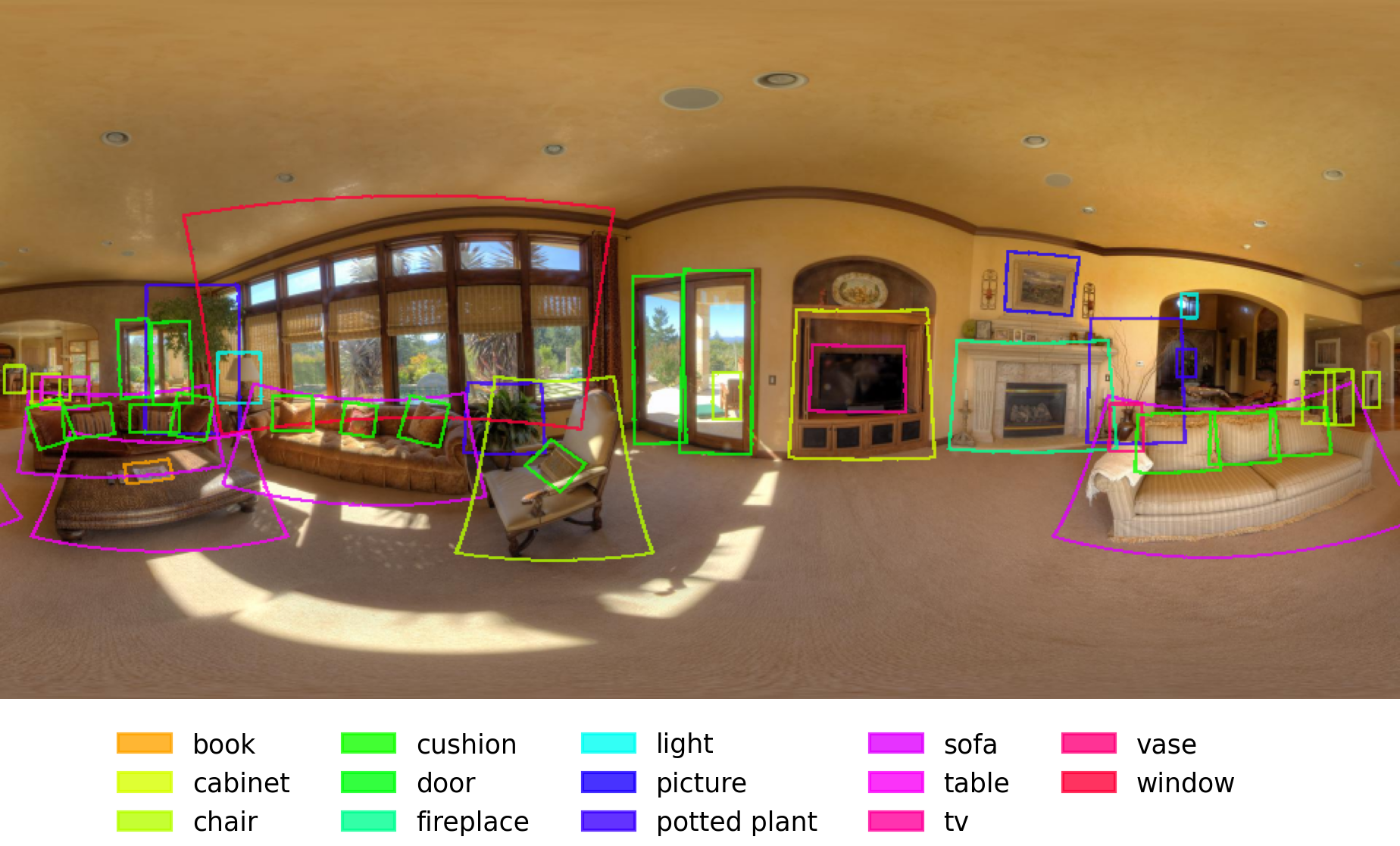} & \includegraphics[valign=c,width=.35\linewidth,keepaspectratio]{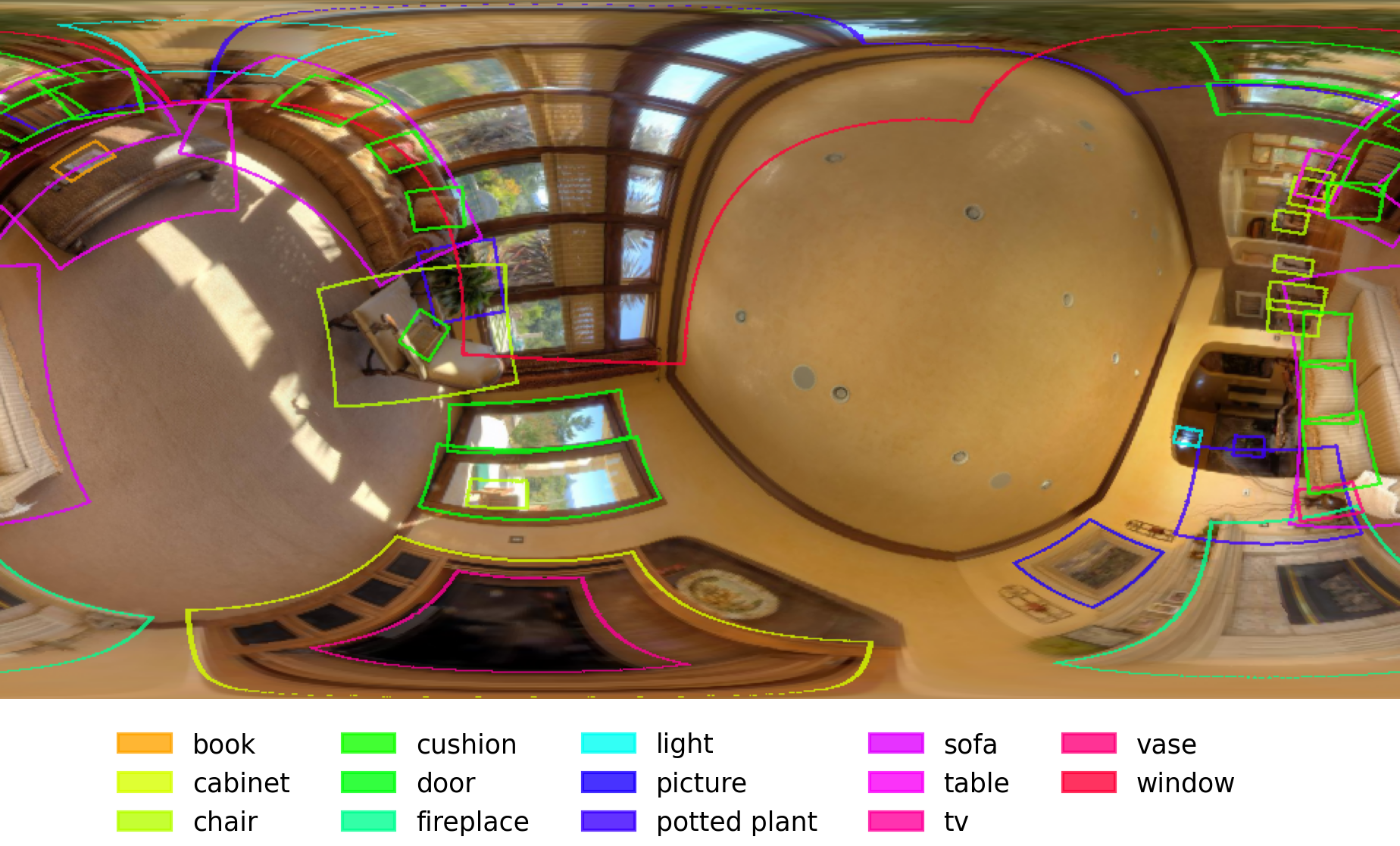}                                                                                                                             \\
		\bottomrule
	\end{tabular}
	\caption{\textbf{Object Detection Qualitative Result.}}
	\label{tab: object detection qualitative}
\end{table*}

%% file: tab/Semantic_Segmentation_Random_Rotation_Qualitative.tex
\begin{table*}
	[!ht]
	\centering
	\begin{tabular}{cc cc}
		\toprule \multicolumn{2}{c|}{\diagbox[outerleftsep=-40pt,outerrightsep=-40pt]{\textbf{Train}}{\textbf{Test}}}                                                                                                                                                                                                                                                                                                                                         & NR                                                                                & RR                                                                                                                                                                                                           \\
		\midrule \multirow{2}{*}[-4.5em]{\makecell{Planar                                                                                                                                                                                                                                                                                                                                            \\DeepLab v3\cite{chenRethinkingAtrousConvolution2017}}} & \multicolumn{1}{|c|}{NR}                                                          & \includegraphics[valign=c,width=.35\linewidth,keepaspectratio]{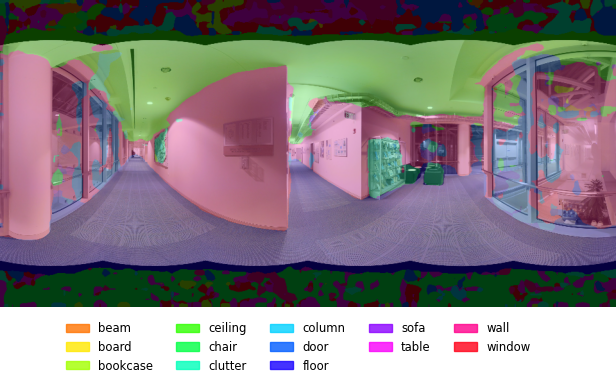}    & \cellcolor{lightred} \includegraphics[valign=c,width=.35\linewidth,keepaspectratio]{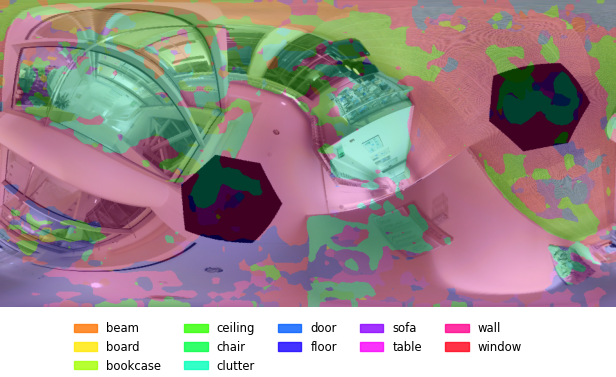} \\
		                                                                                                                                                                                                                                                                                                                                                                                                                                                      & \multicolumn{1}{|c|}{RR}                                                          & \includegraphics[valign=c,width=.35\linewidth,keepaspectratio]{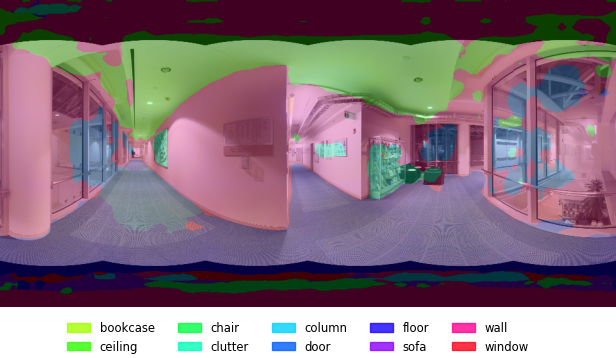}     & \includegraphics[valign=c,width=.35\linewidth,keepaspectratio]{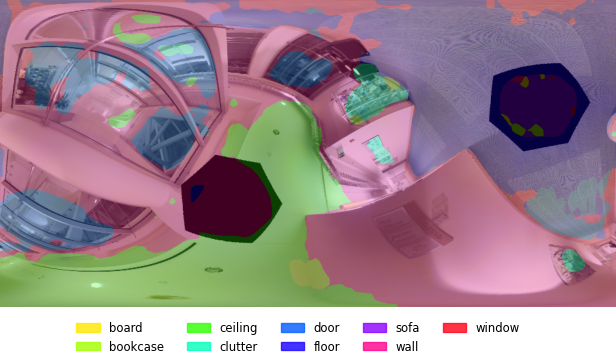}                       \\
		\midrule \multirow{2}{*}[-4.5em]{\makecell{Spherical                                                                                                                                                                                                                                                                                                                                         \\DeepLab v3}}                                           & \multicolumn{1}{|c|}{NR}                                                          & \includegraphics[valign=c,width=.35\linewidth,keepaspectratio]{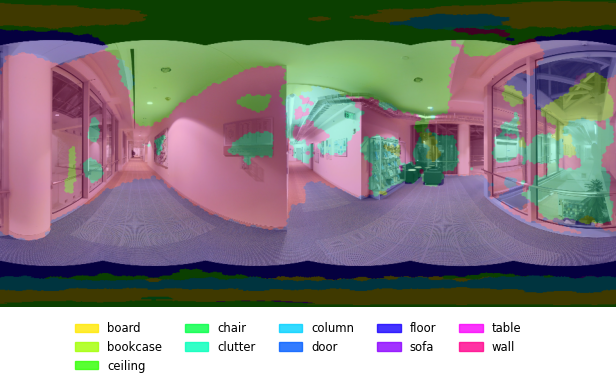} & \includegraphics[valign=c,width=.35\linewidth,keepaspectratio]{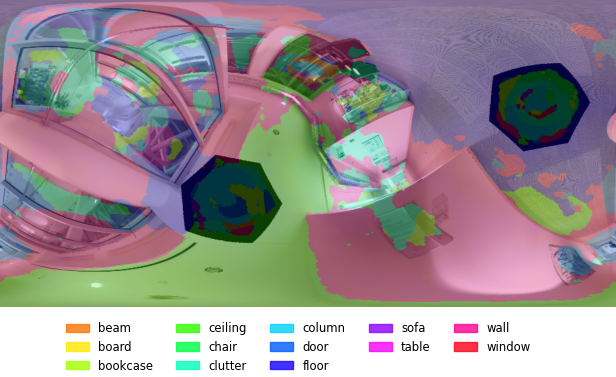}                   \\
		                                                                                                                                                                                                                                                                                                                                                                                                                                                      & \multicolumn{1}{|c|}{RR}                                                          & \includegraphics[valign=c,width=.35\linewidth,keepaspectratio]{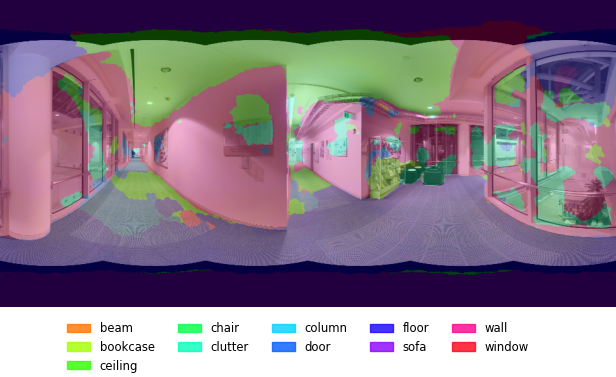}  & \includegraphics[valign=c,width=.35\linewidth,keepaspectratio]{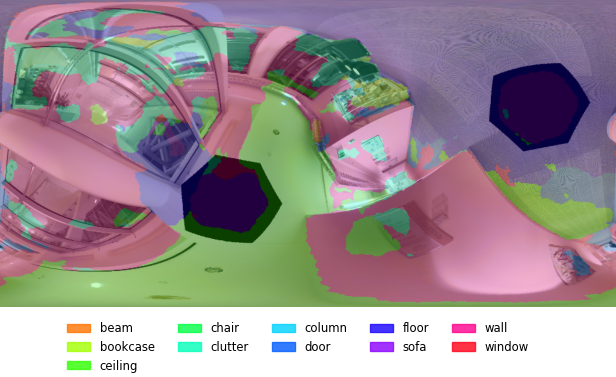}                    \\
		\midrule \multicolumn{2}{c|}{Ground Truth}                                                                                                                                                                                                                                                                                                                                                                                                            & \includegraphics[valign=c,width=.35\linewidth,keepaspectratio]{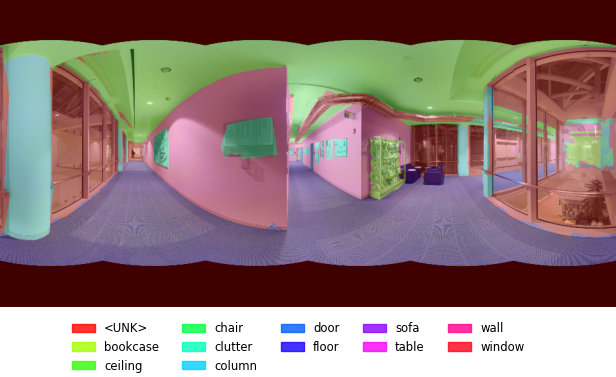} & \includegraphics[valign=c,width=.35\linewidth,keepaspectratio]{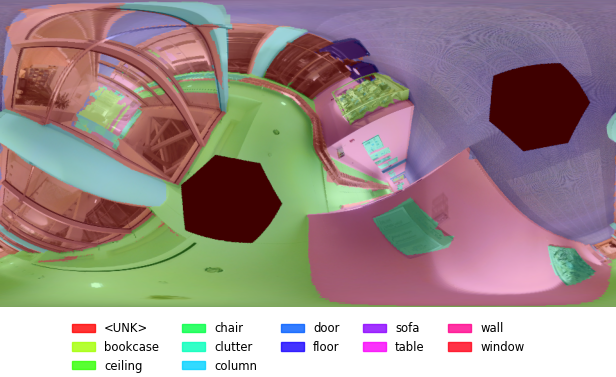}                                                                                                                             \\
		\bottomrule
	\end{tabular}
	\caption{\textbf{Semantic Segmentation Qualitative Result.}}
	\label{tab: semantic segmentation random rotation qualitative}
\end{table*}

%% file: tab/Full-dataset_Lens_Adaptation_Qualitative.tex
\begin{table*}
	[ht]
	\centering
	\begin{tabular}{cc ccc}
		\toprule \multicolumn{2}{c|}{\diagbox[outerleftsep=-40pt,outerrightsep=-40pt]{\textbf{Train}}{\textbf{Test}}}                                                                                                                                                                                                                                                                                                                                                                                                                                                                                                                                               & Pinhole                                                                                 & Fisheye                                                                                                                       & Panoramic                                                                                                                                                                                                                                                    \\
		\midrule \multirow{3}{*}[-7.0em]{\makecell{Planar                                                                                                                                                                                                                                                                                                                                                                                                                                                                                                                                                  \\DeepLab v3\cite{chenRethinkingAtrousConvolution2017}}} & \multicolumn{1}{|c|}{Pinhole}                                                           & \includegraphics[valign=c,width=.2\linewidth,keepaspectratio]{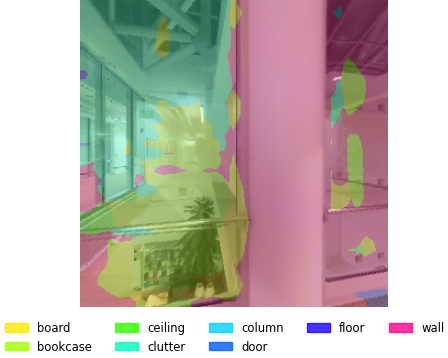}                           & \cellcolor{lightred} \includegraphics[valign=c,width=.2\linewidth,keepaspectratio]{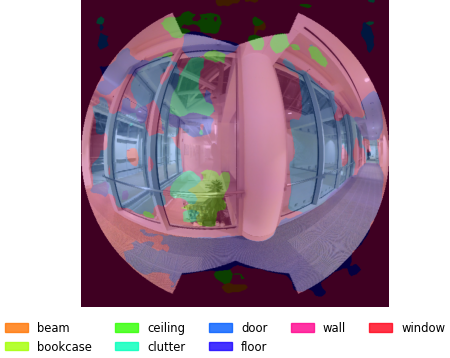}    & \cellcolor{lightred} \includegraphics[valign=c,width=.25\linewidth,keepaspectratio]{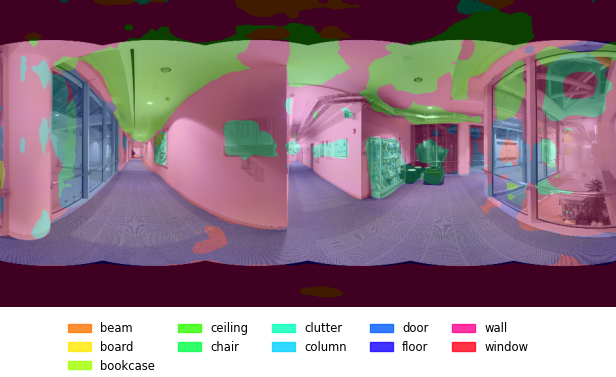}    \\
		                                                                                                                                                                                                                                                                                                                                                                                                                                                                                                                                                                                                                                                            & \multicolumn{1}{|c|}{Fisheye}                                                           & \includegraphics[valign=c,width=.2\linewidth,keepaspectratio]{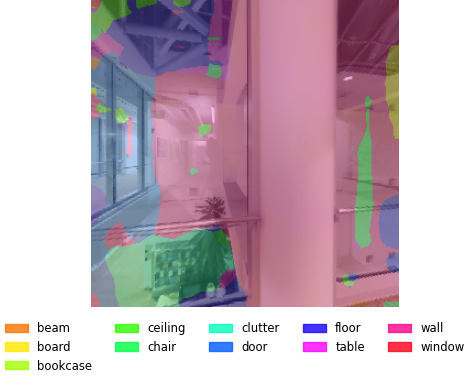}                           & \includegraphics[valign=c,width=.2\linewidth,keepaspectratio]{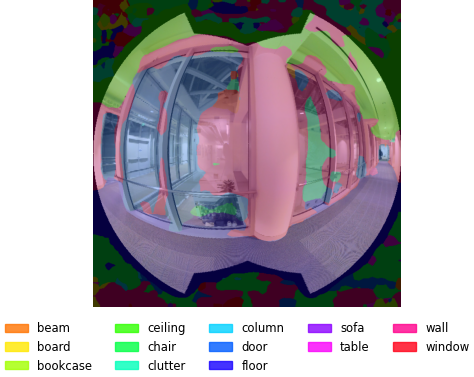}                         & \cellcolor{lightred} \includegraphics[valign=c,width=.25\linewidth,keepaspectratio]{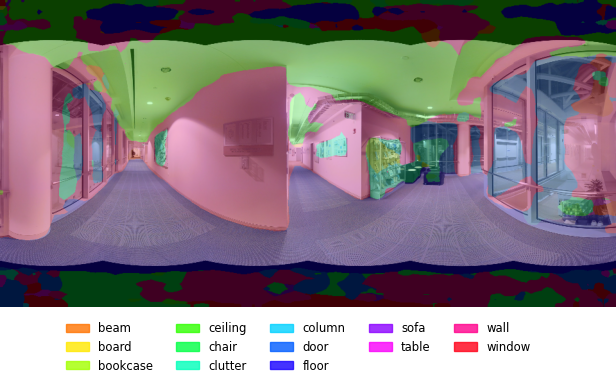}    \\
		                                                                                                                                                                                                                                                                                                                                                                                                                                                                                                                                                                                                                                                            & \multicolumn{1}{|c|}{Panoramic}                                                         & \cellcolor{lightred} \includegraphics[valign=c,width=.2\linewidth,keepaspectratio]{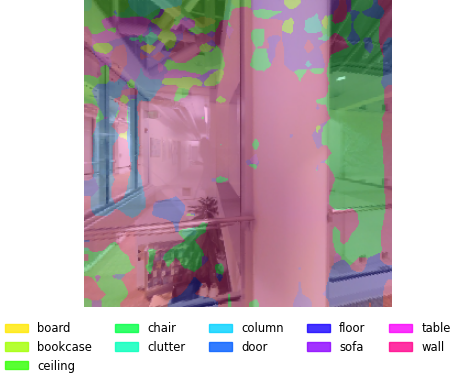}    & \cellcolor{lightred} \includegraphics[valign=c,width=.2\linewidth,keepaspectratio]{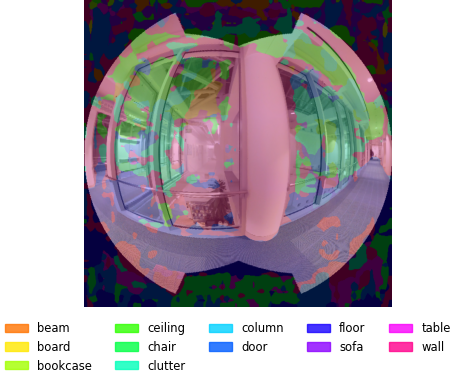}  & \includegraphics[valign=c,width=.25\linewidth,keepaspectratio]{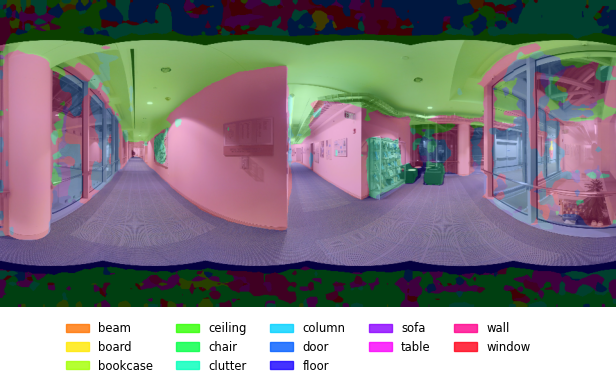}                       \\
		\midrule \multirow{3}{*}[-7.0em]{\makecell{Spherical                                                                                                                                                                                                                                                                                                                                                                                                                                                                                                                                               \\DeepLab v3}}                                           & \multicolumn{1}{|c|}{Pinhole}                                                           & \includegraphics[valign=c,width=.2\linewidth,keepaspectratio]{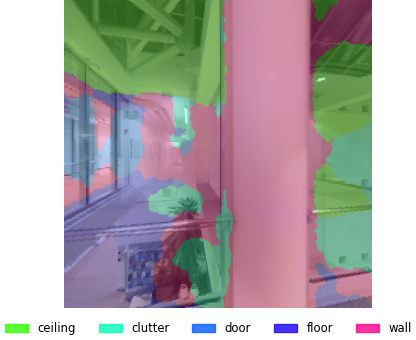}                        & \cellcolor{lightred} \includegraphics[valign=c,width=.2\linewidth,keepaspectratio]{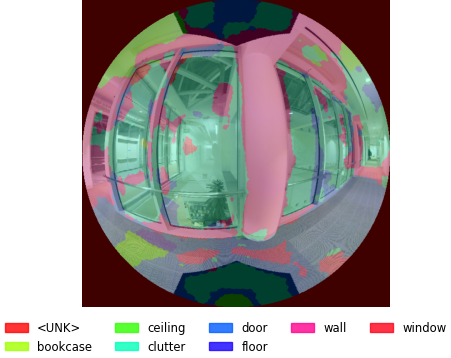} & \cellcolor{lightred} \includegraphics[valign=c,width=.25\linewidth,keepaspectratio]{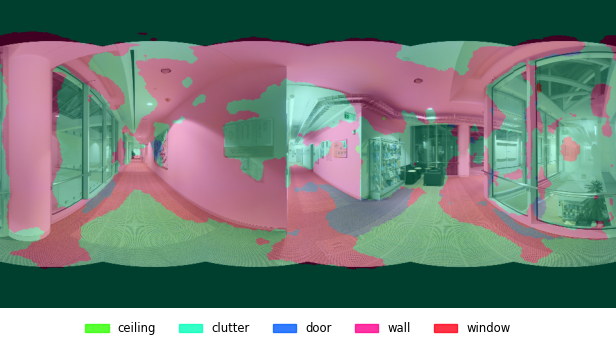} \\
		                                                                                                                                                                                                                                                                                                                                                                                                                                                                                                                                                                                                                                                            & \multicolumn{1}{|c|}{Fisheye}                                                           & \includegraphics[valign=c,width=.2\linewidth,keepaspectratio]{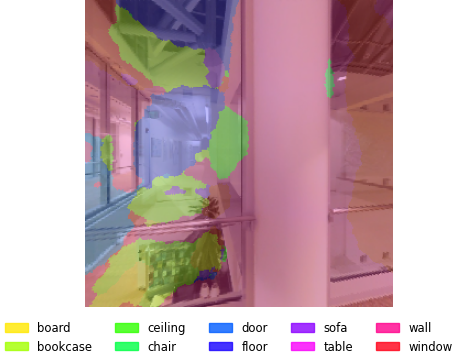}                        & \includegraphics[valign=c,width=.2\linewidth,keepaspectratio]{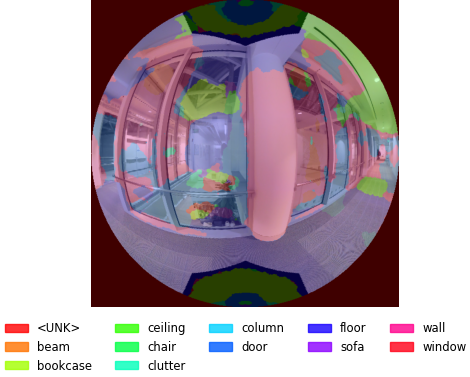}                      & \includegraphics[valign=c,width=.25\linewidth,keepaspectratio]{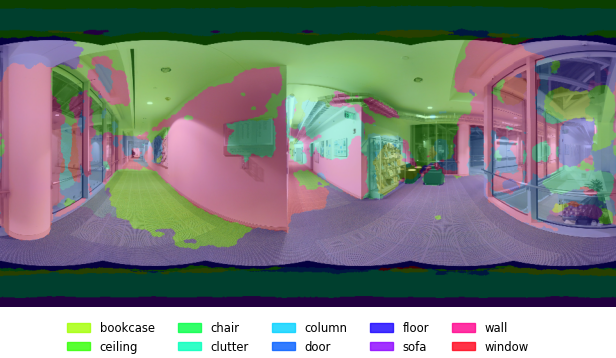}                      \\
		                                                                                                                                                                                                                                                                                                                                                                                                                                                                                                                                                                                                                                                            & \multicolumn{1}{|c|}{Panoramic}                                                         & \cellcolor{lightred} \includegraphics[valign=c,width=.2\linewidth,keepaspectratio]{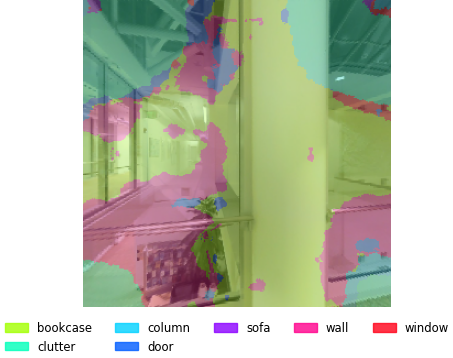} & \includegraphics[valign=c,width=.2\linewidth,keepaspectratio]{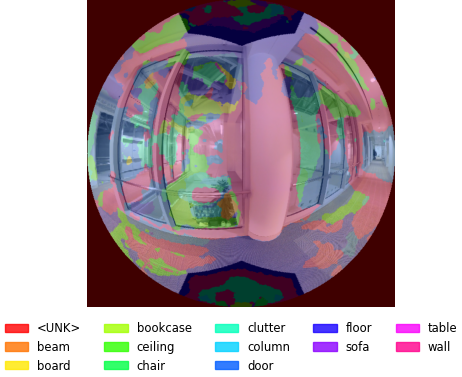}                    & \includegraphics[valign=c,width=.25\linewidth,keepaspectratio]{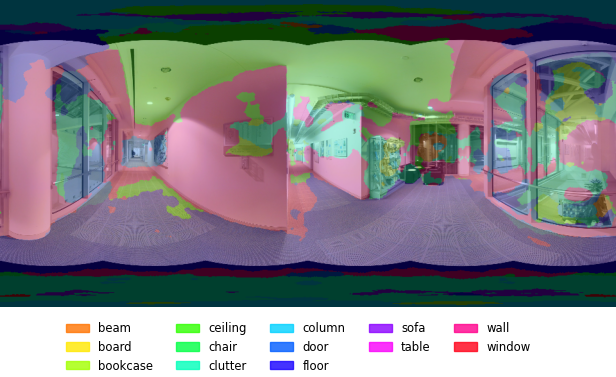}                    \\
		\midrule \multicolumn{2}{c|}{Ground Truth}                                                                                                                                                                                                                                                                                                                                                                                                                                                                                                                                                                                                                  & \includegraphics[valign=c,width=.2\linewidth,keepaspectratio]{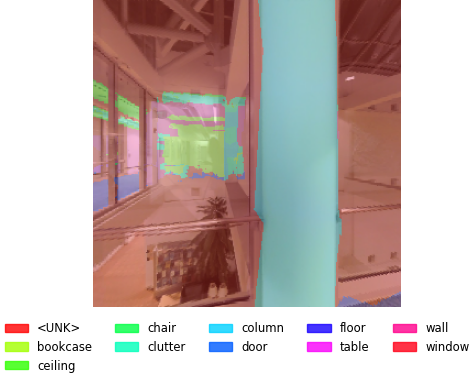} & \includegraphics[valign=c,width=.2\linewidth,keepaspectratio]{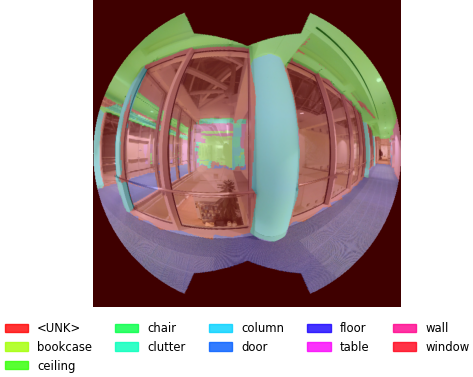}                                       & \includegraphics[valign=c,width=.25\linewidth,keepaspectratio]{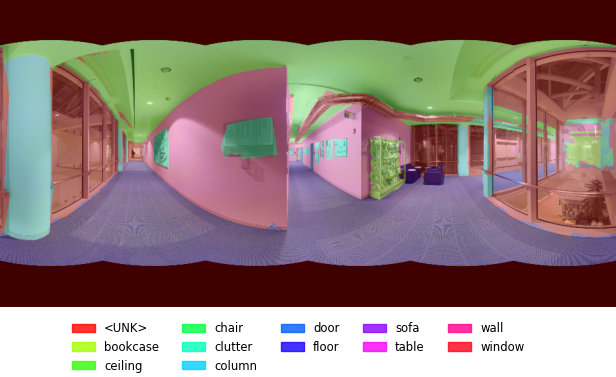}                                                                                                                                                                   \\
		\bottomrule
	\end{tabular}
	\caption{ \textbf{Semantic Segmentation Full-dataset Lens Adaptability Test.}
		Random Rotation is disabled.}
	\label{tab: full-dataset lens adaptation qualitative}
\end{table*}